\documentclass[smallcondensed]{svjour3}       

\usepackage{amsmath}
\usepackage{amsfonts}
\usepackage{mathtools}
\usepackage{amssymb}

\usepackage[utf8]{inputenc}
\usepackage{url}

\usepackage{algorithm}
\usepackage{algcompatible}
\algnewcommand{\YIELD}{\STATE{\textbf{yield }}}

\usepackage{xspace}
\newcommand{\mathsymbol}[2]{\newcommand{#1}{\ensuremath{\mathit{#2}}\xspace}}
\newcommand{\textmacro}[2]{\mathsymbol{#1}{\text{#2}}}
\mathsymbol{\hyperband}{\text{\sc Hyperband}}

\mathsymbol{\dataspace}{\mathcal{D}}
\mathsymbol{\searchspace}{\mathcal{P}}
\mathsymbol{\instancespace}{\mathcal{X}}
\mathsymbol{\labelspace}{\mathcal{Y}}
\mathsymbol{\risk}{\mathcal{R}}
\mathsymbol{\performance}{\phi}
\mathsymbol{\void}{\bot}

\textmacro{\blackbox}{black-box}

\textmacro{\automl}{AutoML}
\textmacro{\naml}{Naive \automl}
\textmacro{\snaml}{Quasi-Naive \automl}
\textmacro{\mlplan}{ML-Plan}
\textmacro{\autoweka}{Auto-WEKA}
\textmacro{\autosklearn}{auto-sklearn}
\textmacro{\gama}{GAMA}
\textmacro{\tpot}{TPOT}
\textmacro{\recipe}{RECIPE}
\textmacro{\exdef}{``Ex-def''}

\usepackage{booktabs}

\begin{document}


\institute{
    Felix Mohr, Universidad de La Sabana, Chía, Colombia (felix.mohr@unisabana.edu.co)\\
    Marcel Wever, Paderborn University, Germany (marcel.wever@uni-paderborn.de)\\
}

\vspace{-1cm}

\author{Felix Mohr, Marcel Wever}
\title{Naive Automated Machine Learning}
\maketitle

\vspace{-1em}

\begin{abstract}
An essential task of Automated Machine Learning (\automl) is the problem of automatically finding the pipeline with the best generalization performance on a given dataset.
This problem has been addressed with sophisticated \blackbox optimization techniques such as Bayesian Optimization, Grammar-Based Genetic Algorithms, and tree search algorithms.
Most of the current approaches are motivated by the assumption that optimizing the components of a pipeline in isolation may yield sub-optimal results.
We present \naml, an approach that does precisely this: It optimizes the different algorithms of a pre-defined pipeline scheme in isolation.
The finally returned pipeline is obtained by just taking the best algorithm of each slot.
The isolated optimization leads to substantially reduced search spaces, and, surprisingly, this approach yields comparable and sometimes even better performance than current state-of-the-art optimizers.
\end{abstract}

\section{Introduction}
An important task in Automated machine learning (\automl) is the one of automatically finding the pre-processing and learning algorithms with the best generalization performance on a given dataset.
The combination of such algorithms is typically called a (machine learning) \emph{pipeline} because several algorithms for data manipulation and analysis are put into (partial) order.
The choices to be made in pipeline optimization include the algorithms used for feature pre-processing and learning as well as the hyper-parameters of the chosen algorithms.

Maybe surprisingly, all common approaches to this problem try to optimize over all decision variables \emph{simultaneously} \cite{autoweka,feurer2015efficient,olson2016tpot,mohr2018ml}, and it has, to our knowledge, never been tried to optimize the different components in isolation.
While it is nearby that there are significant interactions between the optimization decisions, one can argue that achieving a \emph{global} optimum by \emph{local} optimization of components could be at least considered a relevant baseline to compare against.

We present two approaches for pipeline optimization  that do exactly that:
They optimize a pipeline locally instead of globally.
The most extreme approach, \naml, pretends that a locally optimal decision is also globally optimal, i.e., the optimality of a local decision is \emph{independent} of how other components are chosen.
In practice, this means that all components that are not subject to a local optimization process are left blank, except the learner slot, e.g., classifier or regressor, which is configured with some arbitrary default algorithm, e.g., decision tree, in order to obtain a valid pipeline.
Since \naml might sometimes be \emph{too} naive, we consider a marginally less extreme optimizer, called \snaml, which assumes an \emph{order} in which components are optimized and applies the naivety assumption only for the \emph{upcoming} decisions.
That is, it assumes that the quality of local optimization decisions for a component may be influenced by earlier optimized components but not by components that will be optimized subsequently.

On top of naivety, both \naml and \snaml pretend that parameter optimization is irrelevant for choosing the best algorithm for each slot.
That is, they assume that the best algorithm under default parametrization is also the best among all tuned algorithms.
Therefore, both \naml and \snaml optimize a slot by first selecting an algorithm and then applying a random search in the space of parameters of each chosen algorithm.

Our experimental evaluation shows that these simple techniques are surprisingly strong when compared against state-of-the-art optimizers used in \autosklearn and \gama.
While \naml is outperformed in the long run (24h), it is competitive with state-of-the art approaches in the short run (1h runtime).
On the contrary, \snaml is not only competitive in the \emph{long} run (24h) by achieving a de-facto optimal performance in over 90\% of the cases but even outperforms the state-of-the-art techniques in the short run.

While these results might suggest \snaml as a meaningful baseline over which one should be able to substantially improve, we see the actual role of \snaml as the door opener for sequential optimization of pipelines.
The currently applied \blackbox optimizers come with a series of problems discussed in recent literature such as lack of interpretability and flexibility \cite{drozdal2020trust,crisan2021fits}.
The naive approaches follow a sequential optimization approach, optimizing one component after the other.
While interpretability and flexibility are not a topic in this paper, they can be arguably realized more easily in custom sequential optimization approaches than in black-box optimization approaches.
The strong results of \snaml seem like a promise that extensions of \snaml such as \cite{mohr2021replacing} could overcome the above problems of \blackbox optimizers \emph{without sacrificing} global optimality.
We discuss this in more depth in Sec.~\ref{sec:results:discussion}.


\section{Problem Definition}
\label{sec:problem}
Even though the vision of \automl is much broader, a core task of \automl addressed by most \automl contributions is to automatically \emph{compose} and \emph{parametrize} machine learning algorithms to maximize a given metric such as accuracy.

In this paper, we focus on \automl for supervised learning.
Formally, in the supervised learning context, we assume some \emph{instance space} $\instancespace \subseteq \mathbb{R}^d$ and a \emph{label space} \labelspace.
A \emph{dataset} $D \subset \{(x,y)~|~x\in \instancespace, y \in \labelspace\}$ is a \emph{finite} relation between the instance space and the label space, and we denote as \dataspace the set of all possible datasets.
We consider two types of operations over instance and label spaces:
\vspace{-.5em}
\begin{enumerate}
    \item \emph{Pre-processors.}
    A pre-processor is a function $t: \instancespace_A \rightarrow \instancespace_B$, converting an instance $x$ of instance space $\instancespace_A$ into an instance of another instance space $\instancespace_B$.
    \item \emph{Predictors.}
    A predictor is a function $p: \instancespace_p \rightarrow \labelspace$, assigning an instance of its instance space $\instancespace_p$ a label in the label space \labelspace.
\end{enumerate}
\vspace{-.5em}
In this paper, a \emph{pipeline} $P = t_1 \circ .. \circ t_k \circ p$ is a functional concatenation in which $t_i: \instancespace_{i-1} \rightarrow \instancespace_i$ are pre-processors with $\instancespace_0 = \instancespace$ being the original instance space, and $p: \instancespace_k \rightarrow \labelspace$ is a predictor.
Hence, a pipeline is a function $P: \instancespace \rightarrow \labelspace$ that assigns a label to each object of the instance space.
We denote as \searchspace the space of all pipelines of this kind.
In general, the first part of a pipeline could be not only a sequence but also a \emph{pre-processing tree} with several parallel pre-processors that are then merged \cite{olson2016tpot}, but we do not consider such structures in this paper since they are not necessary for our key argument.
An extension to such tree-shaped pipelines is canonical future work.

In addition to the sequential structure, many \automl approaches restrict the search space still a bit further.
First, often a particular best order in which different \emph{types} of pre-processor should be applied is assumed.
For example, we assume that feature selection should be conducted after feature scaling.
So \searchspace will only contain pipelines compatible with this order.
Second, the optimal pipeline uses at most one pre-processor of each type.
These assumptions allow us to express \emph{every} element of \searchspace as a concatenation of $k+1$ functions, where $k$ is the number of considered pre-processor \emph{types}, e.g., feature scalers, feature selectors, etc.
If a pipeline does not adopt an algorithm of one of those types, say the $i$-th type, then $t_i$ will simply be the identity function.

The theoretical goal in supervised machine learning is to find a pipeline that minimizes the prediction error averaged over all instances from the same source as the given data.
This performance cannot be computed in practice, so instead one optimizes some function $\performance: \dataspace \times \searchspace \rightarrow \mathbb{R}$ that estimates the performance of a candidate pipeline based on some validation data.
Typical metrics used for this evaluation include error rate, least squares, AUROC, F1, log-loss, and others.

Consequently, a supervised \automl problem \emph{instance} is defined by a dataset $D\in \dataspace$, a search space \searchspace of pipelines, and a performance estimation metric $\performance: \dataspace \times \searchspace \rightarrow \mathbb{R}$ for solutions.
An \automl \emph{solver} $\mathcal{A}: \dataspace \rightarrow \searchspace$ is a function that creates a pipeline given some training set $D_{train} \subset D$.
The performance of $\mathcal{A}$ is given by
$
    \mathbb{E}\left[ ~\performance \big(D_{test}, \mathcal{A}(D_{train}) \big) \right],
$
where the expectation is taken with respect to the possible (disjoint) splits of $D$ into $D_{train}$ and $D_{test}$.
In practice, this score is typically computed taking a series of random binary splits of $D$ and averaging over the observed scores.
Naturally, the goal of any \automl solver is to optimize this metric, and we assume that $\mathcal{A}$ has access to $\phi$ (but not to $D_{test}$) in order to evaluate candidates with respect to the objective function.


\section{Related Work}
Even though the foundation of \automl is often attributed to the proposal of \autoweka, there have been some works on the topic long before.
Initial approaches date back to the 90s in the field of ``knowledge discovery in database systems'' \cite{engels96planningtasksforkddindatabases,morik2004miningmart}.
Another early work applied in the medical area is GEMS \cite{statnikov2005gems} in which the best pipeline of a pre-defined portfolio of configurations is selected based on a cross-validation.
In contrast, the work in \cite{engels96planningtasksforkddindatabases} searches a huge tree containing all possible pipeline configurations.
There are mainly three approaches following this direction, differing in the way how the search space is defined and how the search process is guided.
The first approach we are aware of was designed for the configuration of RapidMiner modules based on hierarchical planning \cite{kietz2009towards,kietz2012designing} most notably MetaMiner \cite{nguyen2012experimental,nguyen2014using}.
With ML-Plan \cite{mohr2018ml}, the idea of HTN-based graph definitions was later combined with a best-first search using random roll-outs to obtain node quality estimates.
Similarly, \cite{mosaic2019} introduced \automl based on Monte-Carlo Tree Search, which is closely related to ML-Plan.
However, the authors of \cite{mosaic2019} do not discuss the layout of the search tree, which is a crucial detail, because it is the primary channel to inject knowledge into the search problem.

Rather recently, it has also been recognized that random search is a quite competitive optimization algorithm for hyper-parameter optimization.
Relevant approaches in this respect are successive halving (SH) \cite{jamieson2016non} and \hyperband (HB) \cite{lihyperband2017}.
Similar to our naive approaches, these optimizers have a sequential aspect and greedily follow candidates that appear good early in the optimization process.
However, in contrast to the naive approaches, SH and HB are sequential in the \emph{evaluation budget} and not in the pipeline slots.
Our naive approaches can be configured with any type of validation function and in fact could adopt HB or SH for local hyper-parameter tuning.

Another line of research based on Bayesian Optimization was initialized with the advent of \autoweka \cite{autoweka,kotthoff2017auto}.
Like \naml, \autoweka assumes a fixed structure of the pipeline, admitting a feature selection step and a predictor.
The decisions are encoded into a large vector that is then optimized using SMAC \cite{hutter2011sequential}.
\autoweka optimizes pipelines with algorithms of the Java data analysis library WEKA \cite{hall2009weka}.
For the Python framework scikit-learn \cite{pedregosa2011scikit}, the same technique was adopted by \autosklearn \cite{feurer2015efficient}.
In the original version, \autosklearn added a data transformation step to the pipeline; meanwhile, the tool has been extended to support some more pre-processing functionalities in the pipeline.
Besides, \autosklearn features warm-starting and ensembling.
The main difference between these approaches and tree search is that tree search successively \emph{creates} solution candidates as paths of a tree instead of obtaining them from an \emph{acquisition function} as done by \autoweka and \autosklearn.

The idea of warm-starting introduced by \autosklearn was also examined in specific works based on recommendations.
Approaches here include specifically collaborative filtering like OBOE \cite{yang2019oboe}, probabilistic matrix factorization \cite{fusi2018probabilistic}, and recommendations based on average ranks \cite{cachada2017combining} .
These approaches are not necessarily requiring but are specifically designed for cases in which a database of past experiences on other datasets is available.



Another interesting line of research is the application of evolutionary and swarm algorithms.
One of the first approaches was PSMS \cite{escalante2009psms}, which used swarm particles for optimization.
A more recent approaches is TPOT \cite{olson2016tpot}.
In contrast to the above approaches, \tpot allows not just one pre-processing step but an arbitrary number of feature extraction techniques at the same time.
TPOT adopts a genetic algorithm to find good pipelines and adopts the scikit-learn framework to evaluate candidates.
Another approach is \recipe \cite{de2017recipe}, which uses a grammar-based evolutionary approach to evolve pipeline construction.
In this, it is similar to the tree search based approaches.
Focusing on the construction of stacking ensembles, another genetic approach was presented with AutoStacker \cite{chen2018autostacker}.
The most recent development in genetic algorithms for \automl is \gama \cite{gijsbers2019gama}, which we also consider in the experiments of this paper.

A recent line of research adopts a type of \blackbox optimization relying on the framework of multipliers (ADMM) \cite{boyd2011distributed}.
The main idea here is to decompose the optimization problem into two sub-problems for different variable types, considering that algorithm selection variables are Boolean while most parameter variables are continuous.
This approach was first presented in \cite{liu2020admm}.

Finally, a related approach is AutoGluon \cite{erickson2020autogluon}.
It is similar to the naive approaches in that it is also conducting a kind of sequential optimization process.
However, while our naive approach only optimizes base algorithms, the focus of AutoGluon is more on ensemble building through stacking and bagging.

Given this relatively rich list of approaches, it is a bit surprising that the naive approach has never been tried before.
Some tools such as \mlplan consider in a first phase only all possible default configurations.
This is a bit similar to the naive approach presented in this paper but does not yet decompose the search spaces.
Also, after this initial phase, the components are again optimized simultaneously.

\section{\naml and \snaml}

\subsection{Naivety Assumption}
\label{sec:naml:naivety}
\naml pretends that the optimal pipeline is the one that is locally best for each of its pre-processors and the final predictor.
In other words, taking into account pipelines with (up to) $k$ pre-processors and a predictor, we assume that for all datasets $D$ and all $1\leq i\leq k+1$
\begin{equation}
    c_i^* \in \arg \min_{c_i} \performance(D, c_1\circ..\circ c_{k+1})
    \label{eq:naivety}
\end{equation}
is \emph{invariant} to the choices of $c_1,..c_{i-1},c_{i+1},..,c_{k+1}$, which are supposed to be fixed in the above equation.
Note that we here use the letter $c$ instead of $t$ for pre-processors or $p$ for the predictor because $c$ may be any of the two types.

The typical approach to optimize the $c_i$ is not to directly construct those functions but to adopt parametrized model building processes that create these functions.
For example, $c_1$ could be a projection obtained by determining some features which we want to stay with, or $c_{k+1}$ could be a trained neural network.
These induction processes for the components can be described by an algorithm $a_i$ and a parametrization $\theta_i$ of the algorithm.
The component $c_i$ is obtained by running $a_i$ under parameters $\theta_i$ with some training data.
So to optimize $c_i$, we need to optimally choose $a_i$ and $\theta_i$.

We dub the approach \naml, because there is a direct link to the assumption made by the Naive Bayes classifier.
Consider \searchspace an urn and denote as $Y$ the event to observe an optimal pipeline in the urn.
Then
$$\mathbb{P}(Y~|~c_1,..,c_{k+1}) \propto \mathbb{P}(c_1,..,c_{k+1}~|~Y)\mathbb{P}(Y) \overset{naive}{=}\mathbb{P}(c_i~|~Y)\prod_{j=1,j\neq i}^{k+1}\mathbb{P}(c_j~|~Y)\mathbb{P}(Y),$$
in which we consider $c_j$ to be fixed components for $j \neq i$, and only $c_i$ being subject to optimization.
Applying Bayes' theorem again to $\mathbb{P}(c_i~|~Y)$ and observing that the remaining product is a constant regardless the choices of $c_{i\neq j}$, it gets clear that the optimal solution is the one that maximizes the probability of being locally optimal, and that this choice is \emph{independent} of the choice of the other components.

A direct consequence of the naivety assumption is that we can leave all components $c_{j\neq i}$ except the predictor component $c_{k+1}$ even \emph{blank} when optimizing $c_i$.
In practice, this should be done since it substantially reduces the runtime of candidate evaluations.
The reason why we cannot leave the predictor $c_{k+1}$ blank is, of course, that we cannot assess the performance of a pipeline that only has a pre-processor but no predictor.
However, under naivety we could just use \emph{any} available predictor, perhaps the fastest one.

It is clear that the naivety assumption does seldomly hold in practice.
One way to see this is the fact that it would enable us even to use a guessing predictor to optimize the pre-processing steps.
In fact, a reasonable default choice for the predictor would be the \emph{fastest} learner, and the arguably fastest algorithm is one that just guesses an output (or maybe always predicts the most common label).
It is unlikely that such a predictor is of much help when optimizing a pre-processor even if it is part of the candidates for $c_{k+1}$.

However, we can rescue the naivety approach by using some \emph{meaningful} fixed default predictor that somewhat ``represents'' the candidates available for $c_{k+1}$.
A reasonable choice could be a nearest neighbors or a decision tree predictor that at least take the features into account.
Of course, this standard predictor should be fixed a priori and not depend on the dataset.

In some cases, the naivety assumption will not hold even after this repair.
That is, there are datasets, on which combining classifier $c_1$ with pre-processor $p_1$ is better than combining it with $p_2$, but for another classifier $c_2$ it is better to combine it with $p_2$ than with $p_1$.
Such situations can be a problem for the naive approach, and the question is then how strong the performance gaps can get.

\subsection{Separate Algorithm Selection and Algorithm Configuration}
On top of the naivety assumption, \naml additionally pretends that even each component $c_i$ can be optimized by local optimization techniques.
More precisely, it is assumed that the algorithm that yields the best component when using the default parametrization is also the algorithm that yields the best component if all algorithms are run with the best parametrization possible.

Just like for the naivety assumption itself, we stress that this assumption is just an algorithmic \emph{decision}, which does not necessarily hold in practice.
In fact, the results on some datasets in the experiments clearly suggest that this assumption is not always correct.
However, this does not necessarily imply that the results will overly deteriorate by making these assumptions.
In a sense, our goal is precisely to study the \emph{extent} by which state-of-the-art approaches can improve over the naive approach by \emph{not} making this kind of simplifying assumptions.
Moreover, this gap is often surprisingly small, as the experiments in Sec. \ref{sec:results} show.

On the other side, the assumption might be less far-fetched than one might expect.
Our preliminary experiments showed that the tuning of parameters has often no or only a slight improvement over the performance achieved with the default configuration.
Keeping some exceptions like neural networks or SVMs in mind, experiments indicate that the variance of the variable describing the \emph{improvement} of a configuration over the performance with default configuration is relatively low for many learners -- at least for the here considered datasets.
Additionally, in some cases, we can also expand one algorithm with highly influencing parameters into several algorithms in which these parameters have already been set.
For example, we could simply treat support vector machines with different kernels and different (orders of) complexity constants as different algorithms.
In the large majority of algorithms, this practice does not yield an explosion in the algorithm space.
In fact, the only exception we can think of is indeed neural networks.



\subsection{The \naml Optimizer}
The \naml optimizer consists of two phases.
In a first phase, it just selects the best component for each slot of the pipeline based on default hyperparameter values.
To this end, \naml iterates over all slots and, for each slot, builds one pipeline for each component that can be filled into that slot.
The pipelines are constructed via the \emph{getPipeline} function, which creates a pipeline that contains only that single component or, in the case of pre-processors, an additional standard prediction component, e.g., a decision tree.
In Alg. \ref{alg:naiveautoml}, the first phase spans lines \ref{algline:shuffle} to \ref{algline:phase1:end}.
In a second phase, the algorithm runs in rounds in which it tries a new random parametrization for each of the components (in isolation).
If the performance of such a pipeline is better than the currently best, the parameters for that slot's component are updated correspondingly.
This is done until the time-bound is hit.
The whole algorithm works as a generator, and whenever a new best configuration is found, a new best pipeline $p^*$ is obtained.
This pipeline $p^*$ holds for each slot the best-found choice.
In Alg. \ref{alg:naiveautoml}, the second phase is described in lines \ref{algline:phase2:start} to \ref{algline:phase2:end}.

In order to make the performance of \naml, on average, independent of the order in which slots and algorithms are defined in the input, those sets are shuffled at the beginning (l. \ref{algline:shuffle} and \ref{algline:shuffle:inner}), so that the \emph{order} in which both the slots themselves and the candidates per slots are tested are subject to randomness.
Of course, in extensions the order of components can be suggested by warm-starting mechanisms, which is however not the focus of this paper.

\begin{algorithm}[t]
\caption{\naml\ - Optimization Routine}
\label{alg:naiveautoml}

\begin{algorithmic}[1]
\REQUIRE{Components $\mathcal{C} = (C_1,..,C_{k+1})$ for pipeline slots, validation function {\sc Validate}}
\STATE{$S \leftarrow$ shuffled list of $\{1,..,k+1\}$}\label{algline:shuffle}
\FORALL{slot $s \in S$}
    \STATE{$C_s \leftarrow$ shuffled list of candidates for slot $s$}\label{algline:shuffle:inner}
    \STATE{$c^*_s, \theta^*_s, v_s^*,v^* \leftarrow \bot, \bot, -\infty, -\infty$}
    \FORALL{candidate $c_s \in C_s$}
        \STATE{$v_s \leftarrow$ {\sc Validate}({\sc getPipeline}($k, c_s, \bot$))}
        \IF{$v_s > v_s^*$}
            \STATE{$c_s^*, v_s^* \leftarrow p, v_s$}
            \IF{$v_s > v^*$}
                \YIELD $((c^*_{s_1}, \bot), .., (c^*_{s}, \bot))$
            \ENDIF
        \ENDIF
    \ENDFOR
\ENDFOR \label{algline:phase1:end}

\WHILE{timeout not reached}\label{algline:phase2:start}
    \FORALL{slot $s \in S$}
        \STATE{$\theta_s \leftarrow$ random configuration for $c^*_s$}
        \STATE{$v_s \leftarrow$ {\sc Validate}({\sc getPipeline}($k, c_s, \theta_s$))}
        \IF{$v_s > v_s^*$}
            \STATE{$\theta^*_s, v_s^* \leftarrow \theta_s, v_s$}
            \IF{$v_s > v^*$}
                \YIELD $((c^*_{s_1}, \theta^*_{s_1}), .., (c^*_{s_{k+1}}, \theta^*_{s_{k+1}}))$
            \ENDIF
        \ENDIF
    \ENDFOR
\ENDWHILE\label{algline:phase2:end}
\end{algorithmic}

\end{algorithm}

Note that the returned pipelines $p^*$ are never executed as a whole internally, so the algorithm has, in fact, no estimate of their performance.
This is precisely where the naivety enters: The algorithm trusts that each local decision is optimal, so the pipeline composed of those locally optimal decisions is also expected to be globally optimal.
Hence, it is not necessary to have a concrete estimate of the performance of $p^*$.

Even though the first phase of \naml entirely fixes the algorithms of the pipeline, the hyperparameter optimization (HPO) phase optimizes each component in isolation.
Instead of optimizing slot after slot, each main HPO step performs one optimization step for each slot.
This procedure is repeated until the overall timeout is exhausted.

From the above presentation is becomes apparent that \naml is indeed not even a fully defined \automl tool but only an optimizer for the \automl context.
One indicator for this is that \naml itself does not train a final candidate.
A minimalistic \automl tool around \naml would, after gaining back control, train the last received pipeline on the full data.

Even more, it can happen that such an outer algorithm must ``repair'' the finally built pipeline if it is corrupt in the sense that it cannot be successfully trained.
Suppose that the last yielded pipeline is $p^*$.
It can happen (and in practice, it \emph{does} happen occasionally) that $p^*$ is not executable on specific data.
For example, a pipeline $p^*$ for scikit-learn \cite{pedregosa2011scikit} may contain a StandardScaler, which produces negative attribute values for some instances, and a MultinomialNB predictor, which cannot work with negative values.
Since the two components are never executed together \emph{during} search, the optimizer does not detect any problem with the two outputs StandardScaler and MultinomialNB in isolation.
Several repair possibilities would be imaginable, e.g., to replace the pre-processors with earlier found candidates for that slot, or to simply try earlier candidates of $p^*$.
To keep things simple, in this paper, we just removed pre-processors from left to the right until an executable pipeline ${p^*}'$ is created; in the extreme case leading just to a predictor without pre-processors.

\subsection{The \snaml Optimizer}
The \snaml Optimizer makes two minor changes in the above code of \naml.
First, the shuffle operation in line \ref{algline:shuffle} of the algorithm is replaced by a fixed permutation $\sigma$.
Second, the {\sc getPipeline} routine does not leave components of previous decision steps blank (or plugs in the default predictor) but puts in the \emph{default configured} component $c^*_s$ chosen for the respective slot $s$.
More formally, if $\sigma(i) < \sigma(j)$ and the algorithm is building a pipeline with slot $j$ as decision variable, then slot $i$ is filled with $(c^*_{s_i}, \bot)$.
Typically, $\sigma$ will order the predictor first and then assume some order of decisions on the pre-processors.

Under this adjustment, the naivety assumption in Eq. (\ref{eq:naivety}) is relaxed as follows.
Instead of assuming that \emph{all} other components are irrelevant for the best choice of a component in the pipeline, one now only pretends that the \emph{subsequently} chosen components are irrelevant for the optimal choice.
In contrast, the \emph{previously} made decisions \emph{are} relevant for the current optimization question.
Concerning the naivety assumption, they are relevant in the sense that the previously decided components cannot be chosen arbitrarily in the naivety property but are supposed to be fixed according to the choice that was made for that slot.

From a practical viewpoint, the strict \naml approach has almost no advantage over the \snaml approach.
The only plus offered by strict \naml is that one can optimize the different slots in parallel.
Intuitively, if no such parallelization is adopted and components are optimized in sequence, there is no good reason not first to find a best learner (classifier or regressor) and then use in each slot the choices made already earlier in other slots when filling a pipeline.
On the other hand, parallelization can also be adopted for the optimization process of a \emph{single} slot, so we would argue that pure \naml is rather of theoretical interest, e.g., in order to verify the appropriateness of the naivety assumption, but does not seem to have any other relevant practical advantage.

\section{Evaluation}
\label{sec:results}
We compare \naml and \snaml with state-of-the-art optimizers used in the context of AutoML.
We stress that we aim at comparing \emph{optimizers} and not whole AutoML tools.
That is, we explicitly abandon previous knowledge that can be used to warm-start an optimizer and also abandon post-processing techniques like ensembling \cite{feurer2015efficient,gijsbers2019gama} or validation-fold-based model selection \cite{mohr2018ml}.
Those techniques are (largely) orthogonal to the optimizer and hence irrelevant for its performance analysis.
This being said, it is, of course, possible that some optimizers benefit more from certain additional techniques like warm-starting etc. than others.
However, this kind of analysis is in the scope of studies that propose those kinds of techniques.

When comparing the naive approaches with state-of-the-art optimizers, we should recognize that the naive approaches are indeed \emph{very} weak optimizers.
First, in contrast to global optimizers, the naive approaches do not necessarily converge to an optimal solution because large parts of the search space are pruned early.
In other words, the naive approaches can only lose (or at best be competitive) in the long run.
Second, the highly stochastic nature of the algorithms also does not give high hopes for great performance in the short run.
Both \naml and \snaml are closely related to random search, which can be considered one of the most simple baselines\footnote{
In fact, \naml \emph{is} a random search in a decomposed search space: While the HPO phase is an explicit random search, the algorithm selection phase simply iterates over all possible algorithms, which is equivalent to a random search due to the small number of candidates (all of them are considered anyway).}.
The only reason to believe in \naml seems to be that it quickly commits to an apparently locally best component and that hyperparameter-tuned versions of those components will also occur in an optimal pipeline.

These observations then motivate three research questions:

\begin{itemize}
    \setlength{\itemindent}{3em}
    \item[RQ 1:] \emph{Do} the naive approaches find better pipelines than state-of-the-art optimizers in the short run?
    
    \item[RQ 2:] How \emph{often} do global optimizers outperform the naive approaches in the long run, and how \emph{long} do they need to take the lead?
    
    \item[RQ 3:] How \emph{large} is the performance gap between the solutions found by naive approaches compared to the best ones found by global optimizers?
\end{itemize}

To operationalize the terms ``short run'' and ``long run'', we choose time windows of 1h and 1d, respectively.
These time limits are, of course, arbitrary but are common practice and seem to represent a good compromise taking into account the ecological impact of such extensive experiments.

\subsection{Experiment Setup}
\subsubsection{Compared Optimizers and Search Space Definition}
The evaluation is focused on the machine learning package scikit-learn \cite{pedregosa2011scikit}.
On the state-of-the-art side, we compare solutions with the competitive AutoML tools \autosklearn and \gama.
Hence, as one reference optimizer, we consider Bayesian Optimization through the notion of \autosklearn, which adopts SMAC as its optimizer \cite{hutter2011sequential}.
We use version 0.12.6, which underwent substantial changes and improvements compared to the original version \cite{feurer2015efficient}.
As a second baseline, we compare against the \emph{genetic algorithm optimizer} proposed in \gama \cite{gijsbers2019gama}.
In order to isolate possibly confounding factors and to only compare optimization techniques, all pre- and post-processing activities such as warm-starting and ensemble building were deactivated in \autosklearn and \gama.
We are not aware of other approaches that have shown to substantially outperform these tools at the optimizer level.
Some works claim to outperform \autosklearn but only demonstrate that with rank plots, so the extend of improvement is unclear \cite{mosaic2019,liu2020admm}.

To maximize the comparability, we unified the search space among the compared optimizers as far as possible.
Since all optimizers except \autosklearn can be configured relatively easily in their search space and pipeline structure, we adopted the pipeline structure dictated by \autosklearn.
This pipeline consists of three steps, including so-called \emph{data-pre-processors}, which are mainly feature scalers, \emph{feature-pre-processors}, which are mainly feature selectors and decomposition techniques, and finally the estimator.
The appendix shows the concrete list of algorithms used for each category.
We also used the hyperparameter space defined by \autosklearn for each of the components.
Unfortunately, there are some (proprietary) components such as balancing and minority coalescer that cannot be deactivated in \autosklearn but also cannot be easily used in other tools.
This implies that the search spaces are not entirely identical, but an analysis of results suggests that those differences are probably not relevant for the comparison.
The search spaces of the naive approaches and \gama are almost identical.
The only difference is that \gama, at the time of writing, does only support explicitly defined domains for parameter values, which does not match the concept of numerical parameters used in \autosklearn and the naive approaches through the ConfigSpace library \cite{configspace}.
To overcome this problem, we sampled 10000 values for each \emph{parameter} and used these as a discrete space; this sampling mechanism already included log-scale sampling where applicable\footnote{All these efforts were realized in collaboration with and under the approval of the authors of \gama to ensure a maximally faithful adaption of the code for the purpose of this benchmark.}.

Implementations of the naive approaches and the experiments are available for the public \footnote{ \url{https://github.com/fmohr/naiveautoml/tree/mlj2021}}.
The repository also comes along with the data we used to create the result figures and tables.

\subsubsection{Benchmark Datasets}
The evaluation is based on the dataset portfolio proposed in \cite{gijsbers2019open}.
This is a collection of datasets available on openml.org \cite{OpenML2013}.
These datasets cover classification for both binary and multi-class classification with numerical and categorical attributes.
Within this scope, the dataset selection is quite diverse in terms of numbers of instances, numbers of attributes, numbers of classes, and distributions of types of attributes.
The appendix lists the relevant properties of each of these datasets to confirm this diversity.
Our assessment is hence limited to binary and multi-class classification.


For all datasets, categorical attributes were replaced by an Bernoulli encoding (one-hot-encoding) and missing values are replaced by 0 prior to passing it to the optimizer.
This was just with the purpose to avoid implicit search space differences, because \autosklearn comes with some pre-processors specifically tailored for categorical attributes.
Since these are partially proprietary and not easily applicable with \gama and the naive approaches, we simply eliminated this decision variable from the search space.
Hence, the optimizers construct pipelines that are applied to purely numerical datasets.
Of course, the imputation with 0 is often sub-optimal, but since the imputation is the same for all optimizers, it does not affect the comparison among them.

\subsubsection{Validation Mechanism and Performance Metrics}
With respect to validation, we standardized both the outer and the inner evaluation mechanism.
First, the outer evaluation mechanism splits the original data into \emph{train} and \emph{test} data.
For this, we chose a 90\% train fold size and a 10\% test fold size.
Running each optimizer 10 times with different such random splits corresponds to a 10 iterations Monte Carlo cross-validation with 90\% train fold size.
Of course, splits were identical per seed among all optimizers.
Second, the evaluation mechanism for a concrete pipeline candidate was fixed among all approaches to 5-fold cross-validation; no early stopping was applied.

Note that our primary focus here is \emph{not} on test performance but validation performance.
This paper compares \emph{optimizers}, so we should measure them in terms of what they optimize, namely validation performance.
It can clearly happen that strong optimization of that metrics yields no better or even worse performance on the test data (over-fitting).
Even though test performance is, in our view, not relevant for the research questions, we conduct the outer splits and hence provide test performance results in order to maximize insights.

Following the argumentation of Provost et al. \cite{provost98thecaseagainstaccurarcy}, we abstain from the use of accuracy as a performance measure for comparison.
Instead, we use area under the receiver operator curve (AUROC) as a performance measure of models on binary classification data as proposed in \cite{provost98thecaseagainstaccurarcy} and log-loss on multi-class classification data, as suggested in the context of the AutoML benchmark \cite{gijsbers2019open}.
Since these metrics are based on prediction \emph{probabilities}, they allow for more fine granular assessment.
A particular advantage of AUROC is that it is agnostic to class imbalance.

\subsubsection{Resources and used Hardware}
Timeouts were configured as follows.
For the short (long) run, we applied a total overall runtime of 1h (24h), and the runtime for a single pipeline execution was configured to take up to 5 (20) minutes.
The memory was set to 24GB and, despite the technical possibilities, we did \emph{not} parallelize evaluations.
That is, all the tools were configured to run with a single CPU core.
The computations were executed in a compute center with Linux machines, each of them equipped with 2.6Ghz Intel Xeon E5-2670 processors and 32GB memory.

\subsection{Results}
\subsubsection{RQ 1: \emph{Do} the naive approaches find better pipelines than state-of-the-art optimizers in the short run?}

To answer this question, we look at the results for an overall timeout of 1h.
Taking into account the timeouts allowed in some competitions, this time limit can even be considered kind of generous.
However, those competitions typically compare fully-fledged systems, making massive use of warm-starting, so a timeout of 1h seems appropriate when comparing cold-started optimizers.
Also, note that this is not the same as looking onto the first part of the 24h runs since the timeout per execution is also lower; it is hence a different setup.

\begin{figure}[t]
    \centering
    \includegraphics[width=\textwidth]{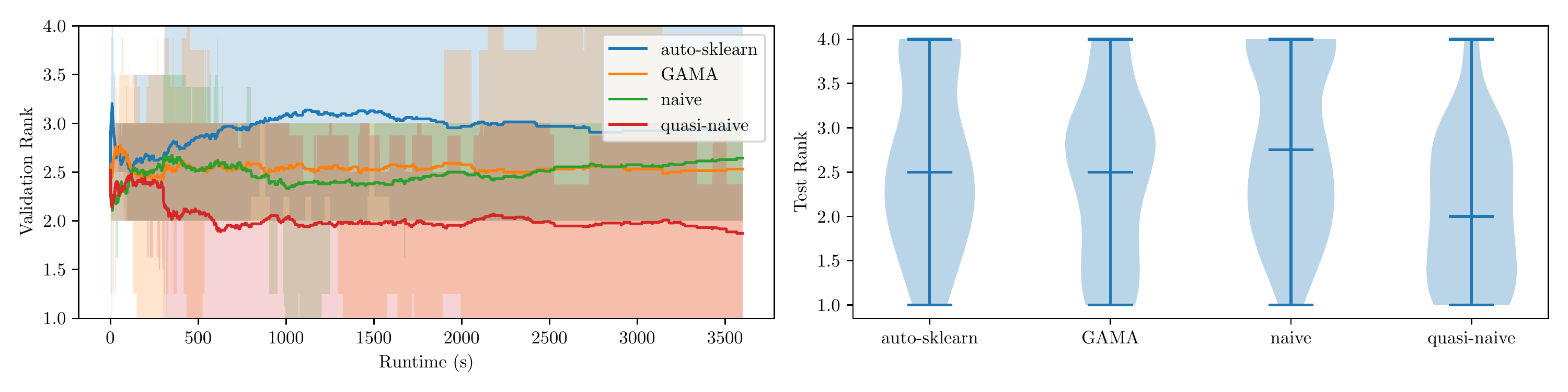}
    \caption{Validation ranks (left) and test ranks (right) for 1h timeouts.}
    \label{fig:results:ranks:1h}
\end{figure}

Fig. \ref{fig:results:ranks:1h} summarizes the results on a very abstract level.
In these plots, both figures show performance \emph{ranks}.
The left plot shows for each point of time $t$, the rank obtained by an approach when using the \emph{validation} performance of the best-seen solution up to $t$.
That is, it shows the internally best 5-CV result observed for any candidate pipeline up to that time.
The lines indicate median ranks, and the shaded areas show the rank IQRs.
Since we compare optimizers and are not primarily interested in test performance, we focus on validation performances.
However, to complement the internal validation results, the right plots show the rankings of the different optimizers with respect to the performance obtained by the finally returned model on the \emph{test} data.
The vertical bars in the violin plots are the respective medians.

The plot shows that the naive approaches are competitive or even stronger than state-of-the-art tools in the short run.
\naml is competitive with \gama both of which outperform auto-sklearn's SMAC in this time horizon.
\snaml even substantially detaches from that group and maintains a clear advantage over the whole time streak.
This advantage is also preserved on the test set, yielding the best test set performance among all approaches.
In at least 50\% of the cases, \snaml ranks best or is the runner-up.

\begin{figure}[t]
    \centering
    \includegraphics[width=\textwidth]{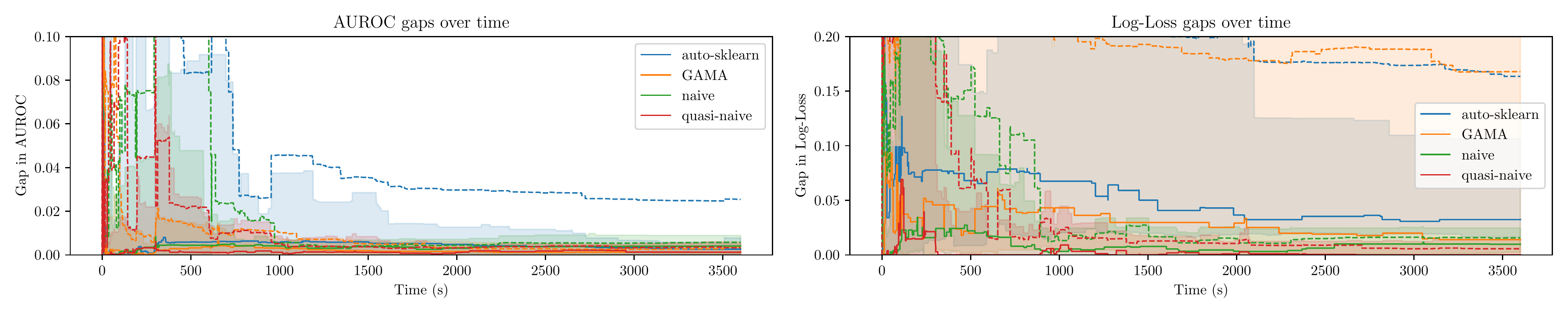}
    \caption{Empirical gaps on the \emph{validation} performance for binary (left) and multi-class (right) classification on 1h timeouts.}
    \label{fig:results:gaps:1h}
\end{figure}

We now discuss a more quantitative metric based on the \emph{empirical gap}.
This metric considers, for each point of time $t$, the best performance observed for \emph{any} candidate up to time $t$ and then computes for each optimizer the gap between its best-found solution and that reference score.
These empirical gaps over time are shown in Fig. \ref{fig:results:gaps:1h}.
Since the empirical gap metric requires comparable scales, it can only be averaged over instances of identical problem types (and hence identical base metrics).
Therefore, we provide it once for binary classification and based on AUROC (left) and once for the multi-class classification datasets based on log-loss (right).
At each point of time, the observations for all datasets corresponding to the respective type (binary or multi-class) are aggregated:
The solid lines are median gaps, the shaded areas are the interquartile ranges (IQR), and the dashed lines are 10\%-trimmed mean gaps.

First, we can see that for both performance measures AUROC and log-loss, the \emph{median} gaps are tiny.
Our interpretation of this is that for at least 50\% of the datasets, the different optimization approaches perform more or less equally.
In terms of log-loss, \autosklearn is slightly worse than the other approaches by a margin of approximately 0.05.

In general, we argue that gaps in log-loss below 0.1 are somewhat negligible.
If the difference in log-loss between two models is below 0.1 this means that the \emph{ratio} of probabilities assigned to the correct class is, on average, around 1.1.
For a three-class problem, this means that, even for situations of rather high uncertainty, if the better model assigns 55\% probability to the correct class, the weaker model also still assigns at least 51\% probability to the correct class and will hence choose it.
Now, this degree of irrelevance increases with a higher certainty of the better model or with higher numbers of classes.
In other words, in concrete situations where the two or three classes with the highest probability are at par, small differences in log-loss will not necessarily but often result in identical behaviors of the models.

As a second observation, we can see that the trimmed \emph{mean} statistics \emph{does} show a substantial difference between the approaches.
This holds specifically for \autosklearn and for \gama in case of log-loss performance.
The fact that those curves are consistently above the IQR area indicates that there are some datasets on which a substantial gap can be observed; the concrete performance curves per dataset in Sec.~\ref{appendix:performancecurves} of the Appendix confirm this observation.

\begin{figure}[t]
    \centering
    \includegraphics[width=\textwidth]{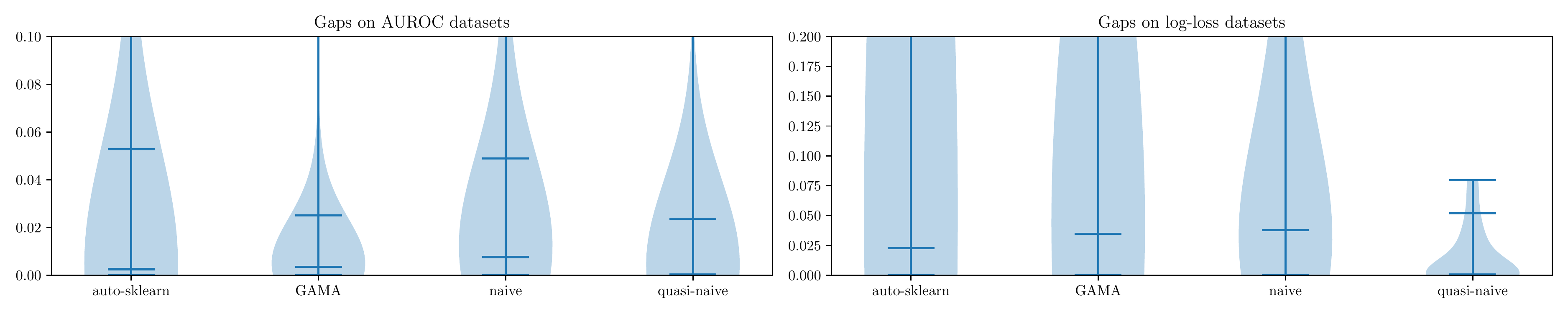}
    \caption{Empirical gaps on the \emph{test} performance for binary (left) and multi-class (right) classification on 1h timeouts.}
    \label{fig:results:test-gaps:1h}
\end{figure}

To complete this analysis, we also look at the test performance gaps of the different approaches.
These are summarized in Fig.~\ref{fig:results:test-gaps:1h}.
For each dataset, the best test score among the four models returned by the optimizers is computed, and then the gap of each approach is the test performance of its model minus the best score (in the case of AUROC, for log-loss, this difference is inverted).

We can observe that \snaml plays a reasonably dominant role in this comparison.
On both problem classes, i.e., binary classification and multi-class classification, it has a median gap of 0, indicating that it sets the best among all found solutions in at least 50\% of the cases.
While the advantage of \snaml in the multi-class scenario is quite pronounced, its tail behavior in the case of AUROC does not seem as strong as the one of \gama.
However, there are only two datasets on which the gap in AUROC is above 0.03.

Putting everything together, our assessment is that the naive approaches indeed compete with or even outperform the other approaches in the short run.
Neither \autosklearn nor \gama can substantially outperform the strict \naml approach in terms of validation performance; they achieve, however, a slightly better test performance in some cases.
Overall, \naml is competitive with \autosklearn and \gama and even outperforms one of them on either binary or multi class classification.
For \snaml we observe the same, but the advantage is much more pronounced.
In the short run, \snaml seems to be the, by far, best and stable choice among the four optimizers.

\subsubsection{RQ 2: How \emph{often} do global optimizers outperform the naive approaches in the long run and how \emph{long} do they need to take the lead?}

\begin{figure}[t]
    \centering
    \includegraphics[width=\textwidth]{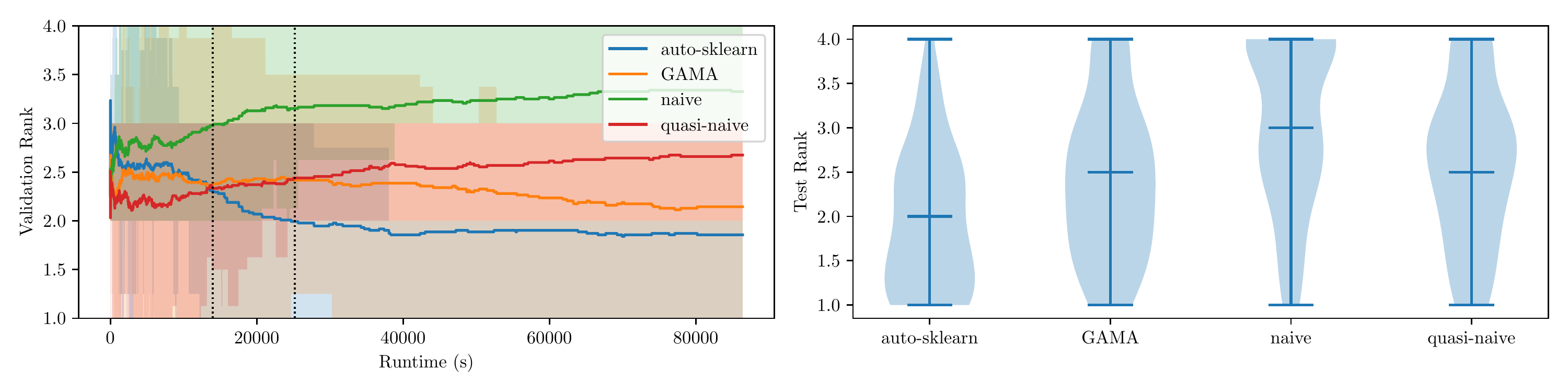}
    \caption{Validation ranks (left) and test ranks (right) for 24h timeouts.}
    \label{fig:results:ranks:1d}
\end{figure}

To get a first idea about the behavior of the optimizers in the long run, we again consider the average gap plots for the timeout of 24h in Fig. \ref{fig:results:ranks:1d}.
As expected, we can observe that, over time, the more sophisticated optimizers gain an advantage over the naive approaches.
We added a vertical dotted black line at the respective points of time where \autosklearn and \gama start to rank better than \snaml.
For \autosklearn, this point sets in after approximately 4h of runtime, and for \gama after 7h of runtime.
However, even though \automl takes and keeps the lead after 4h, it barely improves over a rank of 2, which means that it is on average on par with the set of other optimizers.

The ranking observed on the internal validation performance can also roughly be observed on the test performance ranks.
In general, \autosklearn produces the best or second-best test performance in 50\% of the cases whereas
\gama and \snaml have a slightly worse test rank performance.
Among these two, both have the same median rank, but \gama scores slightly better under the q1-quantile.
\naml is outperformed in terms of ranks in this time horizon.

\begin{figure}[t]
    \centering
    \includegraphics[width=\textwidth]{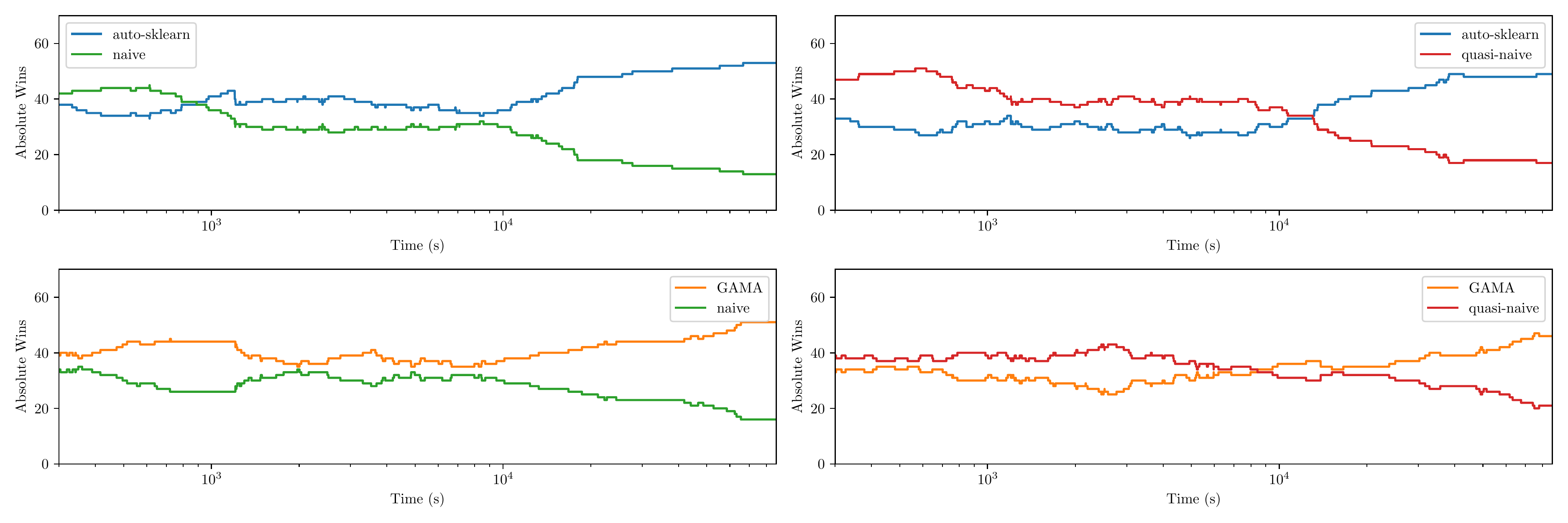}
    \caption{Validation performance duels: Count of datasets on which each of the dueling optimizers obtained the better validation score up to some point of time (x-axis on log-scale).}
    \label{fig:results:wins:1d}
\end{figure}

In order to get a slightly better understanding of the points of time when \naml and \snaml start to be outperformed on which numbers of datasets, Fig. \ref{fig:results:wins:1d} plots the numbers of wins in the duels between the naive approaches on one side and \autosklearn and \gama on the other side over time.
Each of these plots contains two lines, one corresponding to each of the dueling optimizers.
The left plots show the duels between \naml and \autosklearn and \gama, and the right plots show the respective duels of \snaml against \autosklearn and \gama.
On the x-axis, we show the runtime on a log scale.
On the y-axis, we count the number of datasets on which the respective optimizer has the lead (best-observed validation performance up to that point of time).

Looking at the left side of the plots, one observes that \naml is clearly inferior to both \autosklearn and \gama.
\naml is on par with \autosklearn and \gama in the beginning, but the number of datasets on which the others improve over \naml steadily increases after 3 hours.


Assessing the plots for \snaml on the right, we see a similar but less extreme picture.
For the first three hours, the algorithms are fairly balanced with light but consistent advantages of \snaml over both \autosklearn and \gama.
From this viewpoint, \snaml is clearly performing better than \naml in the short run.
Likewise, the advantage of \autosklearn and \gama after the first three hours is much less pronounced than in the case of \naml.
In fact, until the end, \snaml keeps being the \emph{winner} on 30\% of the datasets.

To summarize, we can answer the second research question as follows.
We observe that both \autosklearn and \gama manage to outperform \naml in 75\% of the time in the long run.
While \gama is constantly the winner on a higher number of datasets when dueling with \naml, \autosklearn has slight disadvantages in the first 20 minutes but then takes the lead.
With respect to \snaml, we observe that \autosklearn wins in 70\% and \gama in 65\% of the cases in the long run and it takes them roughly 3 hours to take the lead in this aggregated view.
Needless to say that this is a very condensed view, so we refer to detailed plots \emph{per dataset} over time in the appendix.

The discussion around RQ2 has been entirely qualitative.
We only looked at \emph{orderings} of optimizers but not at absolute performance.
While ranks are exactly what is needed to answer the binary question of \emph{whether} an optimizer outperforms another one, we are clearly also interested in the extent by which the better optimizers outperform the others.
After all \autosklearn seems to make a pretty strong case in the evaluation above, but how \emph{substantial} are those advantages?
    
\subsubsection{RQ 3: How \emph{large} is the performance gap between the solutions found by naive approaches compared to the best ones found by global optimizers?}

\begin{figure}[t]
    \centering
    \includegraphics[width=\textwidth]{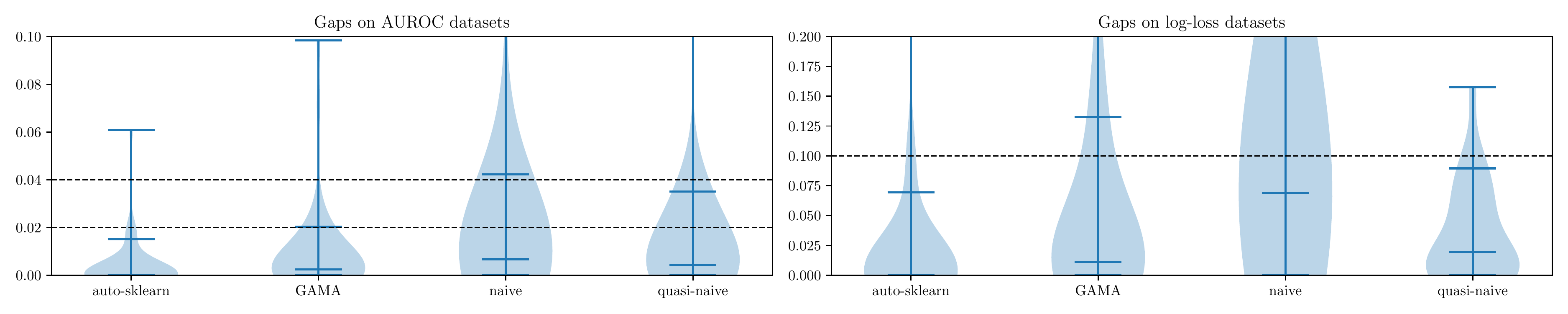}
    \caption{Empirical gaps on the \emph{test} performance for binary (left) and multi-class (right) classification on 24h timeouts.
    Whiskers show the medians, 90\% quantiles, and the maxima.}
    \label{fig:results:gaps:1d}
\end{figure}

To answer this question, we just look at the performance of the final solution on each problem produced by each optimizer.
For each dataset, we take the \emph{median} performance of each algorithm and identify the best among them.
The \emph{gap} of an algorithm is the difference between its own score and the best one.
For the two different problem classes, i.e., binary and multi-class classification, the gaps are summarized in Fig. \ref{fig:results:gaps:1d}.
Whiskers show median, 90\% quantile, and the maximum observation respectively.

These plots now clearly relativize the apparent dominance of \autosklearn suggested in the rank plots.
Looking first on the left plot for the AUROC in binary classification, we can see that \emph{all} of the optimizers have a close-to-zero median; that is, each of the optimizers is, on 50\% of the datasets, performing competitive to the optimal one.
Both \autosklearn and \gama are in 90\% of the cases less than 0.02 away from the best performance (lower auxilliary line).
However, \snaml is also competitive up to 0.02 in 85\% of the cases and to 0.04 in 95\% of the cases (worse only on one dataset; upper auxilliary line).
So while the advantages of \autosklearn and \gama are quantitatively measurable, they are indeed fairly small in the great majority of the cases.
Looking now at the right hand side, the situation is even more balanced on the benchmarks for multi-class classification.
In fact, \autosklearn and \snaml have a comparable 90\% quantile, which is below 0.1 (auxilliary line).
As discussed already for RQ1, we consider differences of less than 0.1 rather negligible.
Put differently, in over 90\% of the cases, both \autosklearn and \snaml exhibit essentially optimal performance on multi-class classification datasets after 24h.
The performance of \gama is not substantially worse though, since also here 80\% of the runs are at most 0.1 worse than the best solution, which can still be regarded considerably good.
In fact, even the performance of \naml is not \emph{too} bad in that at least in 60\% of the cases the performance gap is below 0.1.
However, there is also a good number of cases in which the gap of \naml \emph{is} substantial.

This being said, we answer the research question as follows.
\autosklearn, as the algorithm that shows the best performance on \emph{most} datasets after 24h, exhibits virtually \emph{no} performance advantage over any of the naive approaches on 50\% of the datasets for both binary and multi-class classification.
While it is able to significantly outperform \naml in the long run on \emph{some} datasets, it rarely ever outperforms \snaml.
In the case of multi-class classification benchmarks, \snaml is almost fully on par with \autosklearn, and in binary classification there are 5 datasets on which the performance gap of \snaml is bigger than 0.02 while being worse than 0.04 only once.
On binary classification, the same comparison holds for \gama against \snaml, while on multi-class classification \snaml even performs superior to \gama.

\subsection{Discussion}
\label{sec:results:discussion}
Putting all the results together, the naive approach seems to make a maybe unexpectedly strong case against established optimizers for standard classification problems.
Even the fully naive approach is competitive in the long run in 50\% of the cases.
A possible reason for this could also be that some datasets are ``too easy'' to optimize over, but there is no specific reason to believe that real world datasets are necessarily harder in this sense.
When applying the quasi-naive assumption, we obtain an optimizer that is hardly ever significantly outperformed neither by \autosklearn nor by \gama.
Both \automl and \gama manage to gain slight \emph{qualitative} advantages over \snaml as runtime increases, but the associated \emph{quantitative} advantages are negligible most of the time.

In our view, these are thrilling results as they propose an entirely new way of thinking about the optimization process in \automl.
Until now, pipeline optimization has almost always been treated as a complete \blackbox.
One strength but also weakness of the \blackbox approaches is that they do not require but also not efficiently support domain knowledge.
That is, the knowledge about machine learning is either encoded into the problem or wrapped around the optimizers in pre-processing (e.g., warm-starting) or post-processing (e.g., ensembling).
However, little knowledge can be considered \emph{within} the optimization approach, and the naive approach is different in this regard.
In contrast to black box optimizers, it suggests that the optimization process can be realized \emph{sequentially}.
The ability of sequential optimization opens the door to optimization \emph{flows}, which in turn give room for specialized components within the optimization process \cite{mohr2021replacing}.
For example, based on the observations in the optimization of one slot, it would be possible to activate or deactivate certain optimization modules in the subsequent optimization workflow.
Since this paper has shown even \snaml to be competitive, there is some reason to believe that such more sophisticated approaches might even superior to \blackbox optimization.

Besides potential superiority in terms of absolute performance, sequential optimization frameworks come with a couple of qualitative advantages.
As recent studies have shown, \blackbox approaches have, besides their difficulties of including knowledge into the optimization process, some substantial drawbacks that hinder their application in practice such as flexibility, lack of interactivity, and understandability \cite{drozdal2020trust,wang2019human}.
There are many techniques in the spectrum between \snaml and full \blackbox optimization that are still to be explored and that have the potential to combine high performance while satisfying important additional ``soft'' requirements mentioned above.
The competitive results of \snaml clearly motivate research in this direction.

\section{Conclusion}
In this paper, we have presented two \emph{naive} approaches for the optimization of machine learning pipelines.
Contrary to previous works, these approaches fully (\naml) or largely (\snaml) ignore the general assumption of dependencies between the choices of algorithms within a pipeline.
Furthermore, algorithm selection and hyperparameter optimization are decoupled by first selecting the algorithms of a pipeline only considering their default parametrizations.
Only when the algorithms are fixed, their hyperparameters are optimized.

The results show that the naive approaches are much more competitive than one would maybe expect.
For shorter timeouts of 1h, both naive algorithms performs highly competitive to optimization algorithms of state-of-the-art AutoML tools and sometimes (\snaml in fact even consistently) superior.
In the long run, 24h experiments shows that \snaml is largely on par with \autosklearn and \gama in terms of gaps to the best solution.

The naive approach is a door-opener for sequential optimization flows and hence naturally motivates a series of future work.
Besides the canonical extensions towards more complex pipeline structures such as the shape of a tree or a directed acyclic graph, it seems imperative to further explore the potential of a less naive approach as suggested in \cite{mohr2021replacing}, which adopts a stage-based optimization scheme.
Another interesting direction is to create a more interactive version of \naml in which the expert obtains visual summaries of what choices have been made and with the option for the expert to intervene, e.g., by revising some of the choices.
This could lead to an approach considering different optimization \emph{rounds} for different slots.

\medskip
\noindent
\textbf{Acknowledgements:}
We thank Matthias Feurer and Pieter Gijsbers for their remarkable support in adjusting \autosklearn and \gama for our evaluations.

\section*{Declarations}
Funding: This work was supported by the German Research Foundation (DFG) within the Collaborative Research Center ``On-The-Fly Computing'' (SFB 901)\\
Conflicts of interest/Competing interests: Eyke Hüllermeier\\
Availability of data and material: https://github.com/fmohr/naiveautoml/tree/mlj2021\\
Code availability: https://github.com/fmohr/naiveautoml/tree/mlj2021\\
Authors' contributions: Felix Mohr is the main author of both paper and implementation. Marcel Wever contributed in the manuscript revision as well as the resolution of technical aspects of the evaluation.\\
Ethics approval: not applicable\\
Consent to participate: not applicable\\
Consent for publication: not applicable

\bibliographystyle{spmpsci.bst} 
\bibliography{bibliography}

\newpage
\appendix
\section*{Appendix}
\section{Datasets}
All datasets are available via the openml.org platform \cite{OpenML2013}.
\begin{table}[h!]
    \centering
    \resizebox{\textwidth}{!}{
        \begin{tabular}{rlrrrrrrrrrr}
\toprule
 openmlid &                  name &  instances &  features &  numeric features &  classes & min \% & maj \% & \% missing & \% [0,1] & \% $\mu = 0$ & \% $\sigma = 1$ \\
\midrule
        3 &              kr-vs-kp &       3196 &        36 &                 0 &        2 &   47\% &   52\% &        0\% &      n/a &          n/a &             n/a \\
       12 &         mfeat-factors &       2000 &       216 &               216 &       10 &   10\% &   10\% &        0\% &      n/a &          n/a &             n/a \\
       23 &                   cmc &       1473 &         9 &                 2 &        3 &   22\% &   42\% &        0\% &      n/a &          n/a &             n/a \\
       31 &              credit-g &       1000 &        20 &                 7 &        2 &   30\% &   70\% &        0\% &      0\% &          0\% &             0\% \\
       54 &               vehicle &        846 &        18 &                18 &        4 &   23\% &   25\% &        0\% &      0\% &          0\% &             0\% \\
      181 &                 yeast &       1484 &         8 &                 8 &       10 &    0\% &   31\% &        0\% &     25\% &          0\% &             0\% \\
      188 &            eucalyptus &        736 &        19 &                14 &        5 &   14\% &   29\% &        3\% &      0\% &          0\% &             0\% \\
     1049 &                   pc4 &       1458 &        37 &                37 &        2 &   12\% &   87\% &        0\% &      3\% &          0\% &             0\% \\
     1067 &                   kc1 &       2109 &        21 &                21 &        2 &   15\% &   84\% &        0\% &      0\% &          0\% &             0\% \\
     1111 &    KDDCup09\_appetency &      50000 &       230 &               192 &        2 &    1\% &   98\% &       69\% &      0\% &          0\% &             0\% \\
     1457 &  amazon-commerce-revi &       1500 &     10000 &             10000 &       50 &    2\% &    2\% &        0\% &     22\% &          0\% &             0\% \\
     1461 &        bank-marketing &      45211 &        16 &                 7 &        2 &   11\% &   88\% &        0\% &      0\% &          0\% &             0\% \\
     1464 &  blood-transfusion-se &        748 &         4 &                 4 &        2 &   23\% &   76\% &        0\% &      0\% &          0\% &             0\% \\
     1468 &                cnae-9 &       1080 &       856 &               856 &        9 &   11\% &   11\% &        0\% &     88\% &          0\% &             0\% \\
     1475 &  first-order-theorem- &       6118 &        51 &                51 &        6 &    7\% &   41\% &        0\% &      0\% &          4\% &             2\% \\
     1485 &               madelon &       2600 &       500 &               500 &        2 &   50\% &   50\% &        0\% &      0\% &          0\% &             0\% \\
     1486 &                 nomao &      34465 &       118 &                89 &        2 &   28\% &   71\% &        0\% &     83\% &          0\% &             0\% \\
     1487 &       ozone-level-8hr &       2534 &        72 &                72 &        2 &    6\% &   93\% &        0\% &      0\% &          0\% &             0\% \\
     1489 &               phoneme &       5404 &         5 &                 5 &        2 &   29\% &   70\% &        0\% &      0\% &        100\% &           100\% \\
     1494 &           qsar-biodeg &       1055 &        41 &                41 &        2 &   33\% &   66\% &        0\% &      7\% &          0\% &             0\% \\
     1515 &            micro-mass &        571 &      1300 &              1300 &       20 &    1\% &   10\% &        0\% &      0\% &         17\% &             0\% \\
     1590 &                 adult &      48842 &        14 &                 6 &        2 &   23\% &   76\% &        0\% &      0\% &          0\% &             0\% \\
     4134 &           Bioresponse &       3751 &      1776 &              1776 &        2 &   45\% &   54\% &        0\% &     81\% &          1\% &             0\% \\
     4135 &  Amazon\_employee\_acce &      32769 &         9 &                 0 &        2 &    5\% &   94\% &        0\% &      n/a &          n/a &             n/a \\
     4534 &      PhishingWebsites &      11055 &        30 &                 0 &        2 &   44\% &   55\% &        0\% &      n/a &          n/a &             n/a \\
     4538 &  GesturePhaseSegmenta &       9873 &        32 &                32 &        5 &   10\% &   29\% &        0\% &      0\% &         75\% &             0\% \\
     4541 &         Diabetes130US &     101766 &        49 &                13 &        3 &   11\% &   53\% &        0\% &      0\% &          0\% &             0\% \\
    23512 &                 higgs &      98050 &        28 &                28 &        2 &   47\% &   52\% &        0\% &      0\% &          7\% &             0\% \\
    23517 &           numerai28.6 &      96320 &        21 &                21 &        2 &   49\% &   50\% &        0\% &    100\% &          0\% &             0\% \\
    40498 &    wine-quality-white &       4898 &        11 &                11 &        7 &    0\% &   44\% &        0\% &      0\% &          0\% &             0\% \\
    40668 &             connect-4 &      67557 &        42 &                 0 &        3 &    9\% &   65\% &        0\% &      n/a &          n/a &             n/a \\
    40670 &                   dna &       3186 &       180 &                 0 &        3 &   24\% &   51\% &        0\% &      n/a &          n/a &             n/a \\
    40685 &               shuttle &      58000 &         9 &                 9 &        7 &    0\% &   78\% &        0\% &      0\% &          0\% &             0\% \\
    40701 &                 churn &       5000 &        20 &                16 &        2 &   14\% &   85\% &        0\% &      0\% &          0\% &             0\% \\
    40900 &             Satellite &       5100 &        36 &                36 &        2 &    1\% &   98\% &        0\% &      0\% &          0\% &             0\% \\
    40975 &                   car &       1728 &         6 &                 0 &        4 &    3\% &   70\% &        0\% &      n/a &          n/a &             n/a \\
    40978 &  Internet-Advertiseme &       3279 &      1558 &                 3 &        2 &   13\% &   86\% &        0\% &      0\% &          0\% &             0\% \\
    40981 &            Australian &        690 &        14 &                 6 &        2 &   44\% &   55\% &        0\% &      0\% &          0\% &             0\% \\
    40982 &    steel-plates-fault &       1941 &        27 &                27 &        7 &    2\% &   34\% &        0\% &     11\% &          0\% &             0\% \\
    40983 &                  wilt &       4839 &         5 &                 5 &        2 &    5\% &   94\% &        0\% &      0\% &          0\% &             0\% \\
    40984 &               segment &       2310 &        19 &                19 &        7 &   14\% &   14\% &        0\% &      6\% &          0\% &             0\% \\
    40996 &         Fashion-MNIST &      70000 &       784 &               784 &       10 &   10\% &   10\% &        0\% &      0\% &          0\% &             0\% \\
    41027 &  jungle\_chess\_2pcs\_ra &      44819 &         6 &                 6 &        3 &    9\% &   51\% &        0\% &      0\% &          0\% &             0\% \\
    41138 &            APSFailure &      76000 &       170 &               170 &        2 &    1\% &   98\% &        8\% &      0\% &          0\% &             0\% \\
    41142 &             christine &       5418 &      1636 &              1599 &        2 &   50\% &   50\% &        0\% &      0\% &          0\% &             0\% \\
    41143 &               jasmine &       2984 &       144 &                 8 &        2 &   50\% &   50\% &        0\% &      0\% &          0\% &             0\% \\
    41144 &              madeline &       3140 &       259 &               259 &        2 &   49\% &   50\% &        0\% &      0\% &          0\% &             0\% \\
    41145 &            philippine &       5832 &       308 &               308 &        2 &   50\% &   50\% &        0\% &      0\% &          3\% &             0\% \\
    41146 &               sylvine &       5124 &        20 &                20 &        2 &   50\% &   50\% &        0\% &      0\% &          0\% &             0\% \\
    41147 &                albert &     425240 &        78 &                26 &        2 &   50\% &   50\% &        8\% &      0\% &          0\% &             0\% \\
    41150 &             MiniBooNE &     130064 &        50 &                50 &        2 &   28\% &   71\% &        0\% &      0\% &          0\% &             0\% \\
    41156 &                   ada &       4147 &        48 &                48 &        2 &   24\% &   75\% &        0\% &     83\% &          8\% &             0\% \\
    41157 &                arcene &        100 &     10000 &             10000 &        2 &   44\% &   56\% &        0\% &      0\% &          1\% &             0\% \\
    41158 &                  gina &       3153 &       970 &               970 &        2 &   49\% &   50\% &        0\% &      0\% &          0\% &             0\% \\
    41159 &             guillermo &      20000 &      4296 &              4296 &        2 &   40\% &   59\% &        0\% &      0\% &          0\% &             0\% \\
    41161 &              riccardo &      20000 &      4296 &              4296 &        2 &   25\% &   75\% &        0\% &      0\% &          0\% &             0\% \\
    41162 &                  kick &      72983 &        32 &                14 &        2 &   12\% &   87\% &        6\% &      0\% &          0\% &             0\% \\
    41163 &               dilbert &      10000 &      2000 &              2000 &        5 &   19\% &   20\% &        0\% &      0\% &          0\% &             0\% \\
    41164 &                fabert &       8237 &       800 &               800 &        7 &    6\% &   23\% &        0\% &     96\% &          4\% &             0\% \\
    41165 &                robert &      10000 &      7200 &              7200 &       10 &    9\% &   10\% &        0\% &      0\% &          0\% &             0\% \\
    41166 &               volkert &      58310 &       180 &               180 &       10 &    2\% &   21\% &        0\% &     15\% &         18\% &             0\% \\
    41167 &                dionis &     416188 &        60 &                60 &      355 &    0\% &    0\% &        0\% &      0\% &         10\% &             0\% \\
    41168 &                jannis &      83733 &        54 &                54 &        4 &    2\% &   46\% &        0\% &      4\% &          0\% &             0\% \\
    41169 &                helena &      65196 &        27 &                27 &      100 &    0\% &    6\% &        0\% &      4\% &          0\% &             0\% \\
    42732 &   sf-police-incidents &    2215023 &         9 &                 3 &        2 &   12\% &   87\% &        0\% &      0\% &          0\% &             0\% \\
    42733 &  Click\_prediction\_sma &      39948 &        11 &                 5 &        2 &   16\% &   83\% &        0\% &      0\% &          0\% &             0\% \\
\bottomrule
\end{tabular}

    }
    \caption{Overview of datasets used in the evaluation.}
    \label{tab:datasets}
\end{table}

\newpage
\section{Considered Algorithms}
The following algorithms from the scikit-learn library were considered for the three pipeline slots (same setup for all optimizers).
Please refer to \url{https://github.com/fmohr/naiveautoml/tree/mlj2021} for the exact specification of the search space including the hyper-parameter spaces.

\textbf{Data-Pre-Processors}
\begin{itemize}
    \item Normalizer
    \item VarianceThreshold
    \item QuantileTransformer
    \item StandardScaler
    \item MinMaxScaler
    \item PowerTransformer
    \item RobustScaler
\end{itemize}

\textbf{Feature-Pre-Processors}
\begin{itemize}
    \item FeatureAgglomeration
    \item PCA
    \item PolynomialFeatures
    \item Nystroem
    \item SelectPercentile
    \item KernelPCA
    \item GenericUnivariateSelect
    \item RBFSampler
    \item FastICA
\end{itemize}
	 
\textbf{Classifiers}
\begin{itemize}
    \item SVC (once for each out of four kernels)
	\item KNeighborsClassifier
	\item QuadraticDiscriminantAnalysis
	\item RandomForestClassifier
	\item MultinomialNB
	\item LinearDiscriminantAnalysis
	\item ExtraTreesClassifier
	\item BernoulliNB
	\item MLPClassifier
	\item GradientBoostingClassifier
	\item GaussianNB
	\item DecisionTreeClassifier
\end{itemize}


\clearpage
\newpage
\section{Final Result Tables}
The following tables show the \emph{mean} test score results of the approaches on the different datasets together with the standard deviation.
Best performances are in bold, and entries that are not at least 0.01 (AUROC) or 0.1 worse (log-loss) than the best one or not statistically significantly different (according to a Wilcoxon signed rank test with p=0.05) are underlined.

\subsection{Binary Classification Datasets}
\begin{table}[h!]
    \centering
    \begin{tabular}{r||rrrr}
\toprule
 \multicolumn{1}{c}{id} & \multicolumn{1}{c}{auto-sklearn} &   \multicolumn{1}{c}{GAMA} &  \multicolumn{1}{c}{naive} & \multicolumn{1}{c}{quasi-naive} \\
\midrule
                      3 &             \textbf{1.0$\pm$0.0} &       \textbf{1.0$\pm$0.0} &       \textbf{1.0$\pm$0.0} &            \textbf{1.0$\pm$0.0} \\
                     31 &           \textbf{0.78$\pm$0.04} &     \textbf{0.78$\pm$0.05} &              0.75$\pm$0.09 &       \underline{0.77$\pm$0.05} \\
                   1049 &        \underline{0.94$\pm$0.02} &     \textbf{0.95$\pm$0.02} &              0.91$\pm$0.04 &          \textbf{0.95$\pm$0.02} \\
                   1067 &        \underline{0.83$\pm$0.02} &  \underline{0.83$\pm$0.03} &     \textbf{0.84$\pm$0.03} &          \textbf{0.84$\pm$0.02} \\
                   1111 &        \underline{0.68$\pm$0.03} &              0.67$\pm$0.06 &              0.63$\pm$0.02 &          \textbf{0.69$\pm$0.03} \\
                   1461 &           \textbf{0.78$\pm$0.01} &     \textbf{0.78$\pm$0.01} &     \textbf{0.78$\pm$0.01} &          \textbf{0.78$\pm$0.01} \\
                   1464 &        \underline{0.73$\pm$0.06} &  \underline{0.73$\pm$0.05} &     \textbf{0.74$\pm$0.04} &          \textbf{0.74$\pm$0.06} \\
                   1485 &        \underline{0.93$\pm$0.01} &     \textbf{0.94$\pm$0.13} &  \underline{0.92$\pm$0.01} &       \underline{0.91$\pm$0.03} \\
                   1486 &            \textbf{0.98$\pm$0.0} &      \textbf{0.98$\pm$0.0} &      \textbf{0.98$\pm$0.0} &           \textbf{0.98$\pm$0.0} \\
                   1487 &           \textbf{0.93$\pm$0.04} &              0.91$\pm$0.09 &     \textbf{0.93$\pm$0.05} &       \underline{0.92$\pm$0.06} \\
                   1489 &           \textbf{0.97$\pm$0.01} &  \underline{0.96$\pm$0.01} &     \textbf{0.97$\pm$0.01} &          \textbf{0.97$\pm$0.04} \\
                   1494 &           \textbf{0.93$\pm$0.03} &     \textbf{0.93$\pm$0.03} &              0.91$\pm$0.03 &          \textbf{0.93$\pm$0.03} \\
                   1590 &            \textbf{0.89$\pm$0.0} &     \textbf{0.89$\pm$0.01} &      \textbf{0.89$\pm$0.0} &           \textbf{0.89$\pm$0.0} \\
                   4134 &           \textbf{0.89$\pm$0.01} &  \underline{0.88$\pm$0.01} &  \underline{0.88$\pm$0.01} &       \underline{0.88$\pm$0.03} \\
                   4135 &        \underline{0.83$\pm$0.02} &     \textbf{0.84$\pm$0.02} &              0.66$\pm$0.17 &          \textbf{0.84$\pm$0.03} \\
                   4534 &             \textbf{1.0$\pm$0.0} &       \textbf{1.0$\pm$0.0} &       \textbf{1.0$\pm$0.0} &            \textbf{1.0$\pm$0.0} \\
                  23512 &        \underline{0.79$\pm$0.01} &      \textbf{0.8$\pm$0.04} &  \underline{0.79$\pm$0.04} &       \underline{0.79$\pm$0.01} \\
                  23517 &           \textbf{0.53$\pm$0.01} &     \textbf{0.53$\pm$0.01} &     \textbf{0.53$\pm$0.01} &          \textbf{0.53$\pm$0.01} \\
                  40701 &        \underline{0.91$\pm$0.03} &   \underline{0.9$\pm$0.08} &               0.9$\pm$0.02 &          \textbf{0.92$\pm$0.02} \\
                  40900 &           \textbf{0.99$\pm$0.01} &     \textbf{0.99$\pm$0.01} &               0.9$\pm$0.12 &          \textbf{0.99$\pm$0.15} \\
                  40978 &           \textbf{0.98$\pm$0.01} &  \underline{0.97$\pm$0.02} &  \underline{0.97$\pm$0.03} &          \textbf{0.98$\pm$0.02} \\
                  40981 &           \textbf{0.95$\pm$0.03} &  \underline{0.94$\pm$0.03} &              0.93$\pm$0.04 &          \textbf{0.95$\pm$0.03} \\
                  40983 &             \textbf{1.0$\pm$0.0} &       \textbf{1.0$\pm$0.0} &              0.61$\pm$0.31 &                   0.68$\pm$0.26 \\
                  41138 &           \textbf{0.99$\pm$0.01} &      \textbf{0.99$\pm$0.0} &     \textbf{0.99$\pm$0.01} &          \textbf{0.99$\pm$0.01} \\
                  41142 &           \textbf{0.82$\pm$0.03} &  \underline{0.81$\pm$0.02} &  \underline{0.81$\pm$0.02} &          \textbf{0.82$\pm$0.02} \\
                  41143 &        \underline{0.88$\pm$0.02} &     \textbf{0.89$\pm$0.02} &              0.87$\pm$0.02 &       \underline{0.88$\pm$0.02} \\
                  41144 &           \textbf{0.95$\pm$0.02} &  \underline{0.94$\pm$0.01} &  \underline{0.94$\pm$0.02} &                    0.82$\pm$0.2 \\
                  41145 &            \textbf{0.9$\pm$0.02} &              0.86$\pm$0.07 &              0.79$\pm$0.03 &                   0.84$\pm$0.09 \\
                  41146 &            \textbf{0.99$\pm$0.0} &      \textbf{0.99$\pm$0.0} &      \textbf{0.99$\pm$0.0} &                   0.89$\pm$0.19 \\
                  41147 &                    0.64$\pm$0.05 &  \underline{0.72$\pm$0.04} &      \textbf{0.73$\pm$0.0} &           \textbf{0.73$\pm$0.0} \\
                  41150 &         \underline{0.97$\pm$0.0} &     \textbf{0.98$\pm$0.02} &      \textbf{0.98$\pm$0.0} &           \textbf{0.98$\pm$0.0} \\
                  41156 &           \textbf{0.91$\pm$0.02} &     \textbf{0.91$\pm$0.02} &     \textbf{0.91$\pm$0.02} &          \textbf{0.91$\pm$0.01} \\
                  41157 &                    0.87$\pm$0.11 &              0.86$\pm$0.12 &              0.89$\pm$0.13 &          \textbf{0.95$\pm$0.06} \\
                  41158 &           \textbf{0.99$\pm$0.01} &      \textbf{0.99$\pm$0.0} &      \textbf{0.99$\pm$0.0} &           \textbf{0.99$\pm$0.0} \\
                  41159 &                      0.5$\pm$0.0 &              0.71$\pm$0.12 &  \underline{0.88$\pm$0.01} &          \textbf{0.89$\pm$0.01} \\
                  41161 &                      0.5$\pm$0.0 &              0.93$\pm$0.09 &       \textbf{1.0$\pm$0.0} &            \textbf{1.0$\pm$0.0} \\
                  41162 &           \textbf{0.74$\pm$0.01} &              0.72$\pm$0.02 &  \underline{0.73$\pm$0.01} &       \underline{0.73$\pm$0.01} \\
                  42732 &                      0.5$\pm$0.0 &  \underline{0.62$\pm$0.05} &      \textbf{0.64$\pm$0.0} &           \textbf{0.64$\pm$0.0} \\
                  42733 &                    0.62$\pm$0.01 &              0.62$\pm$0.01 &              0.65$\pm$0.05 &          \textbf{0.72$\pm$0.01} \\
\bottomrule
\end{tabular}
    \caption{Avg. test AUROC on binary classification datasets (1h timeout).}
    \label{tab:results:auc:1h}
\end{table}

\begin{table}[h!]
    \centering
    \begin{tabular}{r||rrrr}
\toprule
 \multicolumn{1}{c}{id} & \multicolumn{1}{c}{auto-sklearn} &   \multicolumn{1}{c}{GAMA} &  \multicolumn{1}{c}{naive} & \multicolumn{1}{c}{quasi-naive} \\
\midrule
                      3 &             \textbf{1.0$\pm$0.0} &       \textbf{1.0$\pm$0.0} &      \textbf{1.0$\pm$0.03} &            \textbf{1.0$\pm$0.0} \\
                     31 &        \underline{0.77$\pm$0.05} &     \textbf{0.78$\pm$0.05} &  \underline{0.77$\pm$0.05} &       \underline{0.77$\pm$0.05} \\
                   1049 &           \textbf{0.95$\pm$0.02} &     \textbf{0.95$\pm$0.02} &  \underline{0.92$\pm$0.05} &       \underline{0.94$\pm$0.03} \\
                   1067 &           \textbf{0.85$\pm$0.02} &              0.83$\pm$0.03 &   \underline{0.8$\pm$0.04} &                   0.81$\pm$0.04 \\
                   1111 &           \textbf{0.77$\pm$0.02} &               0.7$\pm$0.04 &              0.71$\pm$0.03 &                   0.74$\pm$0.02 \\
                   1461 &           \textbf{0.78$\pm$0.01} &     \textbf{0.78$\pm$0.01} &     \textbf{0.78$\pm$0.01} &          \textbf{0.78$\pm$0.01} \\
                   1464 &        \underline{0.72$\pm$0.06} &     \textbf{0.73$\pm$0.04} &     \textbf{0.73$\pm$0.06} &          \textbf{0.73$\pm$0.05} \\
                   1485 &           \textbf{0.95$\pm$0.01} &  \underline{0.94$\pm$0.01} &              0.93$\pm$0.01 &                   0.91$\pm$0.02 \\
                   1486 &            \textbf{0.99$\pm$0.0} &      \textbf{0.99$\pm$0.0} &      \textbf{0.99$\pm$0.0} &           \textbf{0.99$\pm$0.0} \\
                   1487 &           \textbf{0.93$\pm$0.04} &     \textbf{0.93$\pm$0.04} &  \underline{0.92$\pm$0.04} &       \underline{0.92$\pm$0.05} \\
                   1489 &           \textbf{0.97$\pm$0.01} &     \textbf{0.97$\pm$0.01} &     \textbf{0.97$\pm$0.01} &          \textbf{0.97$\pm$0.01} \\
                   1494 &        \underline{0.93$\pm$0.03} &     \textbf{0.94$\pm$0.02} &  \underline{0.92$\pm$0.02} &                   0.92$\pm$0.13 \\
                   1590 &            \textbf{0.89$\pm$0.0} &      \textbf{0.89$\pm$0.0} &      \textbf{0.89$\pm$0.0} &           \textbf{0.89$\pm$0.0} \\
                   4134 &           \textbf{0.89$\pm$0.01} &  \underline{0.88$\pm$0.01} &     \textbf{0.89$\pm$0.01} &          \textbf{0.89$\pm$0.03} \\
                   4135 &           \textbf{0.88$\pm$0.02} &              0.85$\pm$0.03 &     \textbf{0.88$\pm$0.02} &       \underline{0.87$\pm$0.02} \\
                   4534 &             \textbf{1.0$\pm$0.0} &       \textbf{1.0$\pm$0.0} &       \textbf{1.0$\pm$0.0} &            \textbf{1.0$\pm$0.0} \\
                  23512 &           \textbf{0.81$\pm$0.01} &      \textbf{0.81$\pm$0.0} &              0.79$\pm$0.01 &        \underline{0.8$\pm$0.01} \\
                  23517 &           \textbf{0.53$\pm$0.01} &     \textbf{0.53$\pm$0.01} &  \underline{0.52$\pm$0.01} &          \textbf{0.53$\pm$0.01} \\
                  40701 &           \textbf{0.92$\pm$0.03} &     \textbf{0.92$\pm$0.03} &  \underline{0.91$\pm$0.02} &          \textbf{0.92$\pm$0.09} \\
                  40900 &             \textbf{1.0$\pm$0.0} &      \textbf{1.0$\pm$0.01} &               0.91$\pm$0.1 &                   0.95$\pm$0.09 \\
                  40978 &           \textbf{0.98$\pm$0.02} &     \textbf{0.98$\pm$0.01} &  \underline{0.97$\pm$0.04} &          \textbf{0.98$\pm$0.01} \\
                  40981 &           \textbf{0.95$\pm$0.03} &     \textbf{0.95$\pm$0.03} &              0.93$\pm$0.04 &          \textbf{0.95$\pm$0.03} \\
                  40983 &             \textbf{1.0$\pm$0.0} &       \textbf{1.0$\pm$0.0} &              0.71$\pm$0.28 &                   0.75$\pm$0.21 \\
                  41138 &            \textbf{0.99$\pm$0.0} &      \textbf{0.99$\pm$0.0} &      \textbf{0.99$\pm$0.0} &           \textbf{0.99$\pm$0.0} \\
                  41142 &           \textbf{0.84$\pm$0.02} &  \underline{0.83$\pm$0.03} &  \underline{0.83$\pm$0.03} &                   0.82$\pm$0.07 \\
                  41143 &        \underline{0.88$\pm$0.02} &     \textbf{0.89$\pm$0.02} &              0.87$\pm$0.02 &       \underline{0.88$\pm$0.02} \\
                  41144 &           \textbf{0.97$\pm$0.01} &              0.95$\pm$0.01 &              0.95$\pm$0.01 &                   0.88$\pm$0.18 \\
                  41145 &           \textbf{0.93$\pm$0.01} &              0.91$\pm$0.01 &              0.79$\pm$0.02 &                    0.9$\pm$0.02 \\
                  41146 &            \textbf{0.99$\pm$0.0} &      \textbf{0.99$\pm$0.0} &      \textbf{0.99$\pm$0.0} &                   0.94$\pm$0.17 \\
                  41147 &            \textbf{0.77$\pm$0.0} &  \underline{0.76$\pm$0.02} &               0.73$\pm$0.0 &                    0.73$\pm$0.0 \\
                  41150 &            \textbf{0.99$\pm$0.0} &     \textbf{0.99$\pm$0.16} &               0.97$\pm$0.0 &                   0.92$\pm$0.16 \\
                  41156 &           \textbf{0.92$\pm$0.02} &  \underline{0.91$\pm$0.02} &  \underline{0.91$\pm$0.01} &       \underline{0.91$\pm$0.02} \\
                  41157 &                     0.9$\pm$0.08 &              0.84$\pm$0.12 &               0.87$\pm$0.1 &          \textbf{0.94$\pm$0.06} \\
                  41158 &            \textbf{0.99$\pm$0.0} &      \textbf{0.99$\pm$0.0} &      \textbf{0.99$\pm$0.0} &           \textbf{0.99$\pm$0.0} \\
                  41159 &           \textbf{0.92$\pm$0.01} &               0.9$\pm$0.02 &              0.89$\pm$0.01 &          \textbf{0.92$\pm$0.01} \\
                  41161 &             \textbf{1.0$\pm$0.0} &       \textbf{1.0$\pm$0.0} &              0.87$\pm$0.23 &            \textbf{1.0$\pm$0.0} \\
                  41162 &           \textbf{0.75$\pm$0.01} &  \underline{0.74$\pm$0.01} &              0.73$\pm$0.01 &                   0.73$\pm$0.01 \\
                  42732 &        \underline{0.64$\pm$0.01} &      \textbf{0.65$\pm$0.0} &   \underline{0.64$\pm$0.0} &           \textbf{0.65$\pm$0.0} \\
                  42733 &           \textbf{0.64$\pm$0.01} &  \underline{0.63$\pm$0.03} &              0.62$\pm$0.02 &       \underline{0.61$\pm$0.03} \\
\bottomrule
\end{tabular}

    \caption{Avg. test AUROC on binary classification datasets (1d timeout).}
    \label{tab:results:auc:1d}
\end{table}

\clearpage
\newpage
\subsection{Multi-Class Classification Datasets}
\begin{table}[h!]
    \centering
    \begin{tabular}{r||rrrr}
\toprule
 \multicolumn{1}{c}{id} & \multicolumn{1}{c}{auto-sklearn} &   \multicolumn{1}{c}{GAMA} &  \multicolumn{1}{c}{naive} & \multicolumn{1}{c}{quasi-naive} \\
\midrule
                     12 &           \textbf{0.13$\pm$0.06} &  \underline{0.17$\pm$7.05} &  \underline{1.54$\pm$5.17} &          \textbf{0.13$\pm$0.04} \\
                     23 &           \textbf{0.91$\pm$0.04} &     \textbf{0.91$\pm$0.04} &              2.02$\pm$3.08 &       \underline{0.94$\pm$0.22} \\
                     54 &        \underline{0.47$\pm$0.06} &  \underline{0.44$\pm$7.79} &              1.05$\pm$0.36 &          \textbf{0.42$\pm$0.05} \\
                    181 &           \textbf{1.05$\pm$0.08} &  \underline{1.13$\pm$0.19} &  \underline{1.09$\pm$0.19} &       \underline{1.07$\pm$0.08} \\
                    188 &           \textbf{1.15$\pm$0.07} &  \underline{1.16$\pm$0.07} &              1.28$\pm$0.09 &       \underline{1.21$\pm$0.07} \\
                   1457 &                    2.14$\pm$0.05 &              2.04$\pm$0.52 &                1.3$\pm$0.8 &          \textbf{0.75$\pm$0.15} \\
                   1468 &                    0.27$\pm$0.09 &   \underline{0.2$\pm$0.06} &  \underline{0.18$\pm$0.62} &          \textbf{0.13$\pm$0.04} \\
                   1475 &        \underline{1.12$\pm$0.05} &     \textbf{1.08$\pm$0.06} &  \underline{1.12$\pm$0.07} &       \underline{1.15$\pm$0.04} \\
                   1515 &         \underline{0.57$\pm$0.1} &   \underline{0.56$\pm$0.2} &      \textbf{0.5$\pm$0.08} &       \underline{0.53$\pm$0.09} \\
                   4538 &        \underline{0.92$\pm$0.04} &     \textbf{0.85$\pm$0.05} &  \underline{0.87$\pm$0.03} &       \underline{0.89$\pm$0.15} \\
                   4541 &        \underline{0.94$\pm$0.01} &  \underline{0.93$\pm$0.01} &   \underline{0.93$\pm$0.0} &           \textbf{0.92$\pm$0.0} \\
                  40498 &         \underline{0.8$\pm$0.03} &              1.17$\pm$0.58 &     \textbf{0.77$\pm$0.25} &        \underline{0.79$\pm$0.2} \\
                  40668 &                    0.65$\pm$0.12 &      \textbf{0.45$\pm$0.3} &  \underline{0.59$\pm$0.01} &       \underline{0.59$\pm$0.01} \\
                  40670 &        \underline{0.11$\pm$0.05} &  \underline{0.17$\pm$0.03} &  \underline{0.19$\pm$0.07} &           \textbf{0.1$\pm$0.02} \\
                  40685 &             \textbf{0.0$\pm$0.0} &       \textbf{0.0$\pm$0.0} &       \textbf{0.0$\pm$0.0} &            \textbf{0.0$\pm$0.0} \\
                  40975 &            \textbf{0.0$\pm$0.01} &       \textbf{0.0$\pm$0.0} &              0.24$\pm$0.22 &       \underline{0.02$\pm$0.21} \\
                  40982 &           \textbf{0.53$\pm$0.05} &               0.76$\pm$0.6 &              0.68$\pm$0.06 &       \underline{0.54$\pm$0.46} \\
                  40984 &        \underline{0.08$\pm$0.03} &  \underline{0.07$\pm$0.03} &  \underline{0.07$\pm$0.03} &          \textbf{0.06$\pm$0.03} \\
                  40996 &                      2.3$\pm$0.0 &              1.37$\pm$0.58 &  \underline{0.37$\pm$0.01} &          \textbf{0.36$\pm$0.01} \\
                  41027 &        \underline{0.23$\pm$0.06} &      \textbf{0.2$\pm$0.03} &               0.42$\pm$0.1 &       \underline{0.27$\pm$0.03} \\
                  41163 &                    0.22$\pm$0.06 &              0.17$\pm$0.23 &     \textbf{0.05$\pm$0.04} &          \textbf{0.05$\pm$0.01} \\
                  41164 &           \textbf{0.83$\pm$0.02} &              0.96$\pm$0.12 &  \underline{0.89$\pm$0.08} &       \underline{0.84$\pm$0.08} \\
                  41165 &                      2.3$\pm$0.0 &              2.26$\pm$2.37 &     \textbf{1.71$\pm$0.02} &       \underline{1.72$\pm$0.02} \\
                  41166 &        \underline{1.18$\pm$0.35} &               1.91$\pm$5.4 &  \underline{1.04$\pm$0.01} &          \textbf{1.02$\pm$0.03} \\
                  41167 &                     5.87$\pm$0.0 &              5.37$\pm$6.06 &     \textbf{2.31$\pm$0.03} &          \textbf{2.31$\pm$0.03} \\
                  41168 &           \textbf{0.76$\pm$0.04} &     \textbf{0.76$\pm$0.05} &   \underline{0.8$\pm$0.06} &       \underline{0.78$\pm$0.02} \\
                  41169 &                    3.83$\pm$0.33 &              3.45$\pm$0.44 &              3.26$\pm$0.02 &          \textbf{3.07$\pm$0.02} \\
                  42734 &        \underline{0.72$\pm$0.05} &  \underline{0.65$\pm$0.04} &  \underline{0.64$\pm$0.05} &          \textbf{0.62$\pm$0.01} \\
\bottomrule
\end{tabular}

    \caption{Avg. test Log-Loss on multi-class classification datasets (1h timeout).}
    \label{tab:results:logloss:1h}
\end{table}

\begin{table}[h!]
    \centering
    \begin{tabular}{r||rrrr}
\toprule
 \multicolumn{1}{c}{id} & \multicolumn{1}{c}{auto-sklearn} &   \multicolumn{1}{c}{GAMA} &  \multicolumn{1}{c}{naive} & \multicolumn{1}{c}{quasi-naive} \\
\midrule
                     12 &           \textbf{0.12$\pm$0.06} &              0.47$\pm$0.96 &              1.45$\pm$5.03 &       \underline{0.13$\pm$0.05} \\
                     23 &            \textbf{0.9$\pm$0.04} &  \underline{0.91$\pm$0.04} &              2.73$\pm$5.54 &       \underline{0.93$\pm$0.06} \\
                     54 &           \textbf{0.36$\pm$0.04} &              3.7$\pm$10.47 &                        nan &       \underline{0.43$\pm$0.11} \\
                    181 &           \textbf{1.05$\pm$0.05} &  \underline{4.04$\pm$8.23} &              1.18$\pm$0.23 &       \underline{1.06$\pm$0.09} \\
                    188 &           \textbf{1.12$\pm$0.07} &  \underline{1.16$\pm$0.06} &                1.3$\pm$0.1 &       \underline{1.19$\pm$0.07} \\
                   1457 &        \underline{1.31$\pm$0.24} &  \underline{1.23$\pm$0.41} &                2.3$\pm$0.8 &          \textbf{1.13$\pm$0.29} \\
                   1468 &        \underline{0.19$\pm$0.09} &   \underline{0.19$\pm$0.1} &  \underline{0.18$\pm$1.08} &          \textbf{0.16$\pm$0.37} \\
                   1475 &           \textbf{1.07$\pm$0.03} &     \textbf{1.07$\pm$0.05} &  \underline{1.12$\pm$0.07} &          \textbf{1.07$\pm$0.03} \\
                   1515 &           \textbf{0.41$\pm$0.14} &  \underline{0.48$\pm$0.17} &  \underline{0.43$\pm$0.07} &       \underline{0.46$\pm$0.06} \\
                   4538 &        \underline{0.85$\pm$0.02} &      \textbf{0.8$\pm$0.03} &  \underline{0.85$\pm$0.02} &                   0.91$\pm$0.23 \\
                   4541 &             \textbf{0.9$\pm$0.0} &      \textbf{0.9$\pm$0.01} &  \underline{0.92$\pm$0.01} &            \textbf{0.9$\pm$0.0} \\
                  40498 &           \textbf{0.76$\pm$0.03} &              1.39$\pm$4.88 &                        nan &       \underline{0.85$\pm$0.32} \\
                  40668 &           \textbf{0.24$\pm$0.12} &              0.39$\pm$0.13 &              0.38$\pm$0.05 &                   0.36$\pm$0.05 \\
                  40670 &           \textbf{0.09$\pm$0.02} &  \underline{0.12$\pm$0.02} &  \underline{0.17$\pm$0.05} &       \underline{0.11$\pm$0.09} \\
                  40685 &             \textbf{0.0$\pm$0.0} &       \textbf{0.0$\pm$0.0} &      \textbf{0.0$\pm$0.01} &            \textbf{0.0$\pm$0.0} \\
                  40975 &             \textbf{0.0$\pm$0.0} &       \textbf{0.0$\pm$0.0} &              0.36$\pm$0.25 &       \underline{0.01$\pm$0.14} \\
                  40982 &        \underline{0.52$\pm$0.06} &  \underline{0.53$\pm$0.08} &   \underline{0.62$\pm$0.0} &          \textbf{0.51$\pm$0.07} \\
                  40984 &           \textbf{0.07$\pm$0.02} &     \textbf{0.07$\pm$0.03} &     \textbf{0.07$\pm$0.03} &       \underline{0.32$\pm$1.03} \\
                  40996 &           \textbf{0.29$\pm$0.03} &  \underline{0.39$\pm$0.07} &  \underline{0.37$\pm$0.01} &                   0.53$\pm$0.36 \\
                  41027 &        \underline{0.18$\pm$0.05} &     \textbf{0.11$\pm$0.07} &              0.49$\pm$0.21 &       \underline{0.19$\pm$0.02} \\
                  41163 &           \textbf{0.03$\pm$0.02} &     \textbf{0.03$\pm$0.01} &  \underline{0.05$\pm$0.01} &       \underline{0.04$\pm$0.01} \\
                  41164 &            \textbf{0.8$\pm$0.03} &  \underline{0.81$\pm$0.05} &  \underline{0.85$\pm$0.03} &       \underline{0.82$\pm$0.03} \\
                  41165 &        \underline{1.64$\pm$0.15} &     \textbf{1.62$\pm$0.14} &   \underline{1.7$\pm$0.03} &       \underline{1.69$\pm$0.18} \\
                  41166 &           \textbf{0.86$\pm$0.07} &     \textbf{0.86$\pm$0.07} &              1.02$\pm$0.03 &       \underline{0.94$\pm$0.03} \\
                  41167 &           \textbf{2.14$\pm$0.57} &            10.68$\pm$10.68 &              2.69$\pm$0.11 &       \underline{2.31$\pm$0.03} \\
                  41168 &           \textbf{0.68$\pm$0.02} &      \textbf{0.68$\pm$0.2} &  \underline{0.78$\pm$0.01} &       \underline{0.74$\pm$0.01} \\
                  41169 &                    2.77$\pm$0.17 &              3.44$\pm$1.52 &              2.79$\pm$0.06 &          \textbf{2.62$\pm$0.03} \\
                  42734 &           \textbf{0.61$\pm$0.01} &  \underline{0.62$\pm$0.03} &  \underline{0.63$\pm$0.04} &           \textbf{0.61$\pm$0.0} \\
\bottomrule
\end{tabular}
    \caption{Avg. test Log-Loss on multi-class classification datasets (1d timeout).}
    \label{tab:results:logloss:1d}
\end{table}

\clearpage
\newpage
\section{Performance Plots over time}
\label{appendix:performancecurves}
\includegraphics[width=\textwidth]{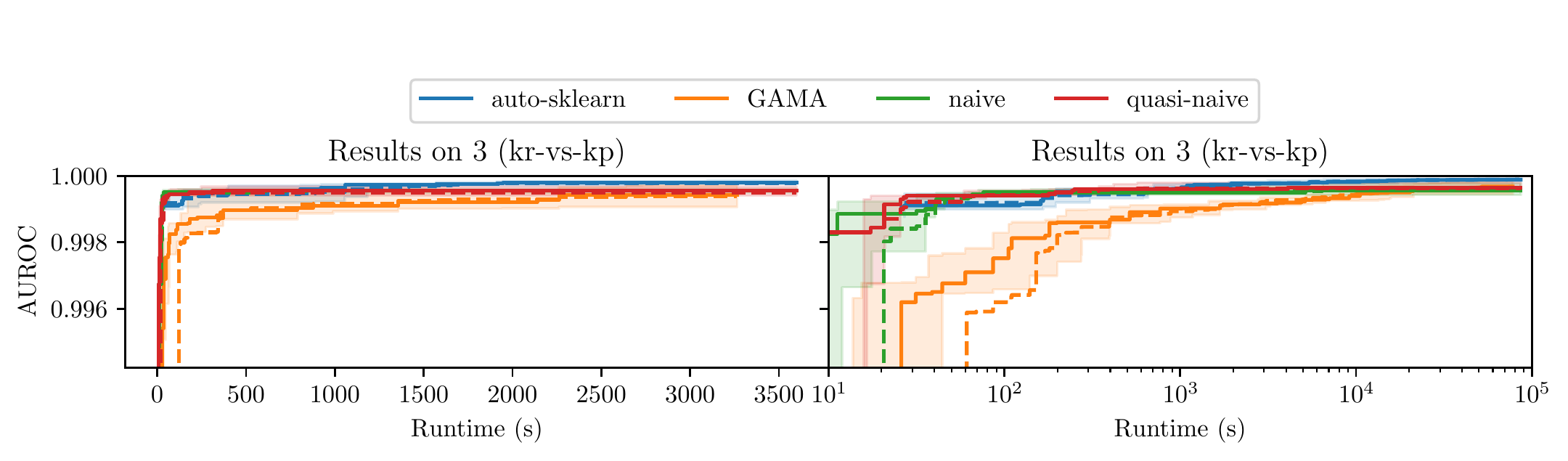}\\[-2em]
\includegraphics[width=\textwidth]{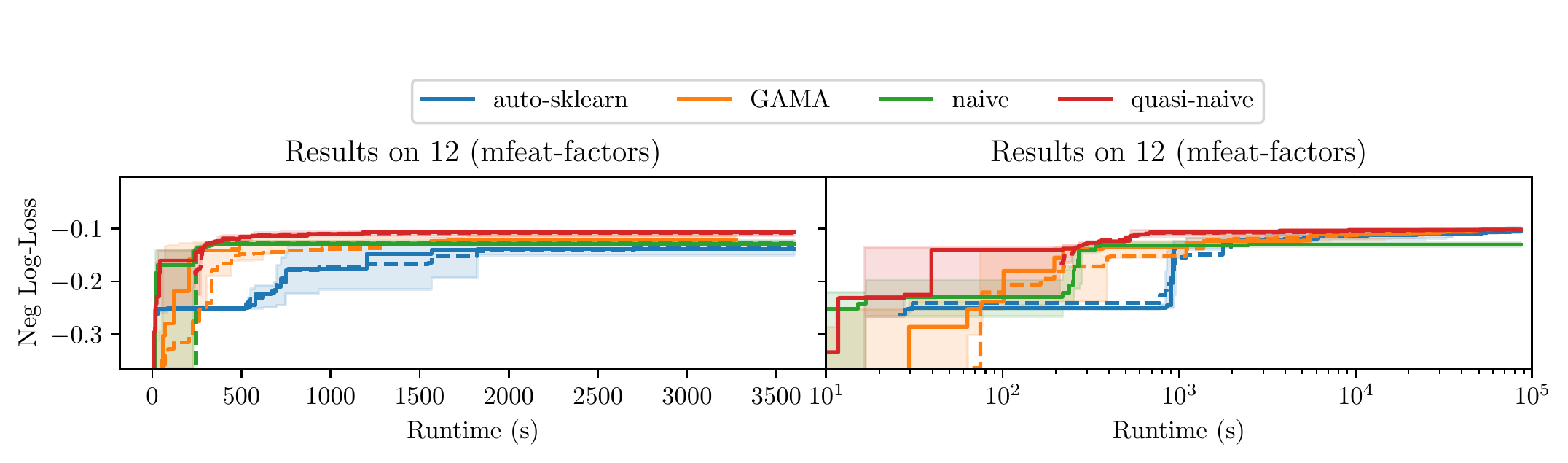}\\[-2em]
\includegraphics[width=\textwidth]{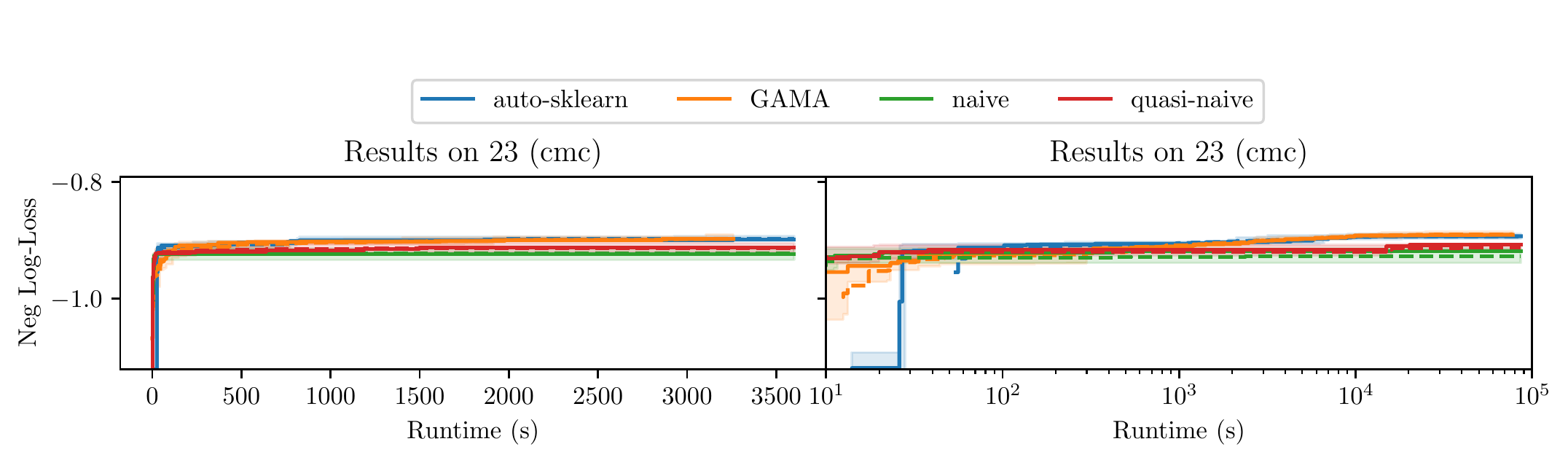}\\[-2em]
\includegraphics[width=\textwidth]{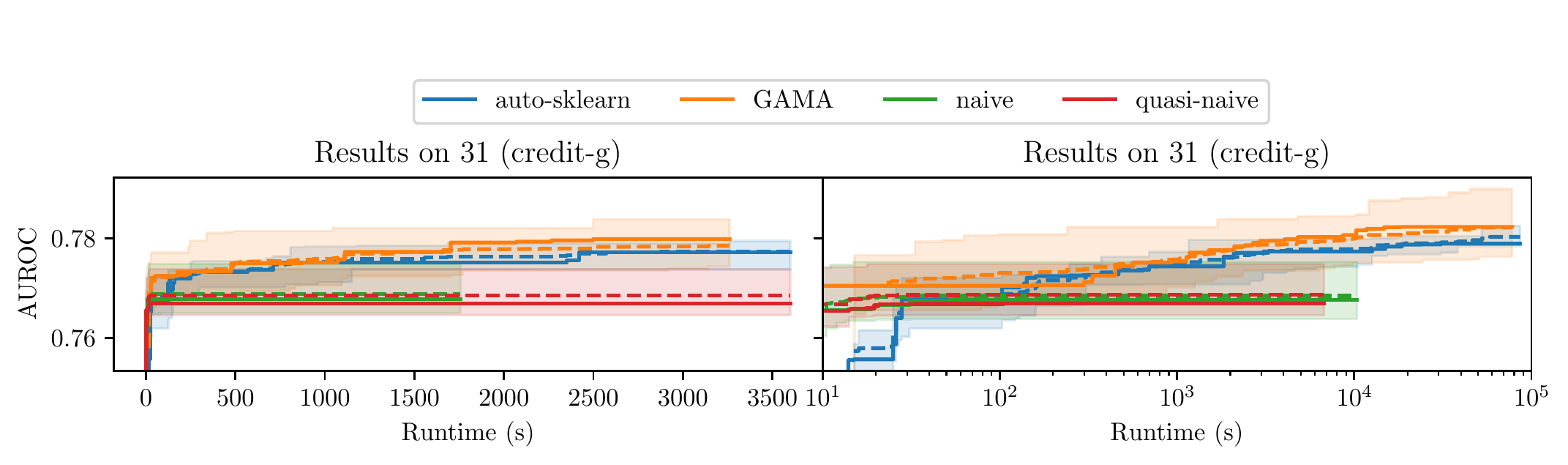}\\[-2em]
\includegraphics[width=\textwidth]{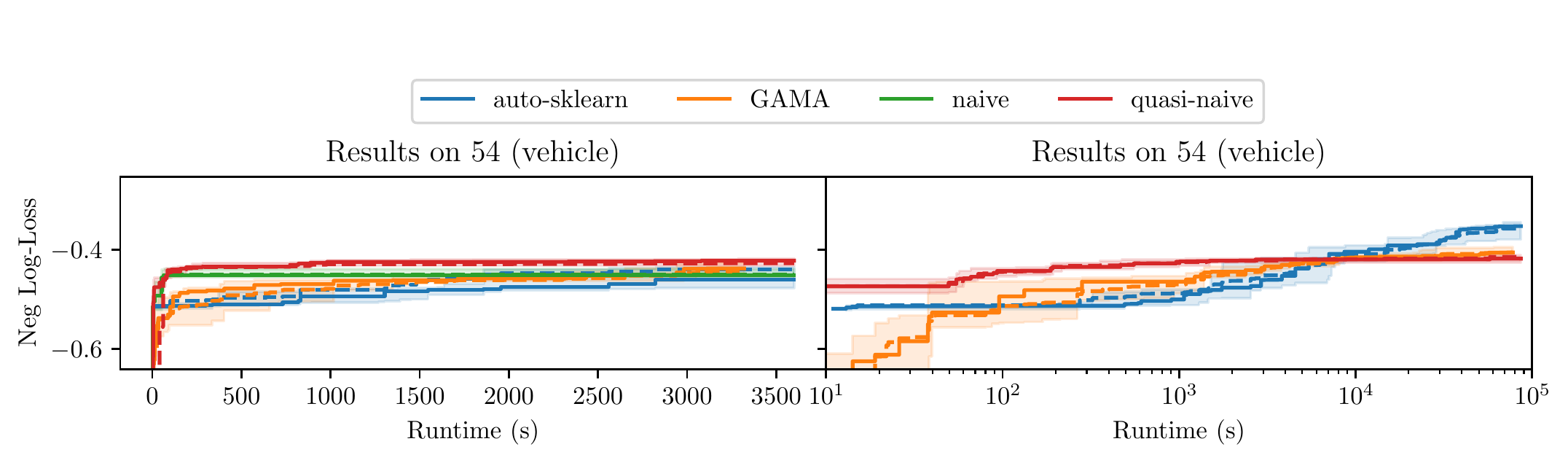}\\[-2em]
\includegraphics[width=\textwidth]{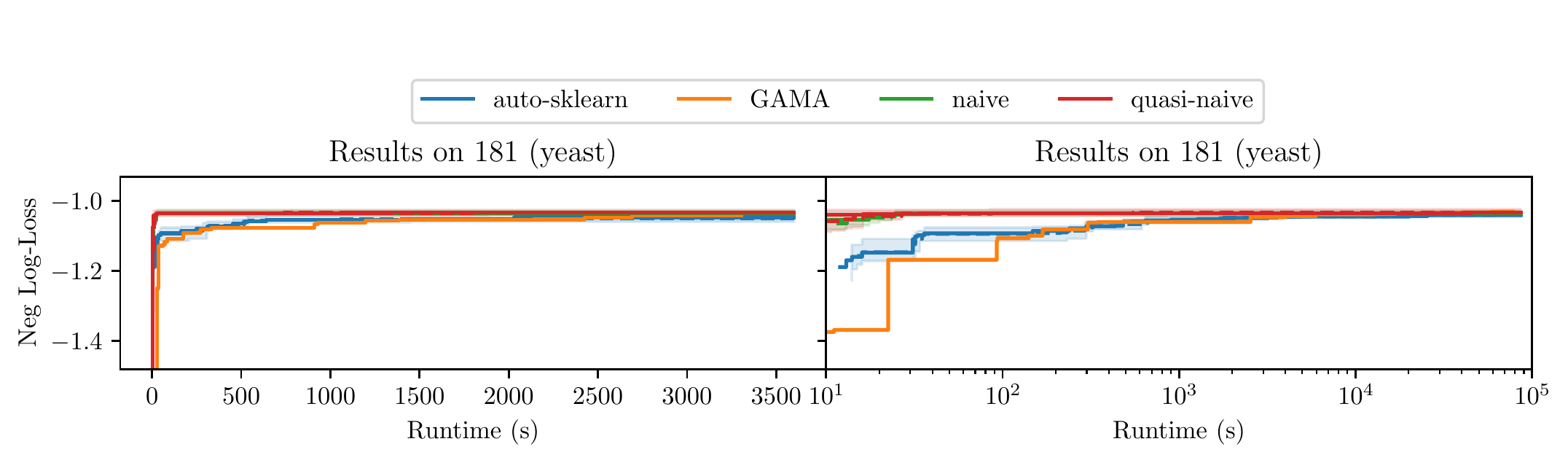}\\[-2em]
\includegraphics[width=\textwidth]{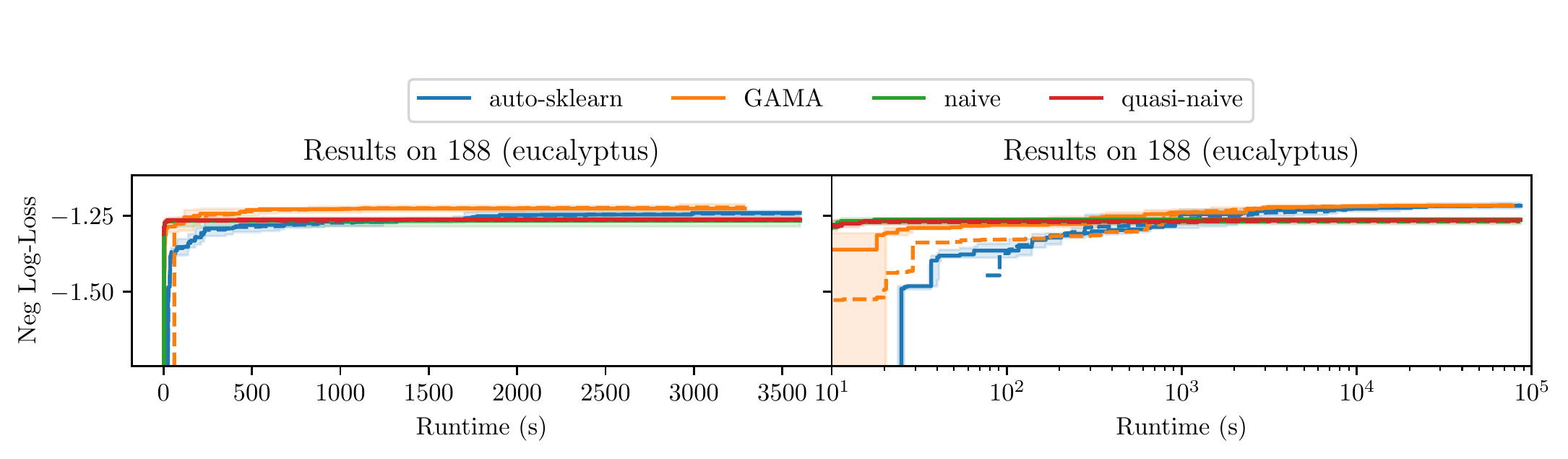}\\[-2em]
\includegraphics[width=\textwidth]{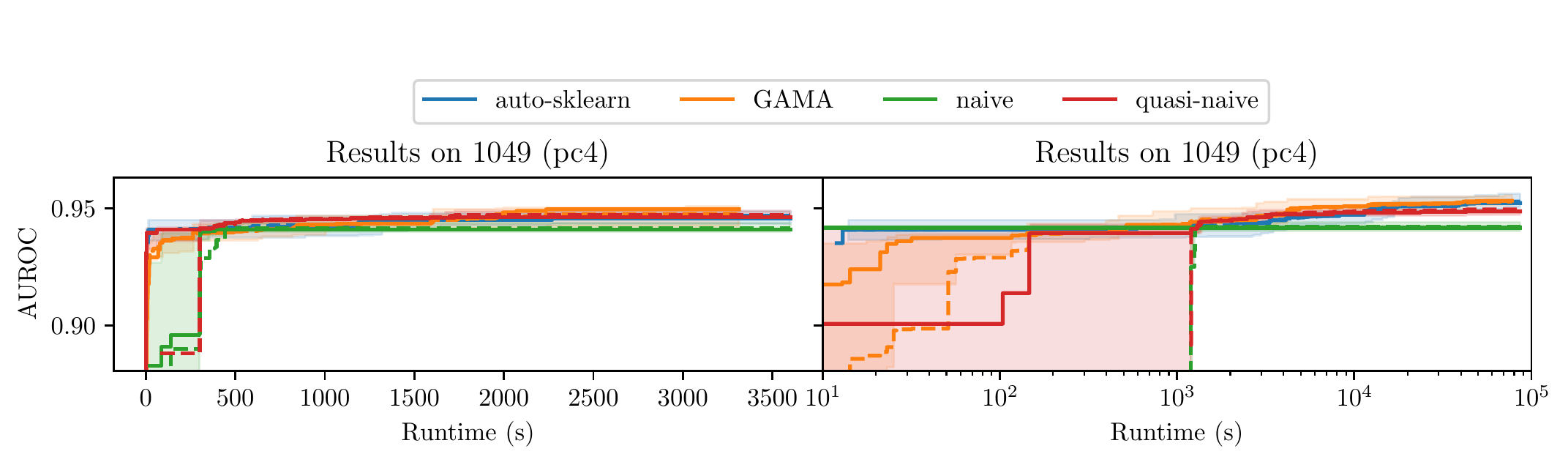}\\[-2em]
\includegraphics[width=\textwidth]{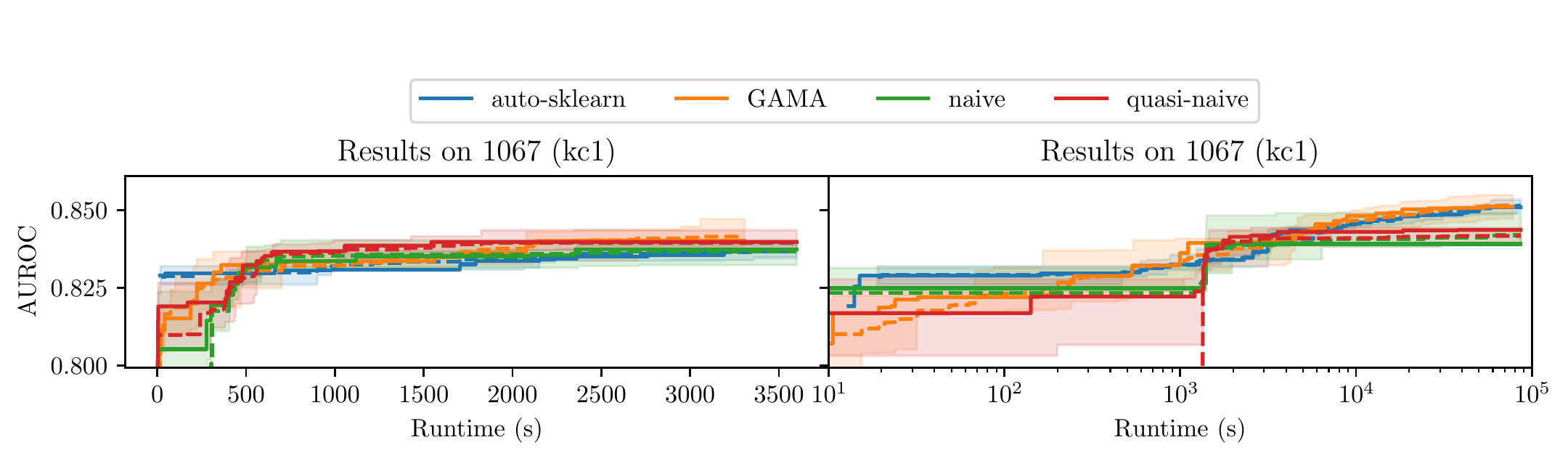}\\[-2em]
\includegraphics[width=\textwidth]{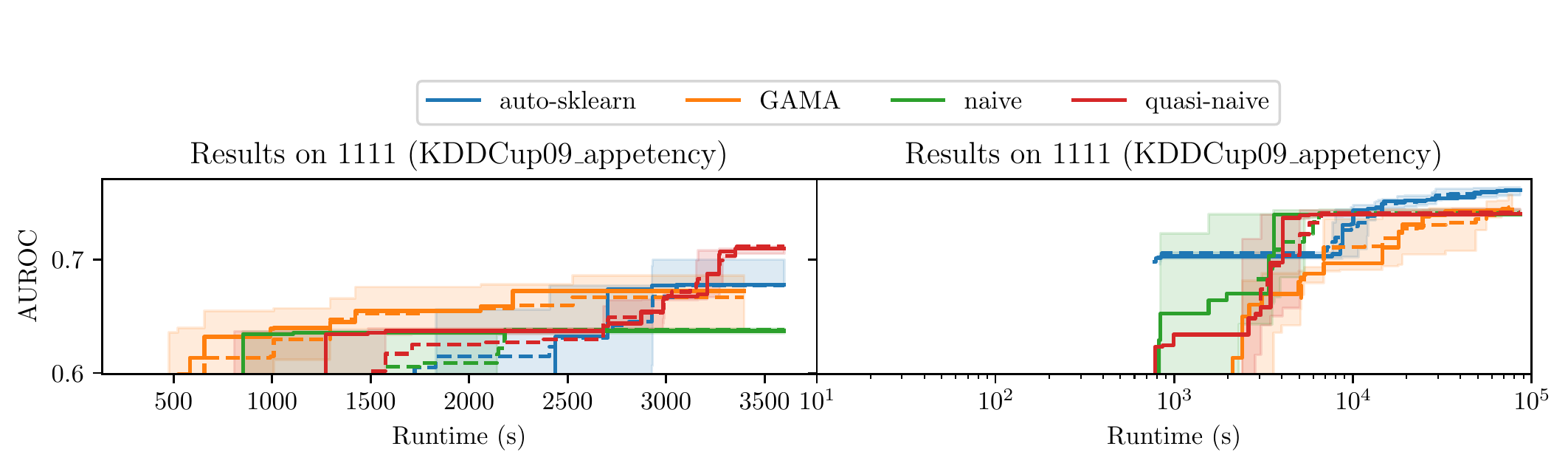}\\[-2em]
\includegraphics[width=\textwidth]{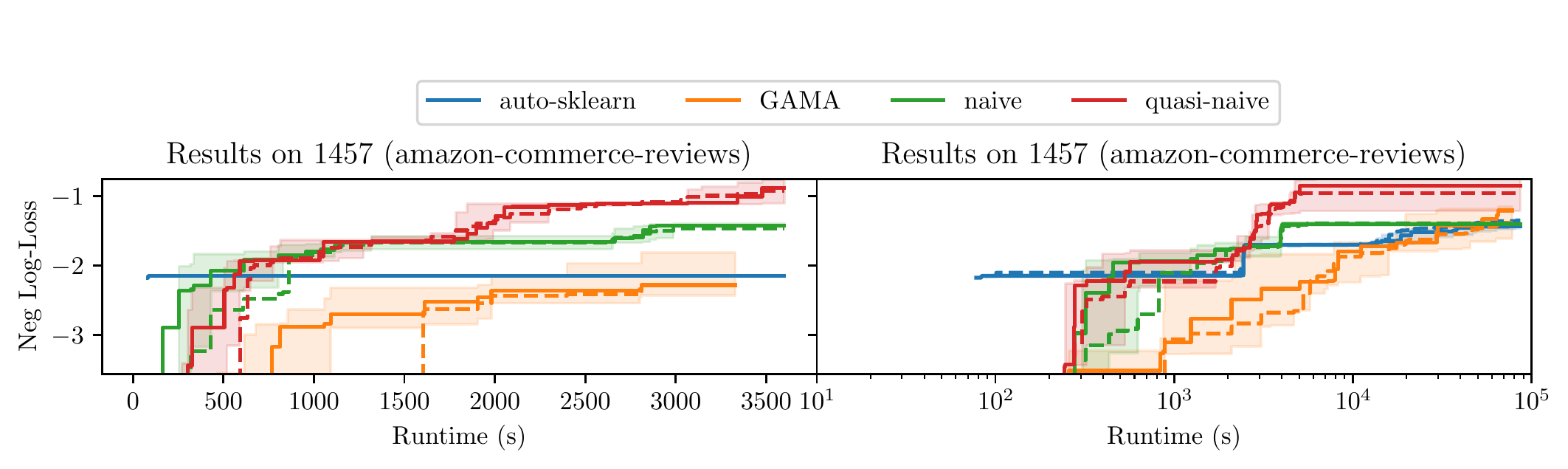}\\[-2em]
\includegraphics[width=\textwidth]{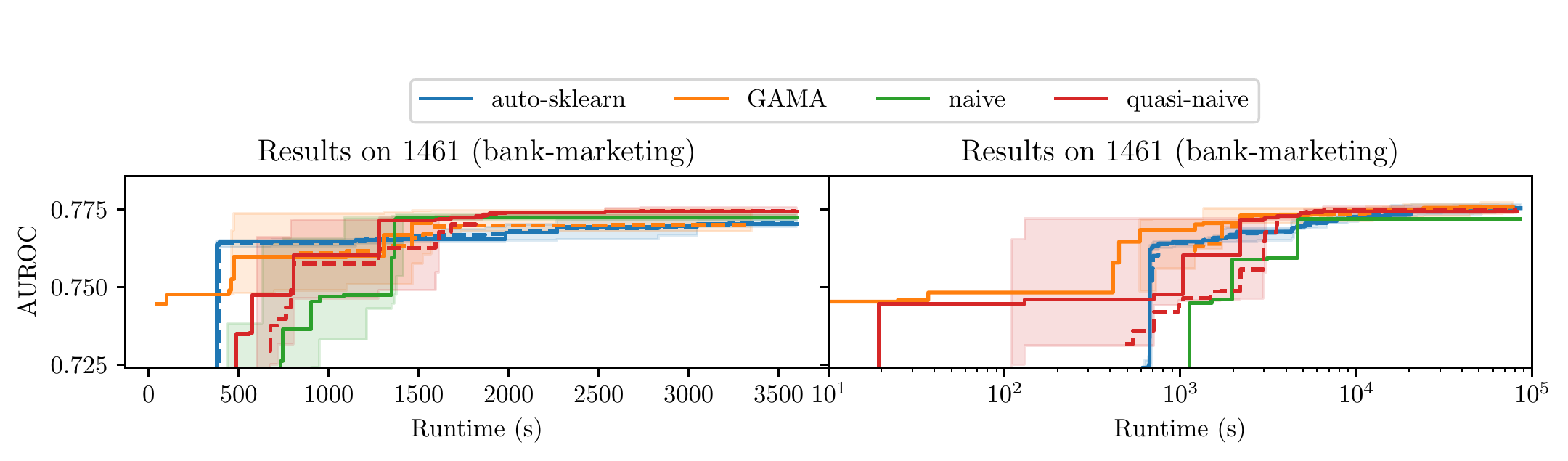}\\[-2em]
\includegraphics[width=\textwidth]{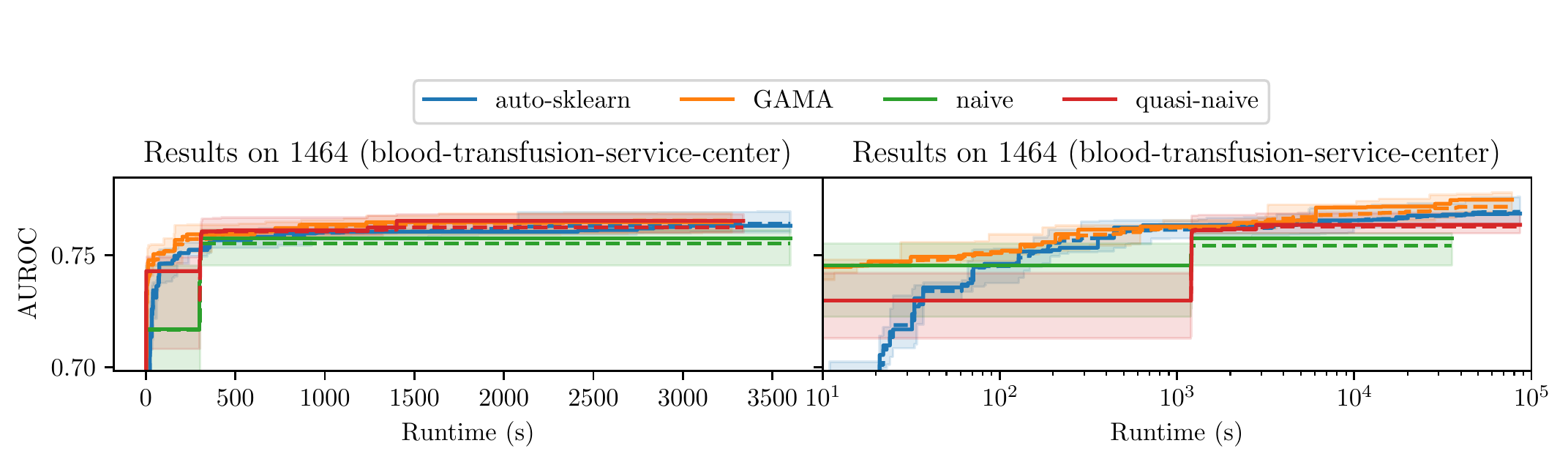}\\[-2em]
\includegraphics[width=\textwidth]{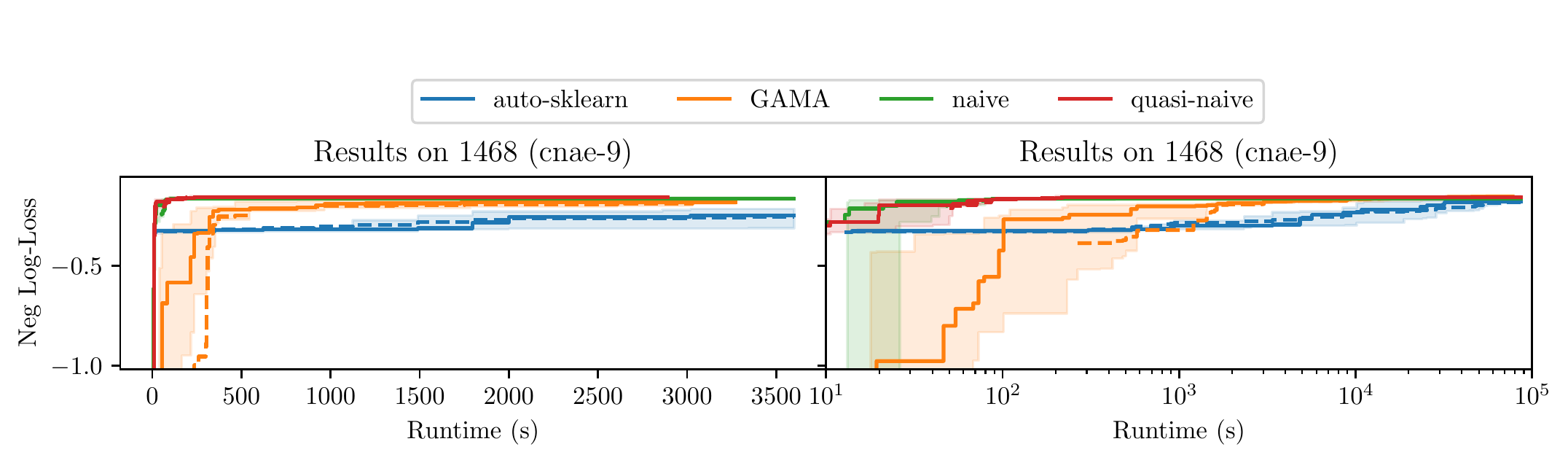}\\[-2em]
\includegraphics[width=\textwidth]{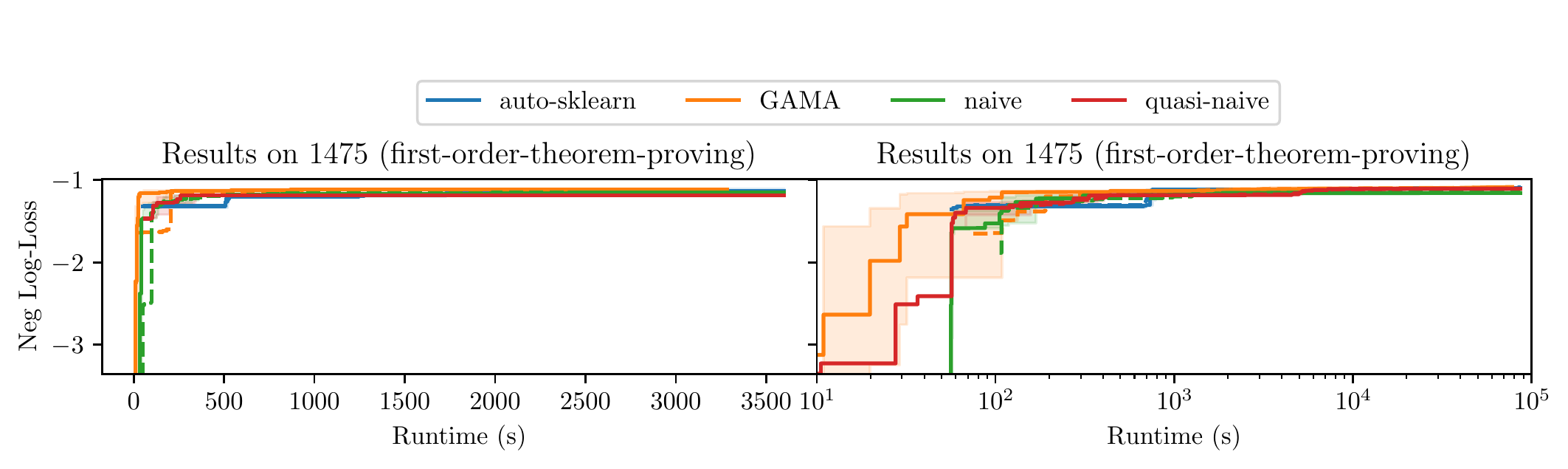}\\[-2em]
\includegraphics[width=\textwidth]{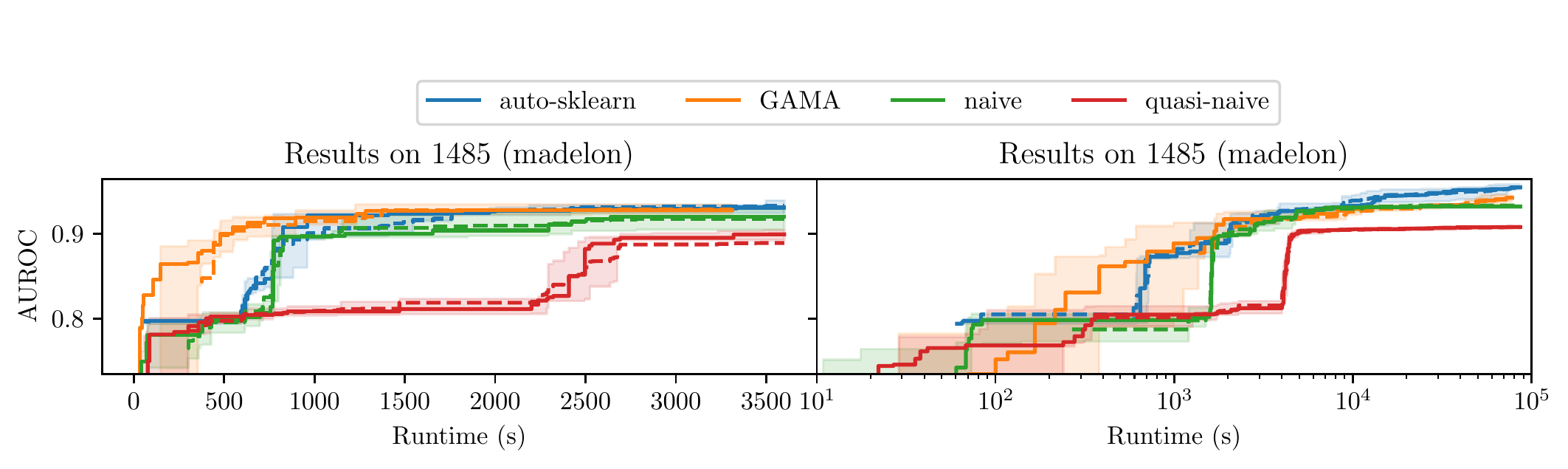}\\[-2em]
\includegraphics[width=\textwidth]{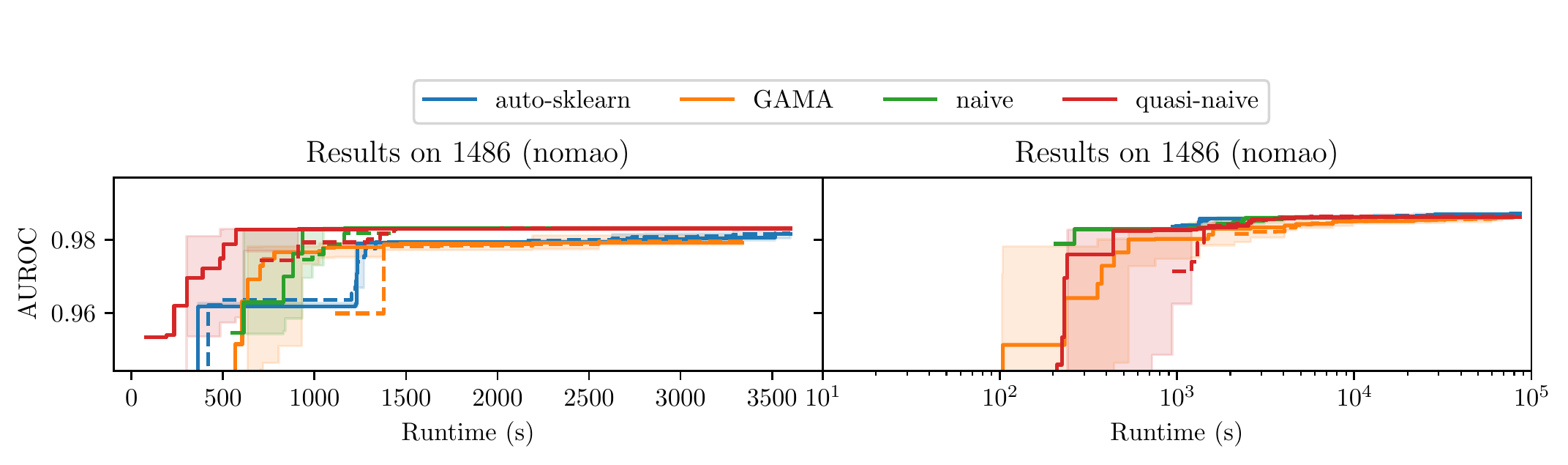}\\[-2em]
\includegraphics[width=\textwidth]{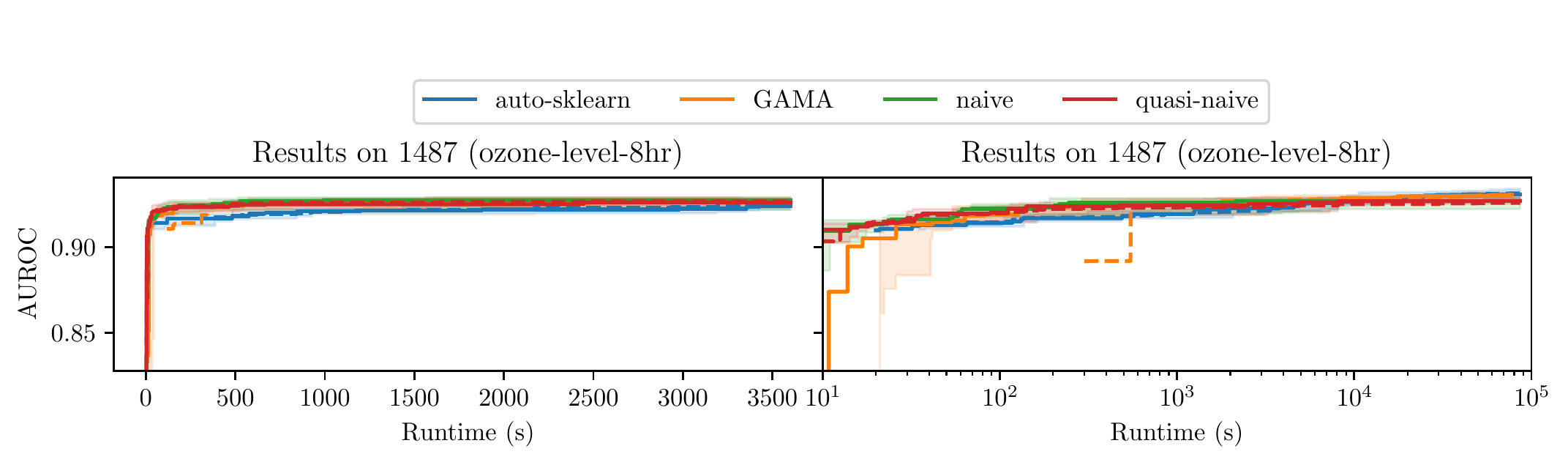}\\[-2em]
\includegraphics[width=\textwidth]{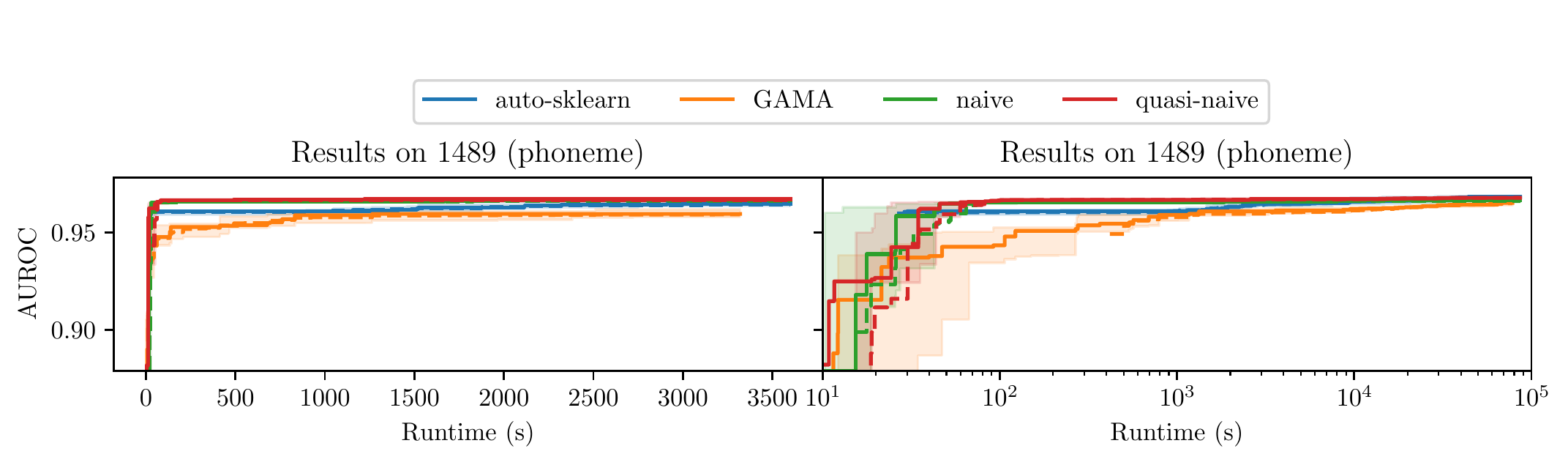}\\[-2em]
\includegraphics[width=\textwidth]{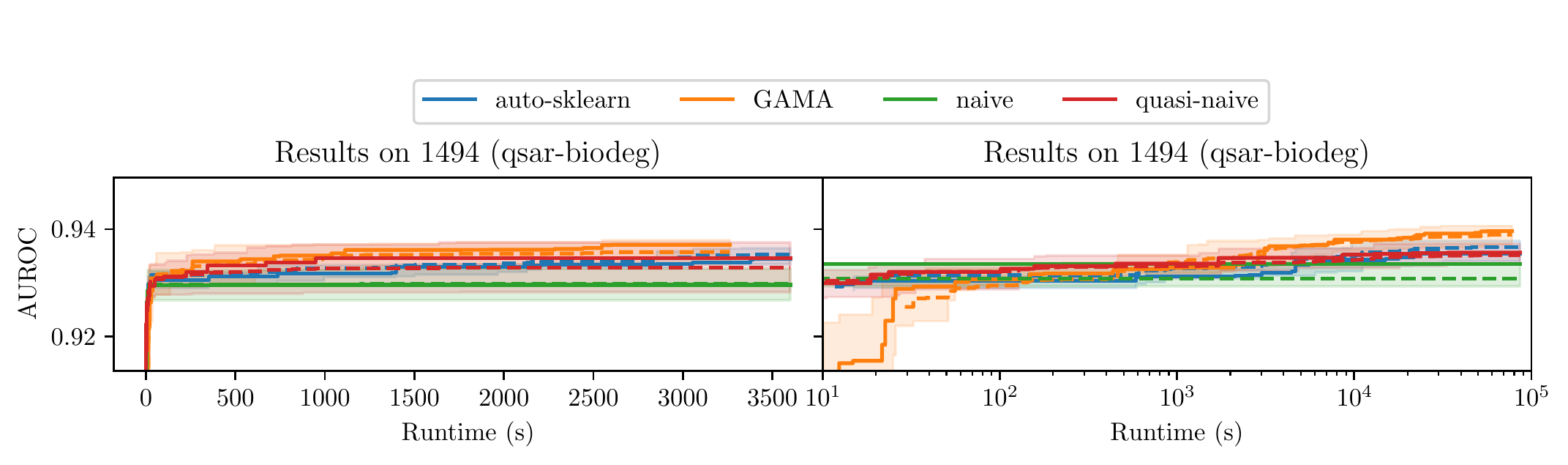}\\[-2em]
\includegraphics[width=\textwidth]{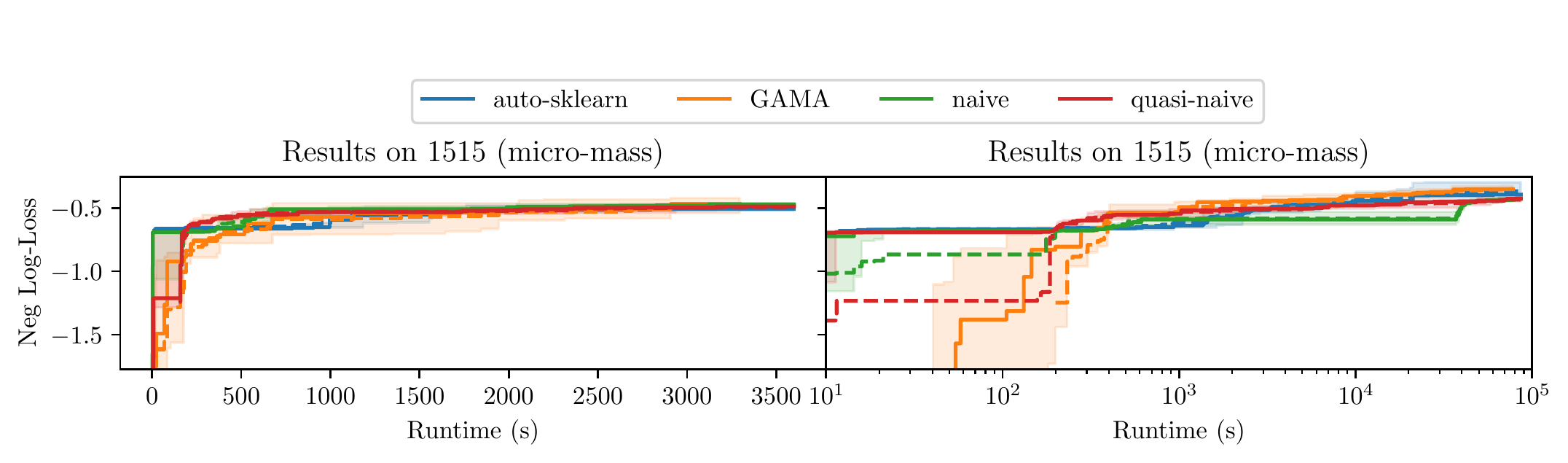}\\[-2em]
\includegraphics[width=\textwidth]{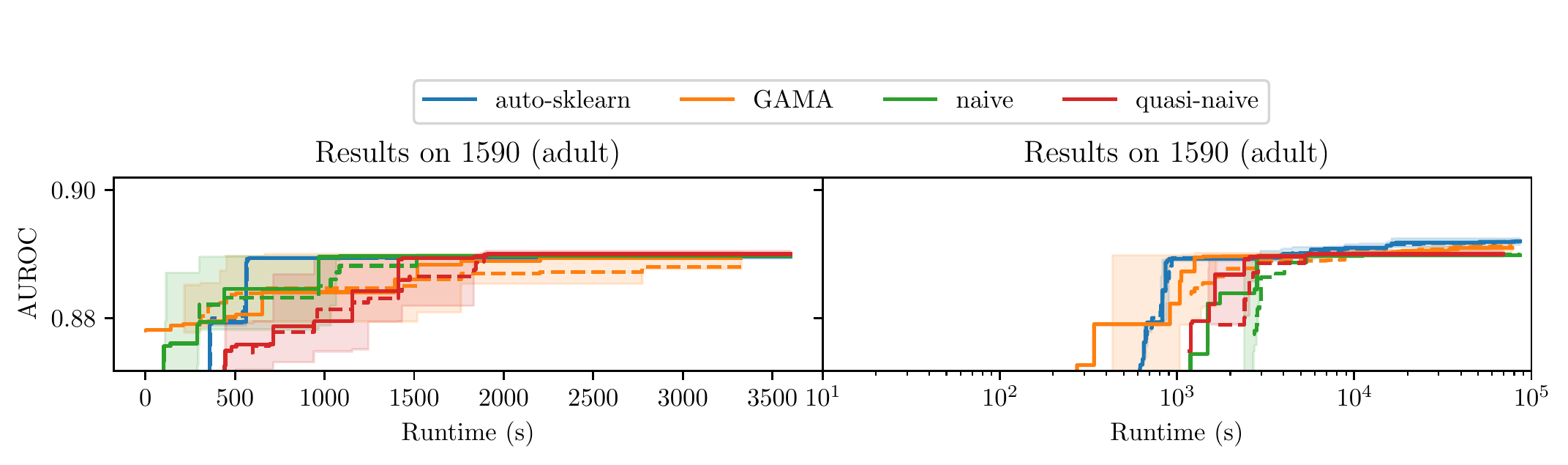}\\[-2em]
\includegraphics[width=\textwidth]{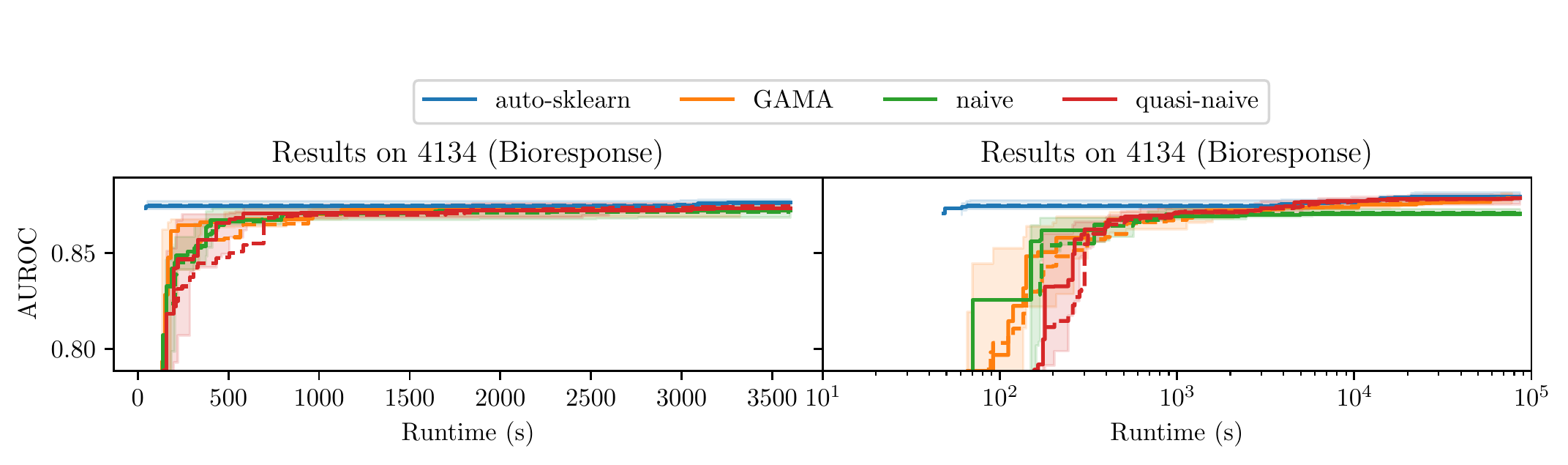}\\[-2em]
\includegraphics[width=\textwidth]{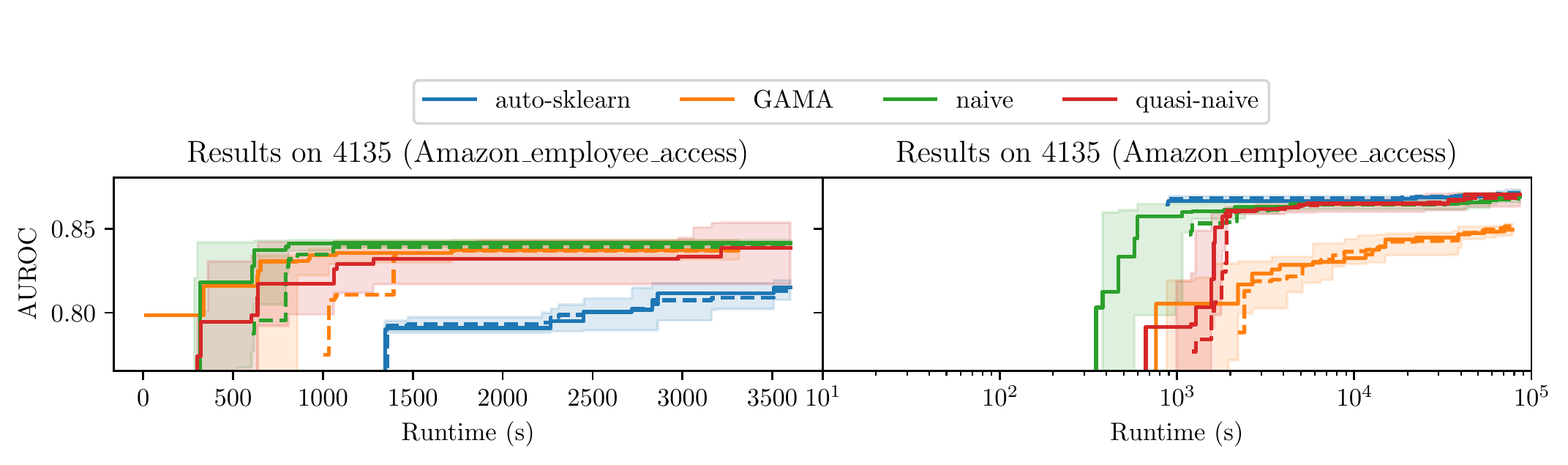}\\[-2em]
\includegraphics[width=\textwidth]{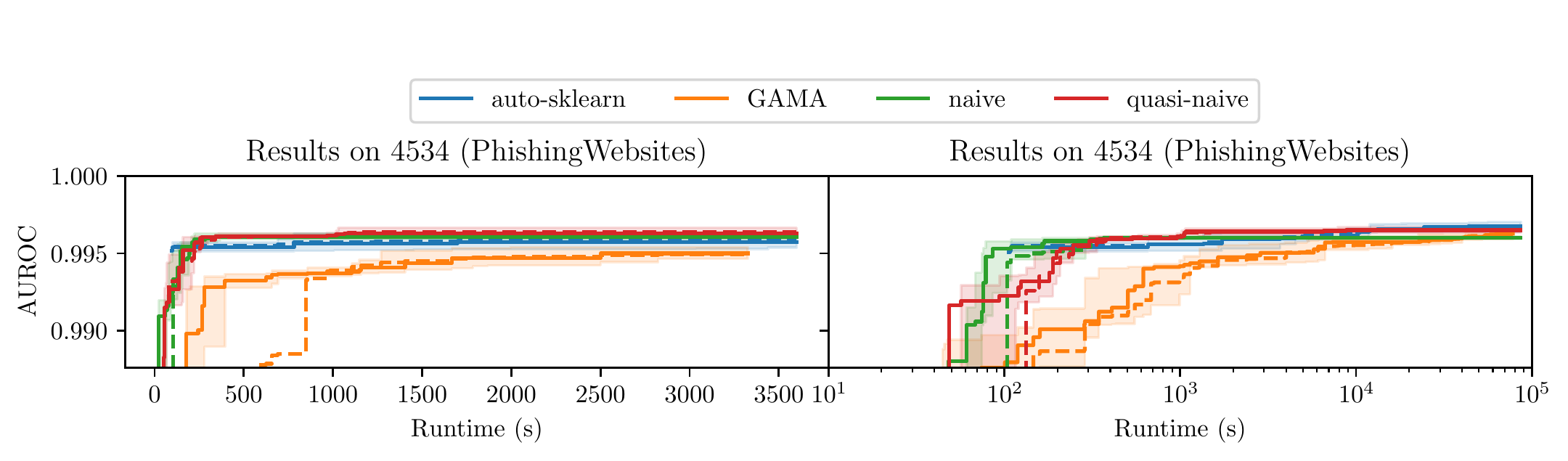}\\[-2em]
\includegraphics[width=\textwidth]{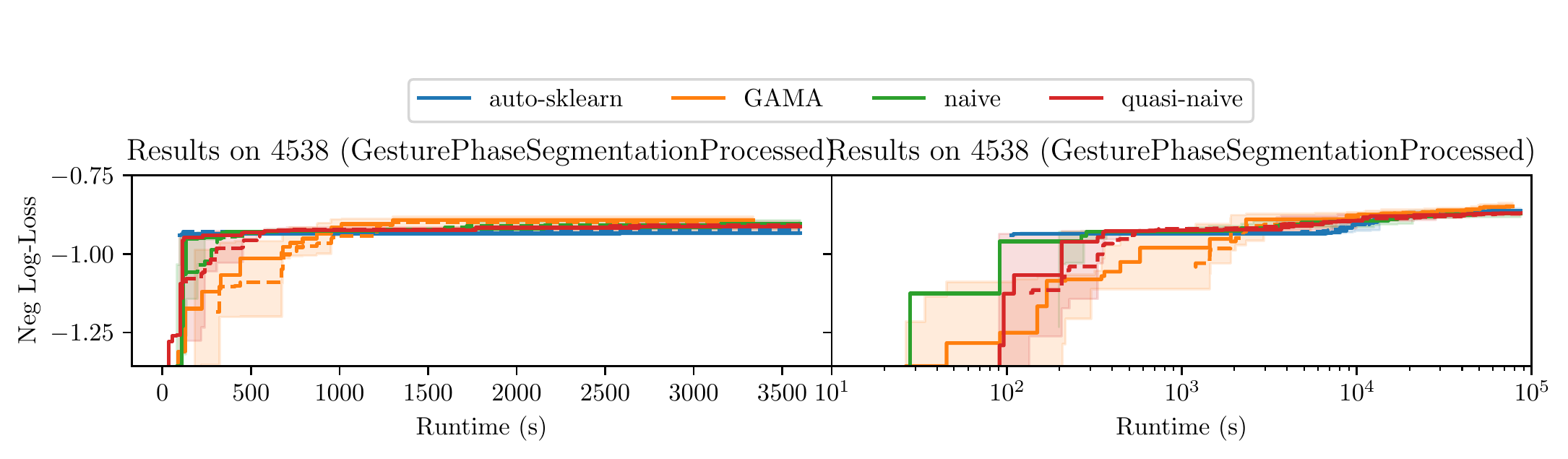}\\[-2em]
\includegraphics[width=\textwidth]{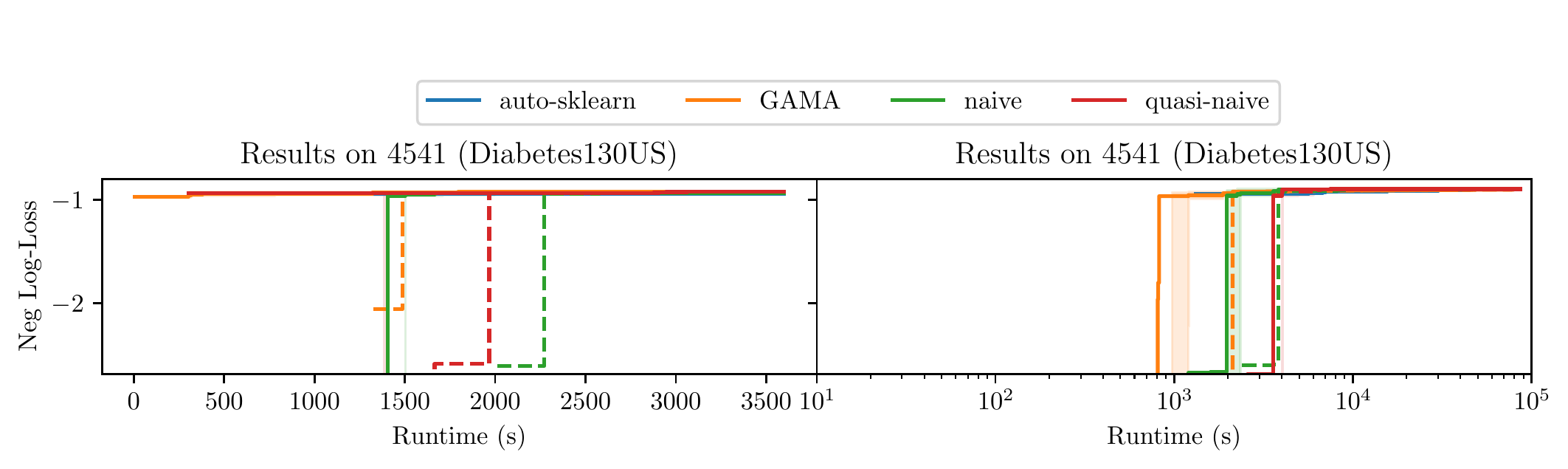}\\[-2em]
\includegraphics[width=\textwidth]{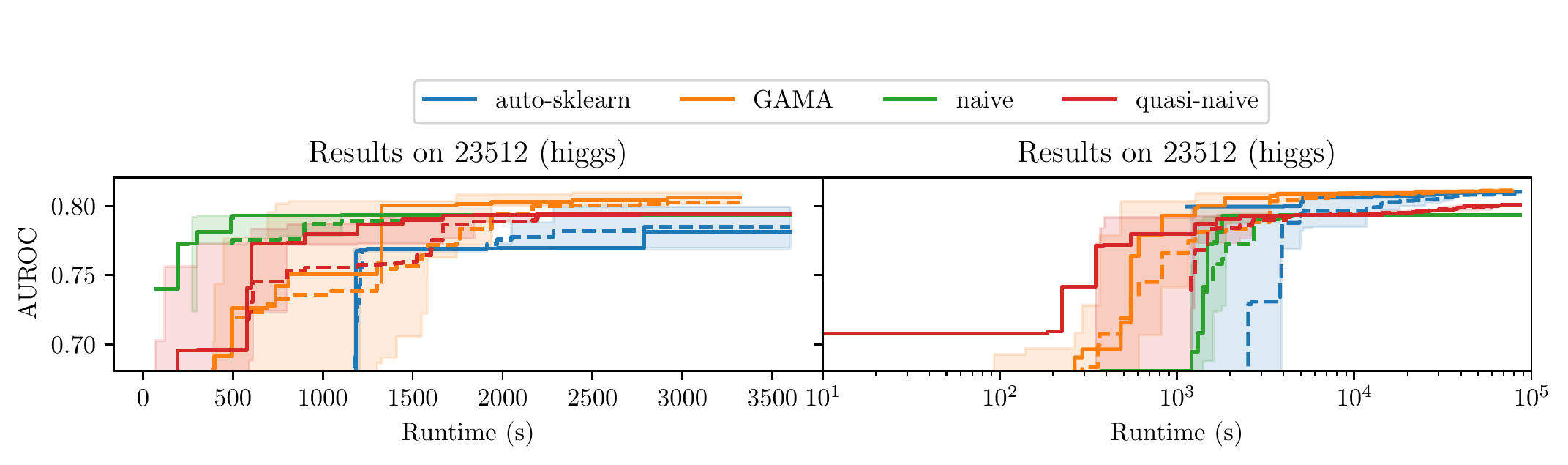}\\[-2em]
\includegraphics[width=\textwidth]{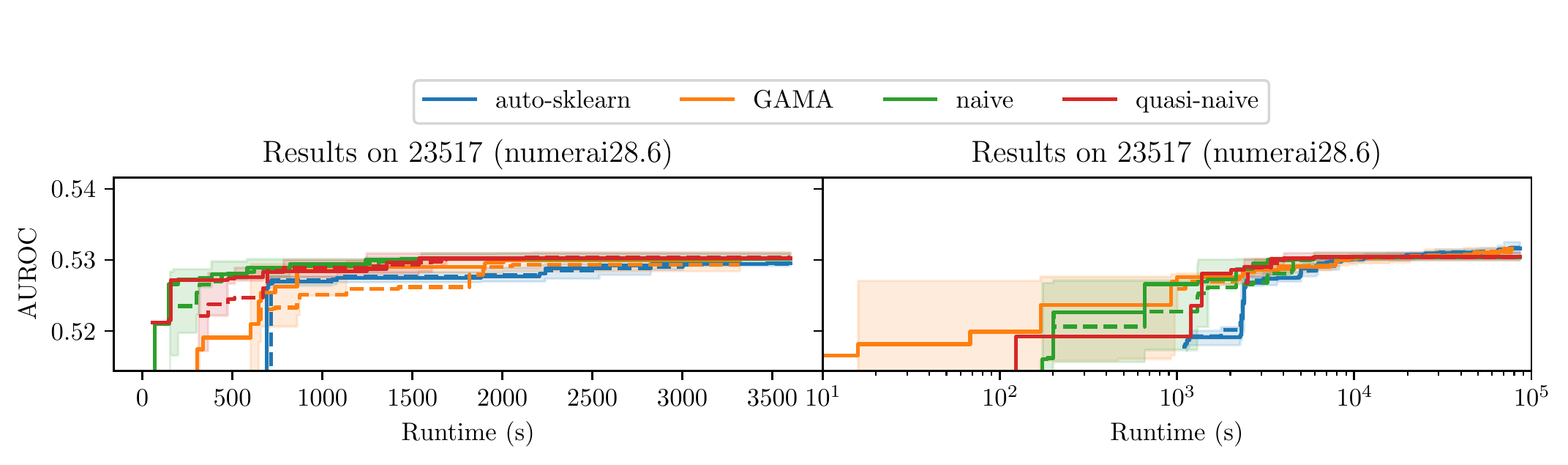}\\[-2em]
\includegraphics[width=\textwidth]{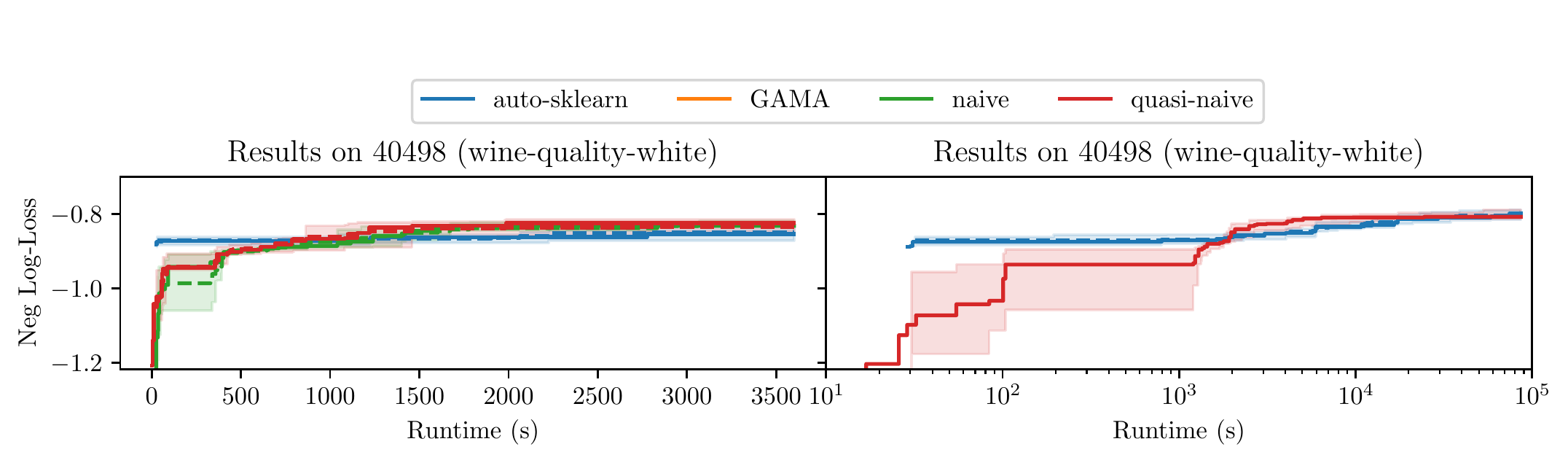}\\[-2em]
\includegraphics[width=\textwidth]{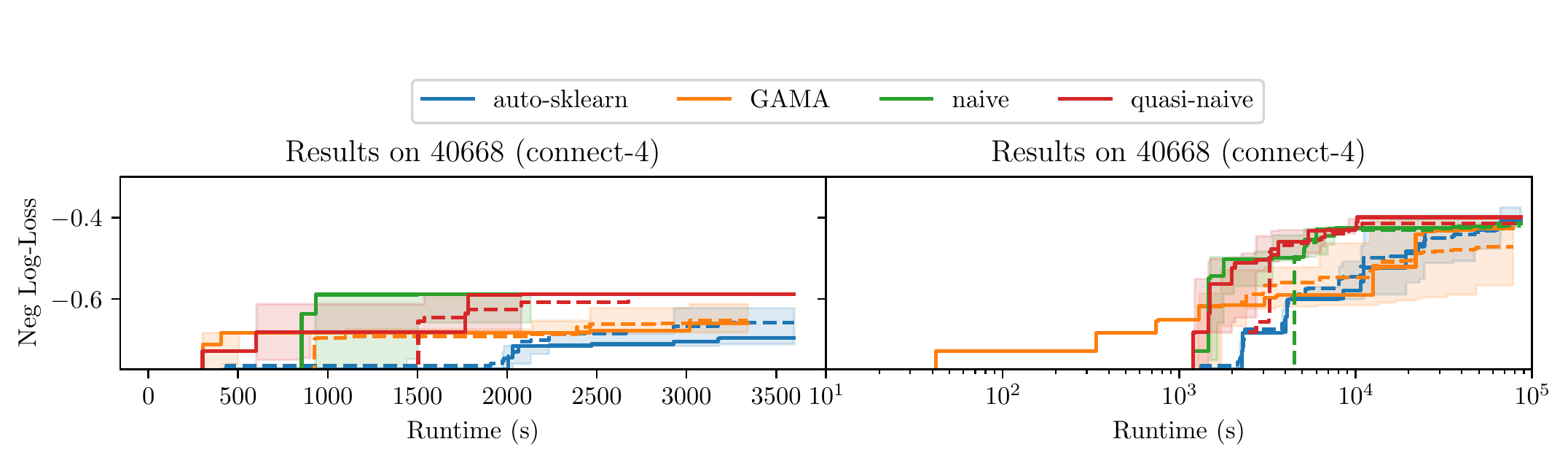}\\[-2em]
\includegraphics[width=\textwidth]{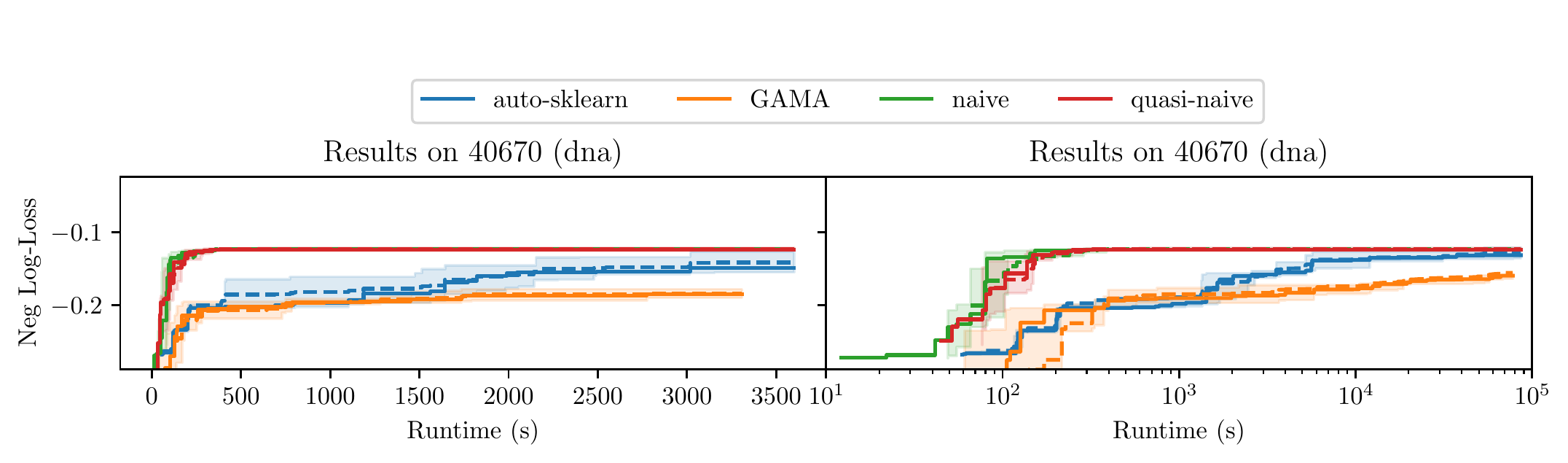}\\[-2em]
\includegraphics[width=\textwidth]{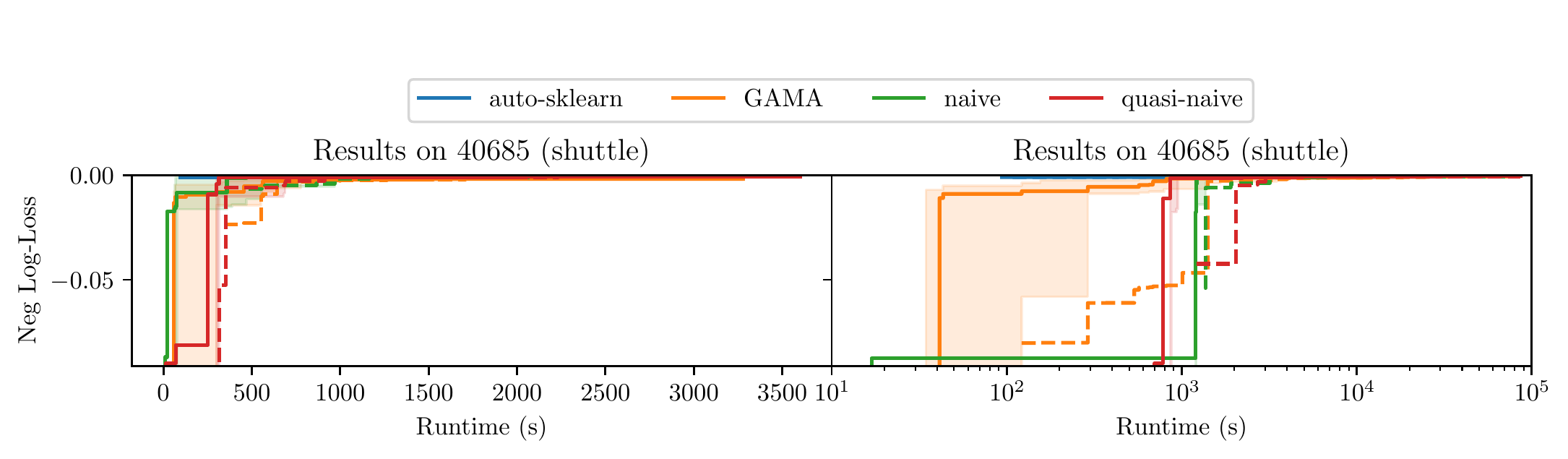}\\[-2em]
\includegraphics[width=\textwidth]{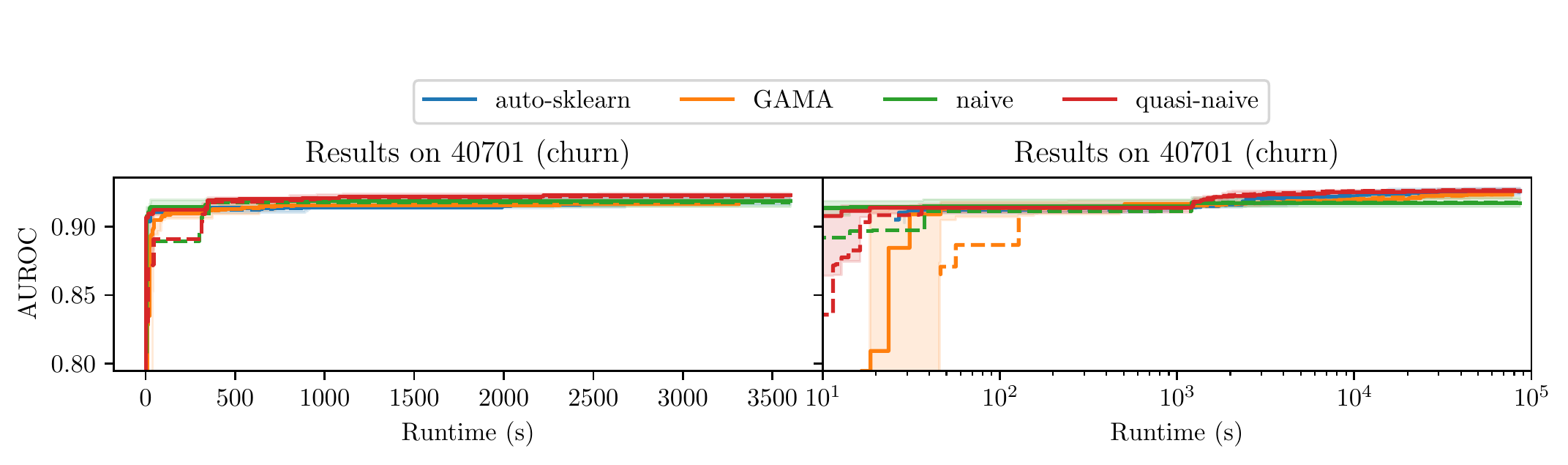}\\[-2em]
\includegraphics[width=\textwidth]{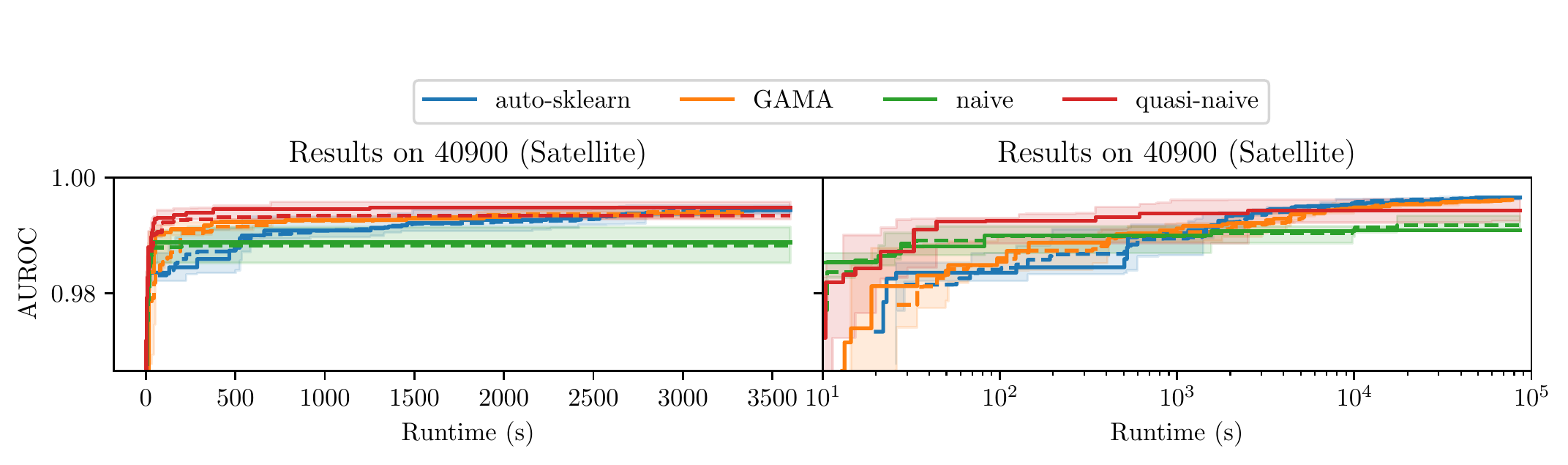}\\[-2em]
\includegraphics[width=\textwidth]{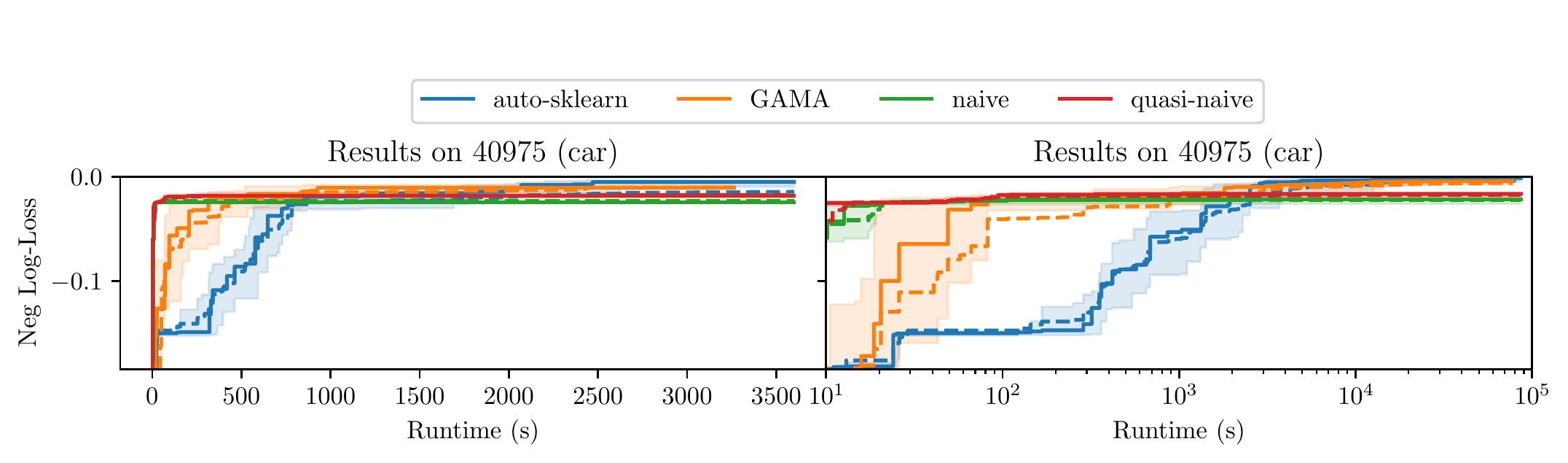}\\[-2em]
\includegraphics[width=\textwidth]{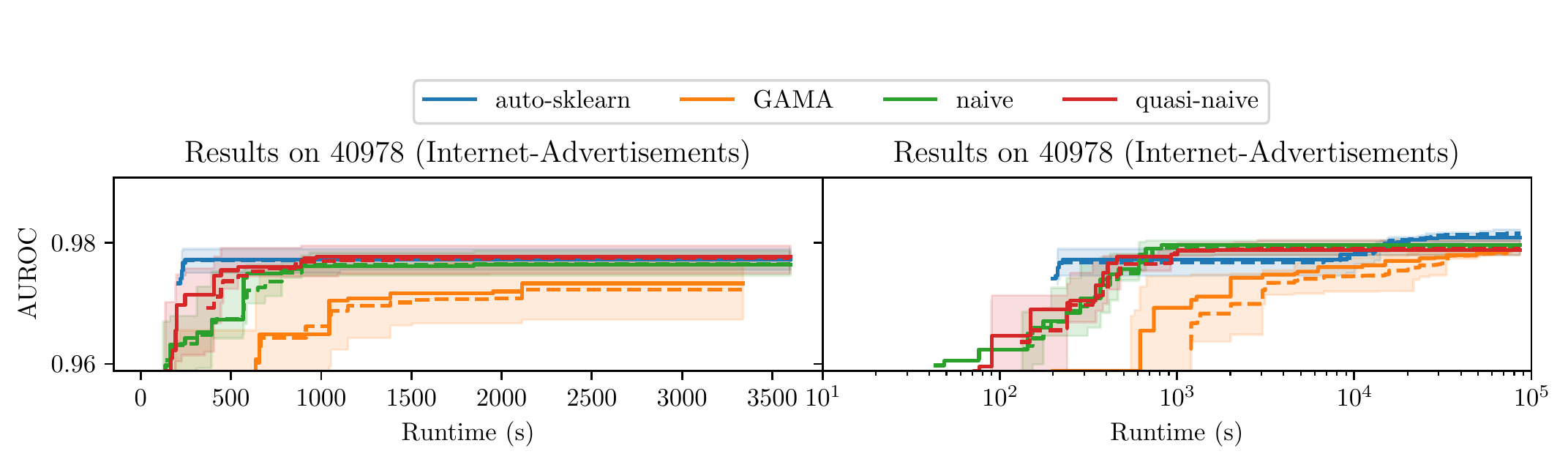}\\[-2em]
\includegraphics[width=\textwidth]{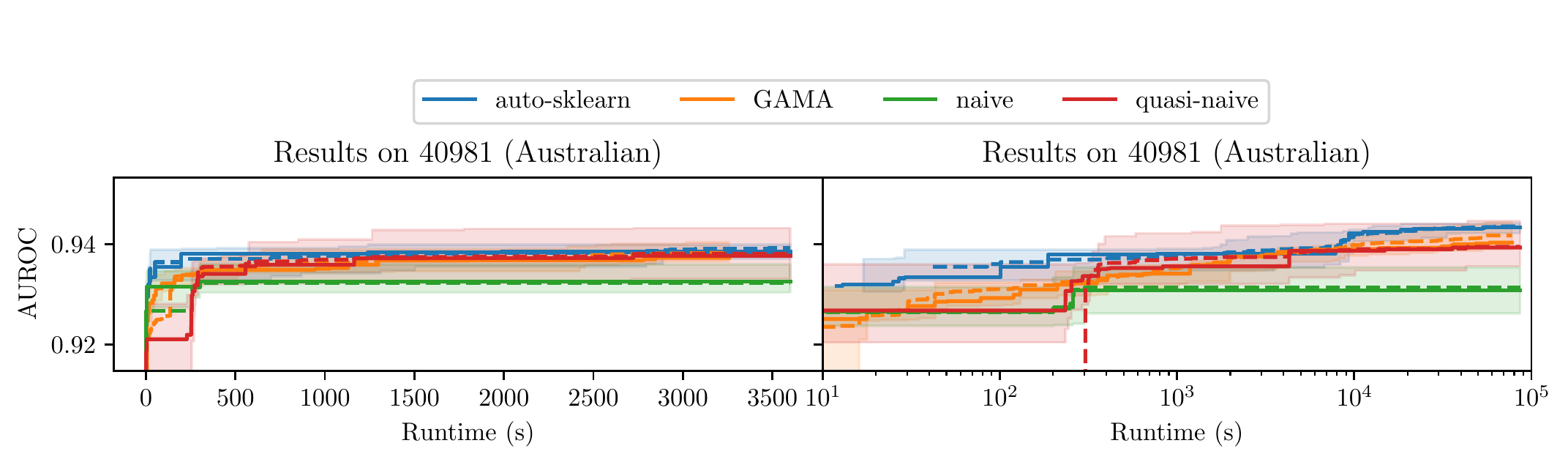}\\[-2em]
\includegraphics[width=\textwidth]{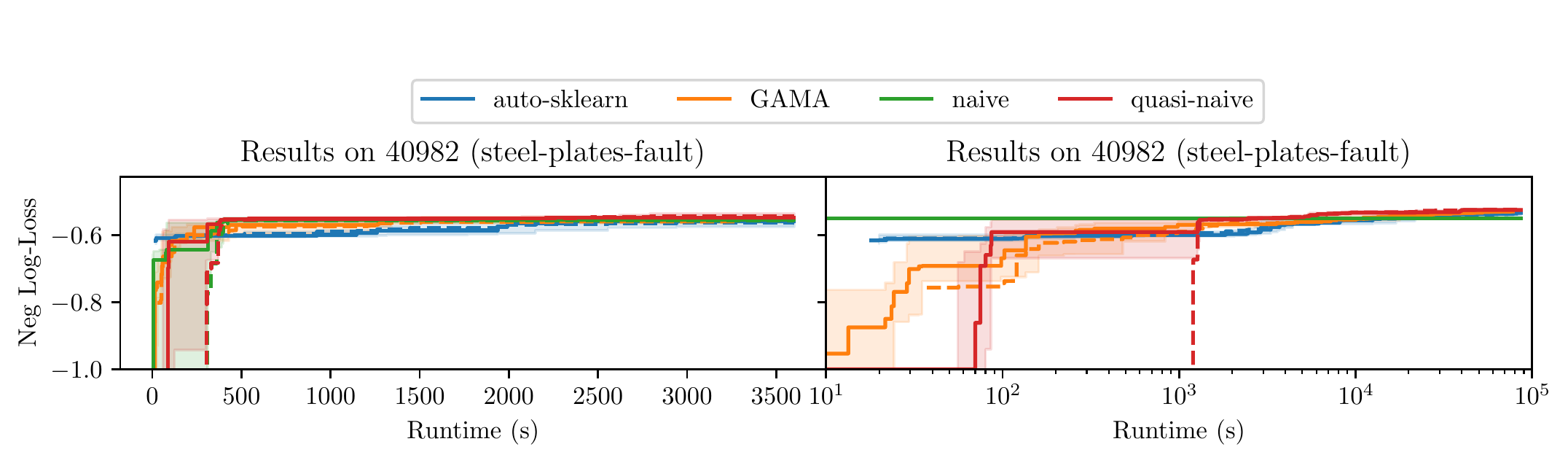}\\[-2em]
\includegraphics[width=\textwidth]{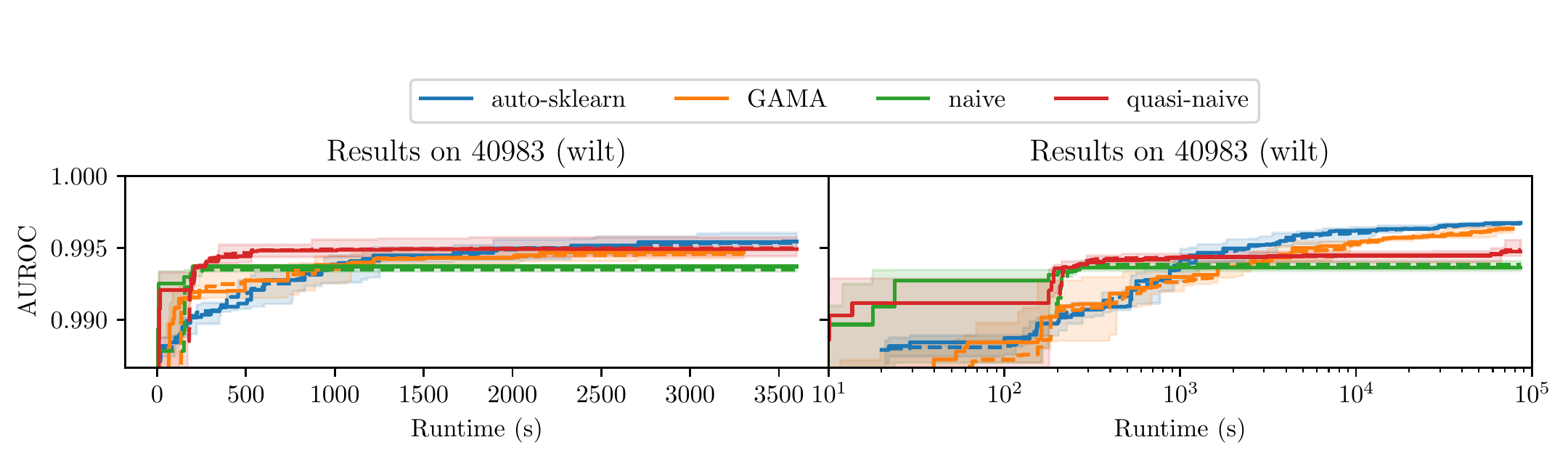}\\[-2em]
\includegraphics[width=\textwidth]{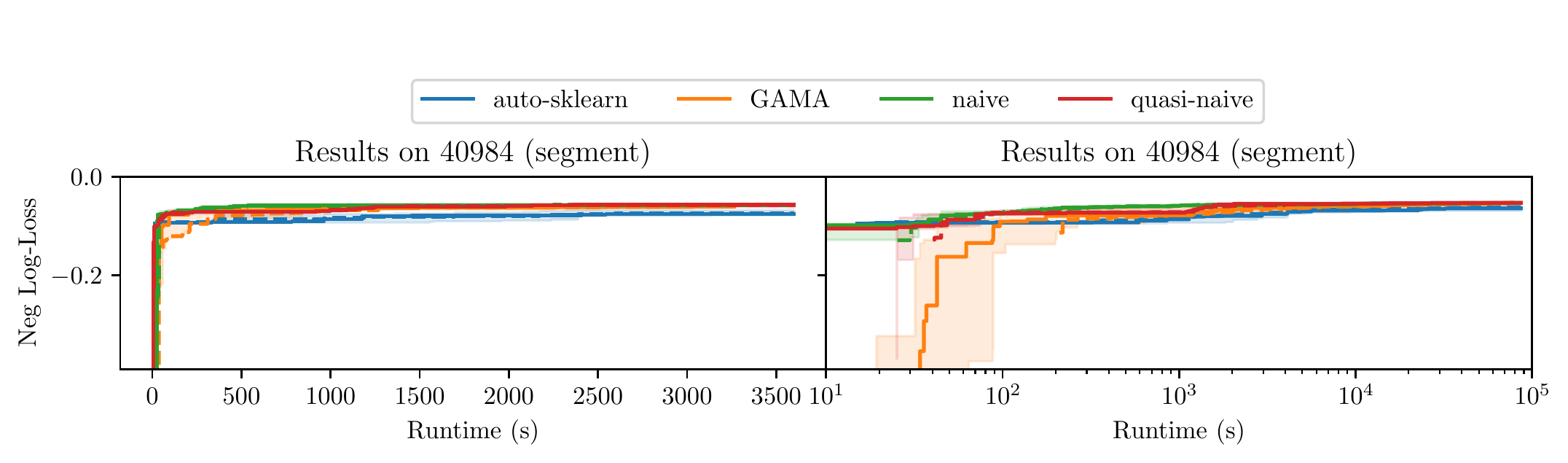}\\[-2em]
\includegraphics[width=\textwidth]{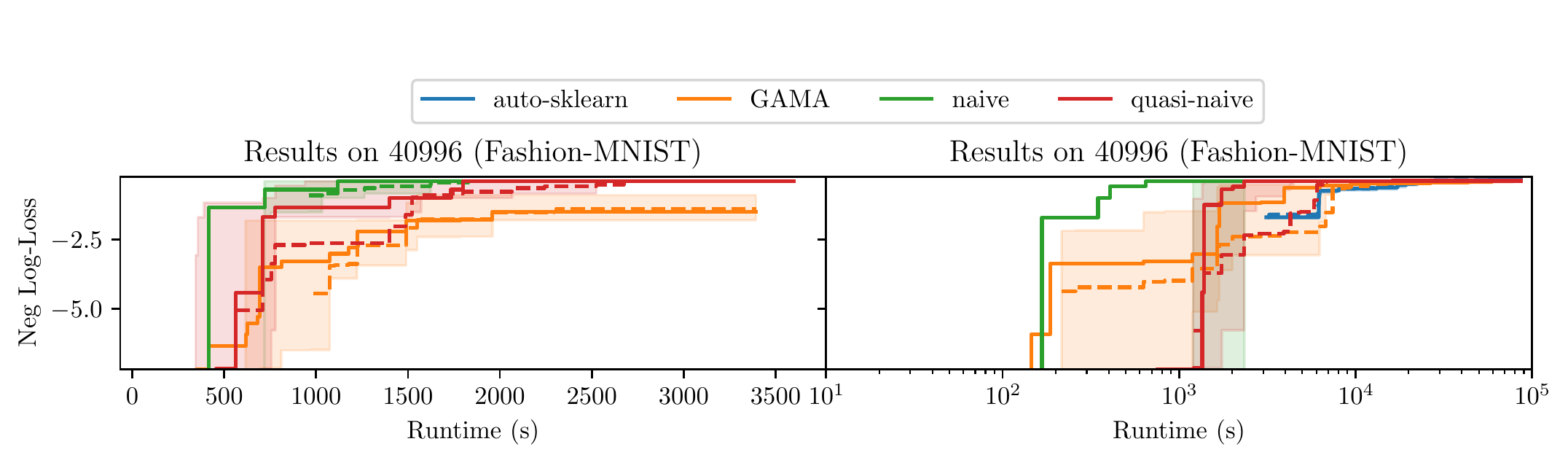}\\[-2em]
\includegraphics[width=\textwidth]{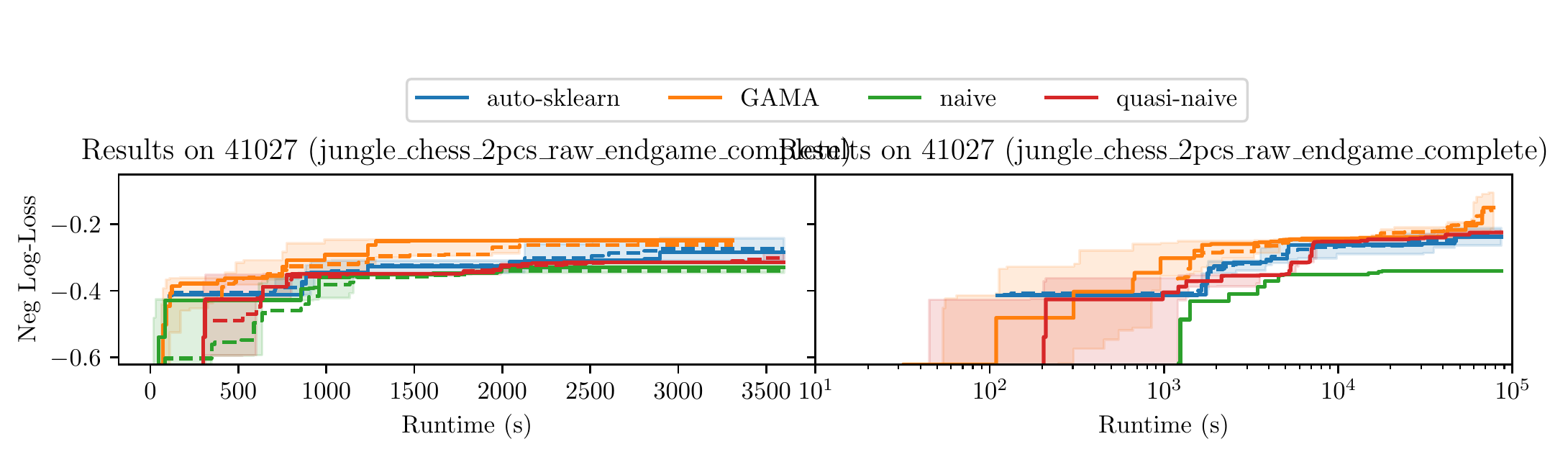}\\[-2em]
\includegraphics[width=\textwidth]{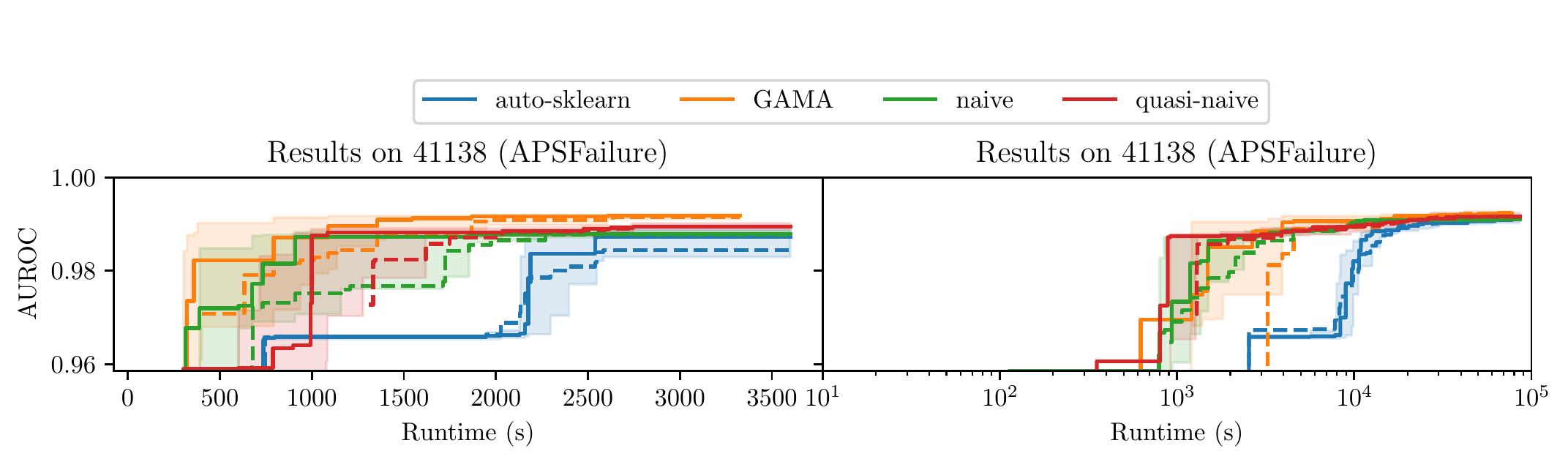}\\[-2em]
\includegraphics[width=\textwidth]{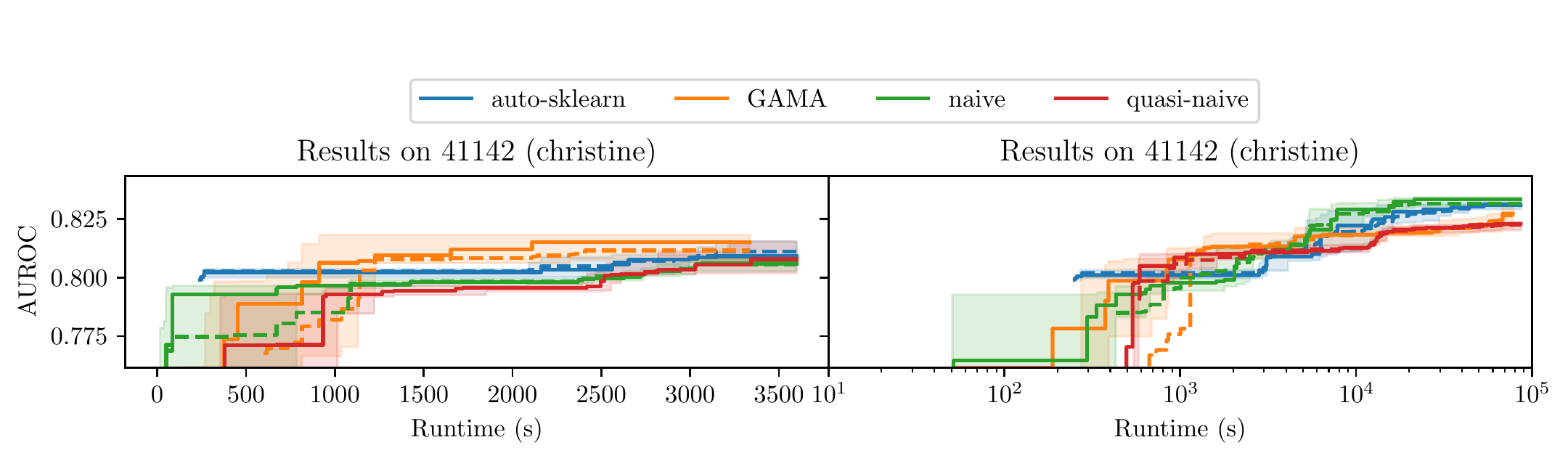}\\[-2em]
\includegraphics[width=\textwidth]{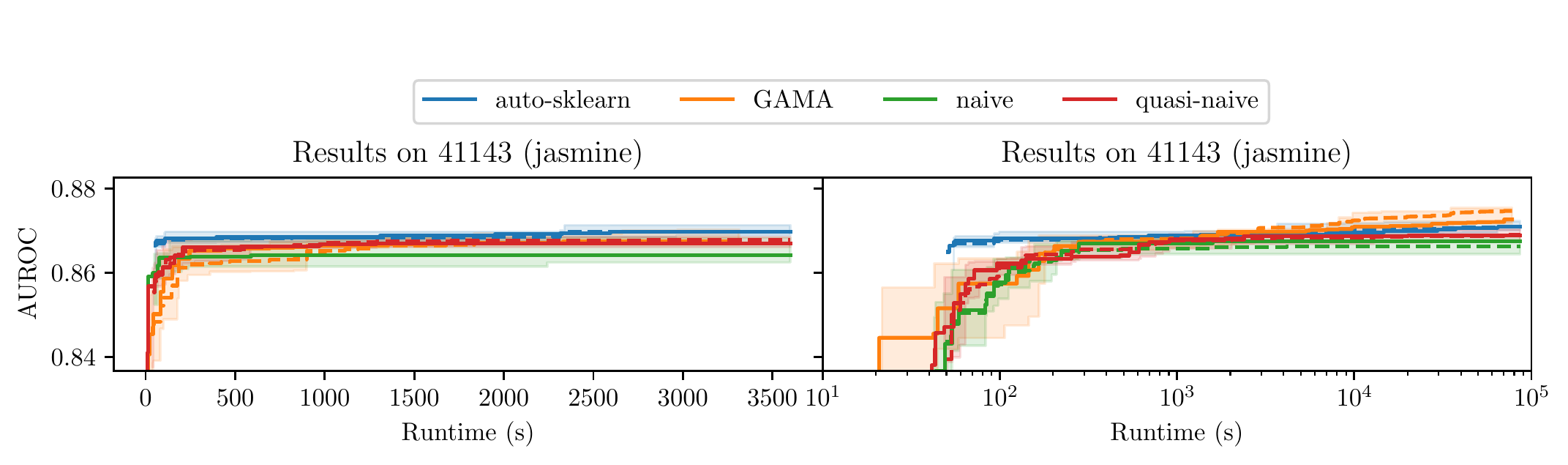}\\[-2em]
\includegraphics[width=\textwidth]{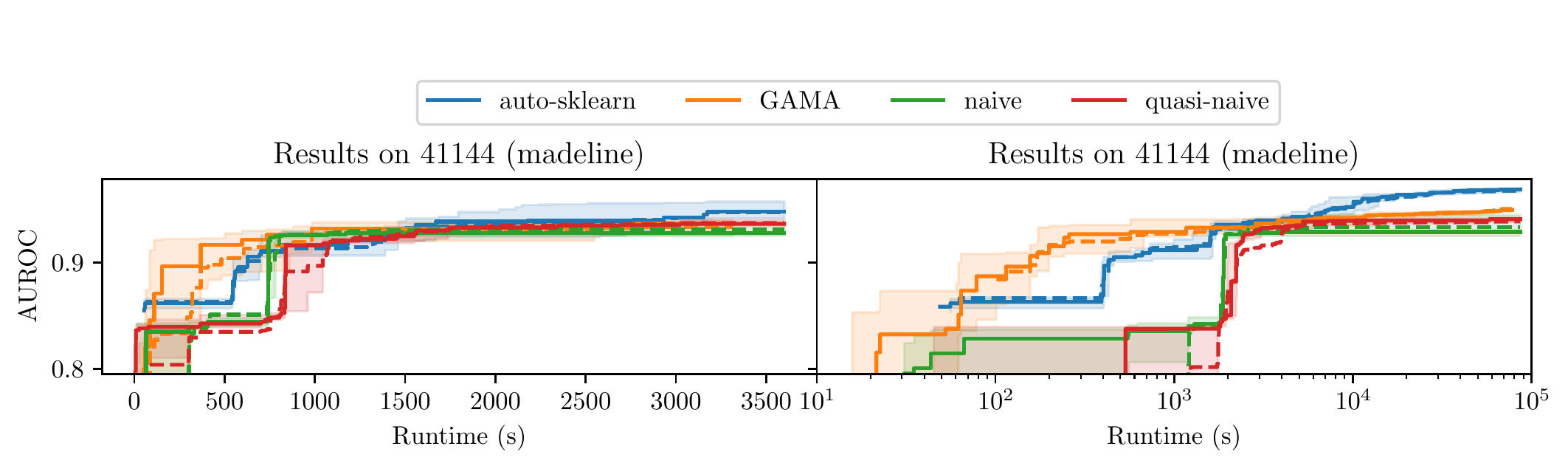}\\[-2em]
\includegraphics[width=\textwidth]{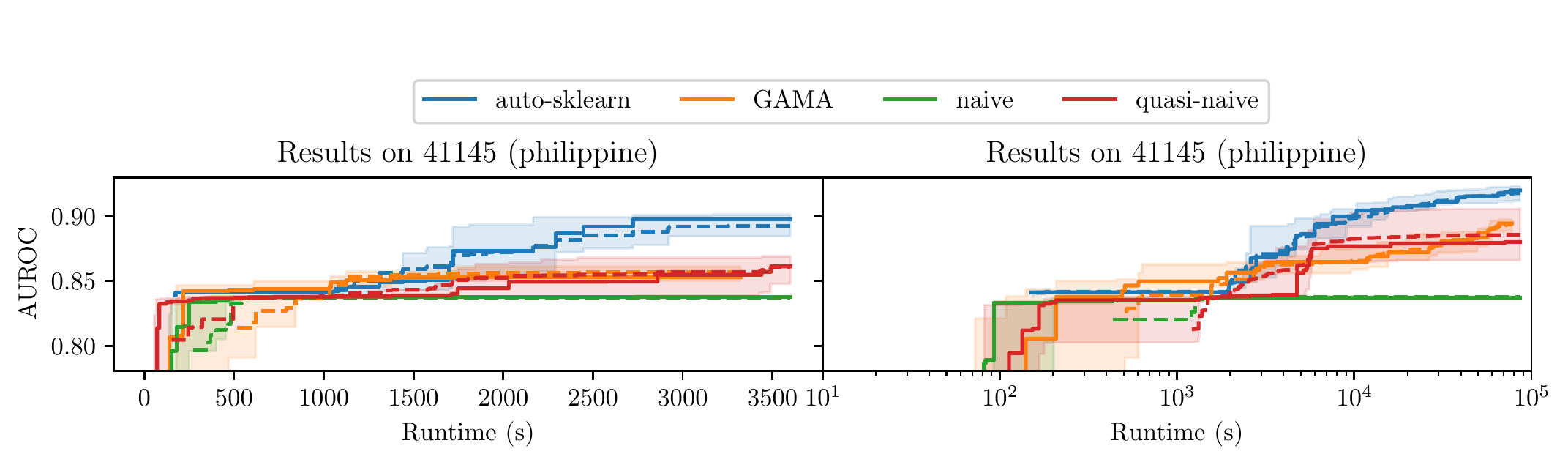}\\[-2em]
\includegraphics[width=\textwidth]{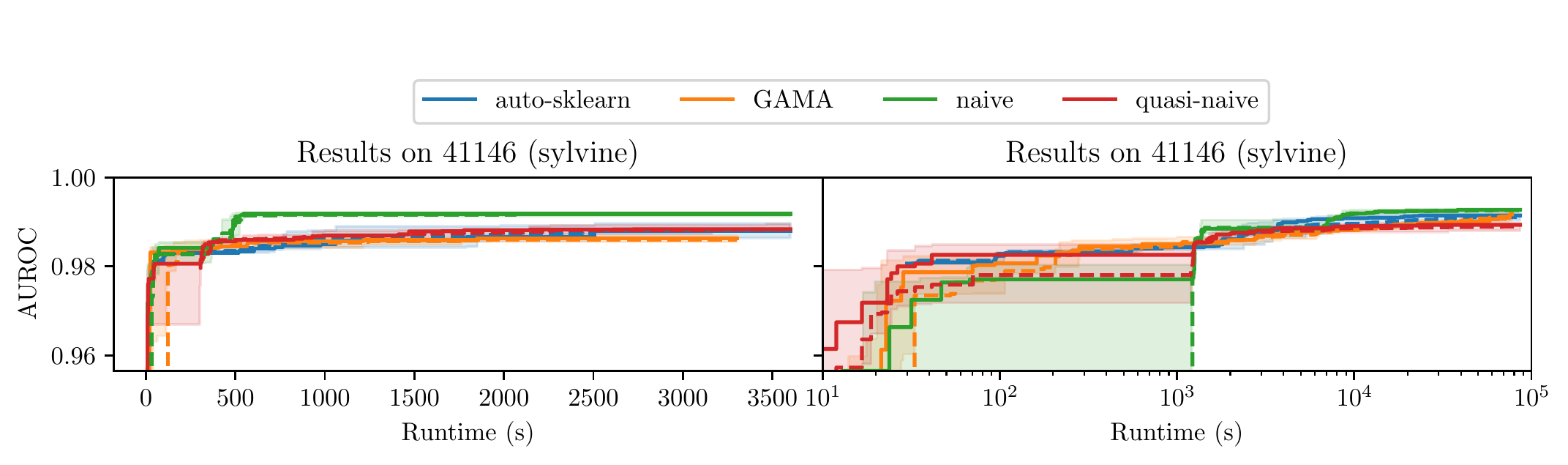}\\[-2em]
\includegraphics[width=\textwidth]{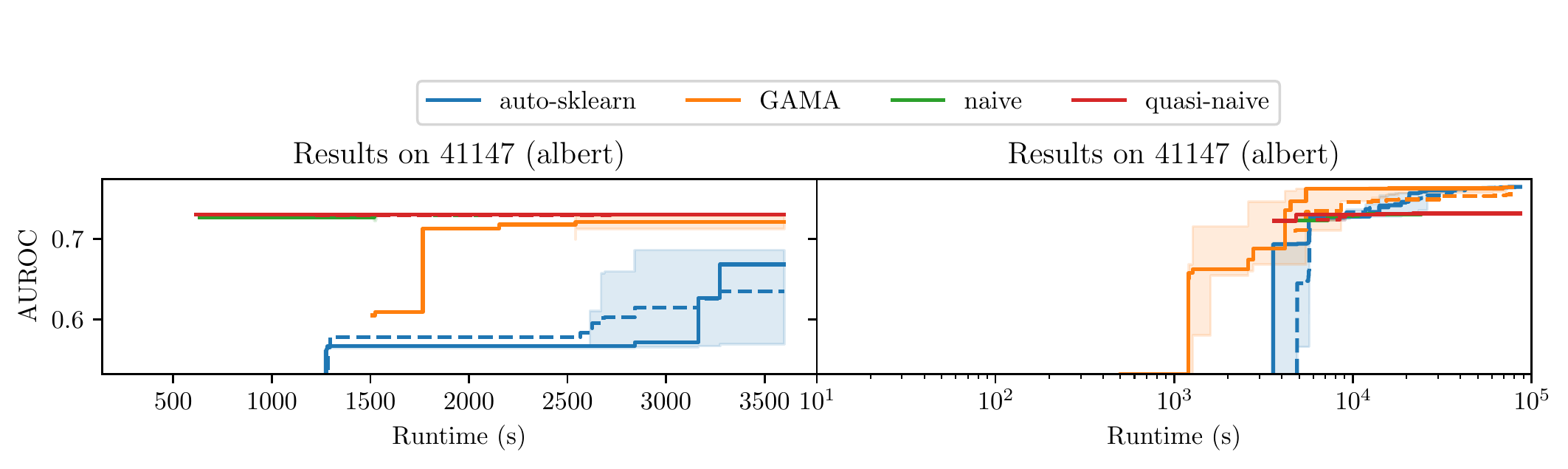}\\[-2em]
\includegraphics[width=\textwidth]{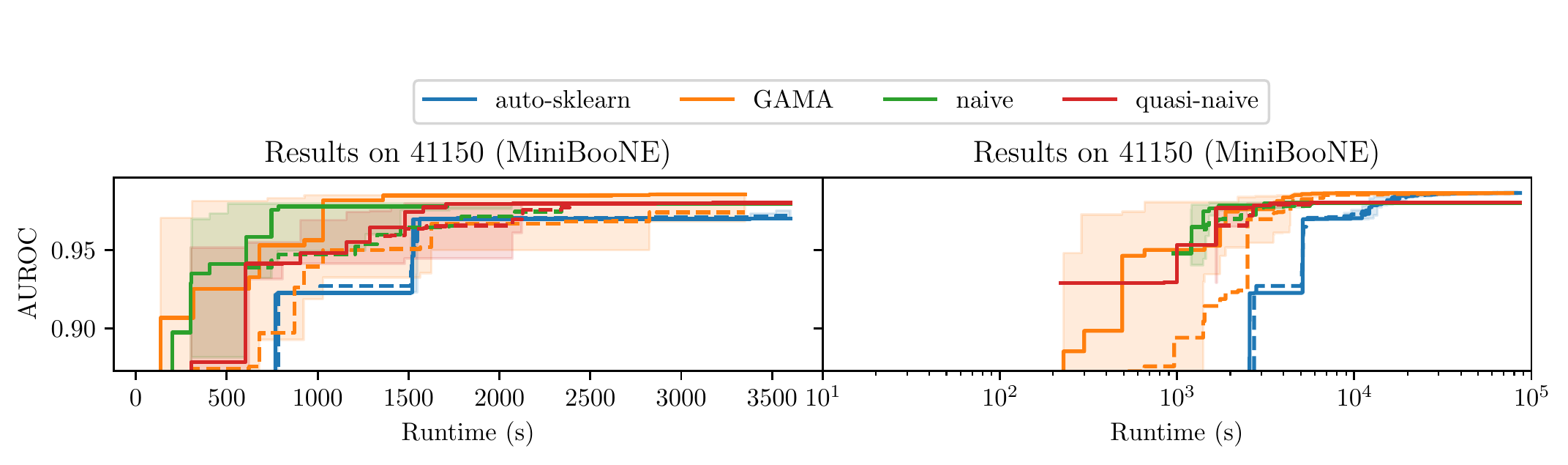}\\[-2em]
\includegraphics[width=\textwidth]{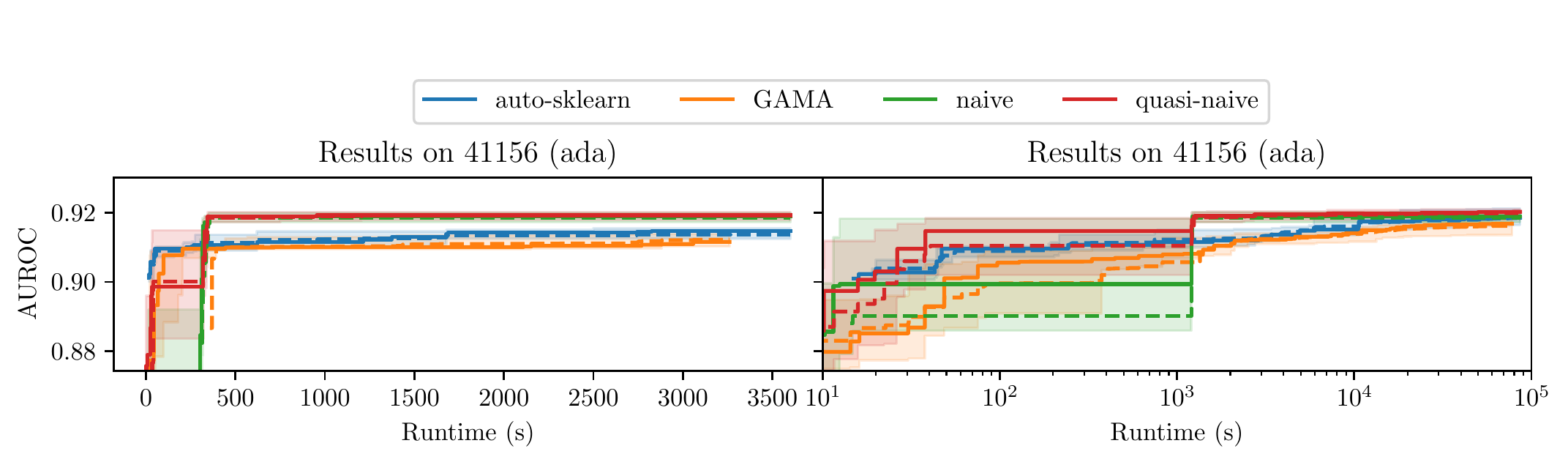}\\[-2em]
\includegraphics[width=\textwidth]{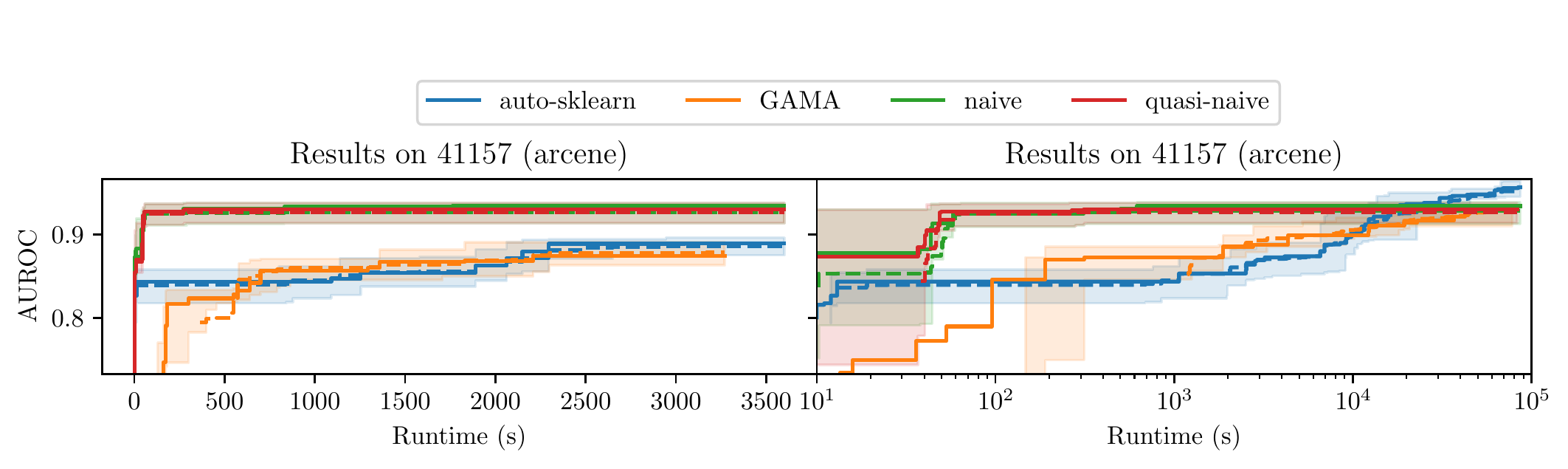}\\[-2em]
\includegraphics[width=\textwidth]{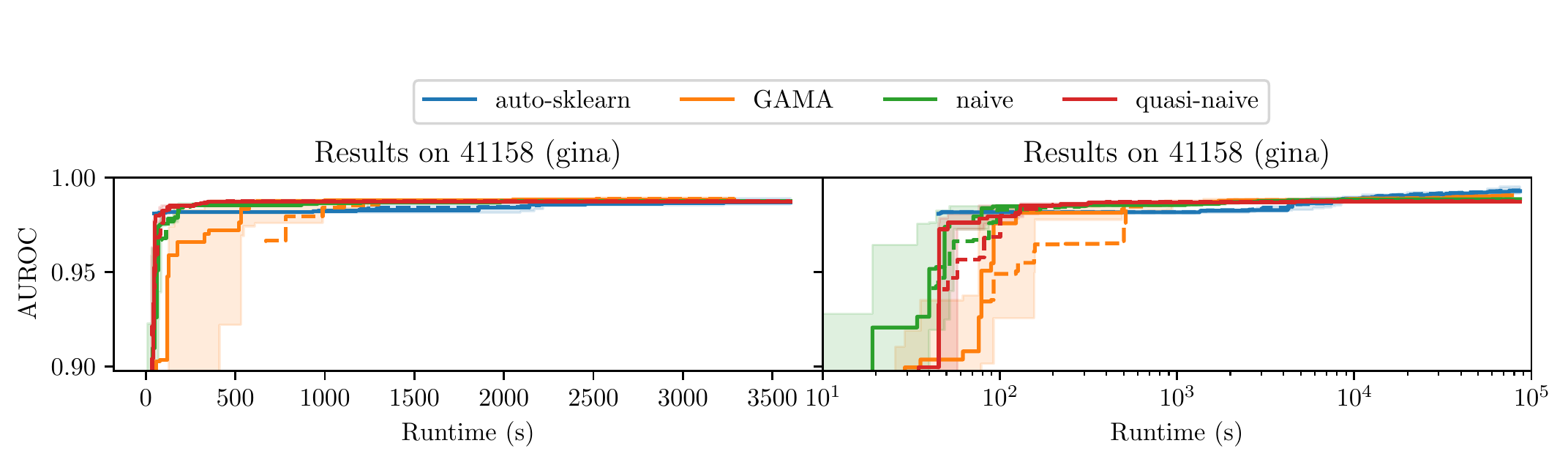}\\[-2em]
\includegraphics[width=\textwidth]{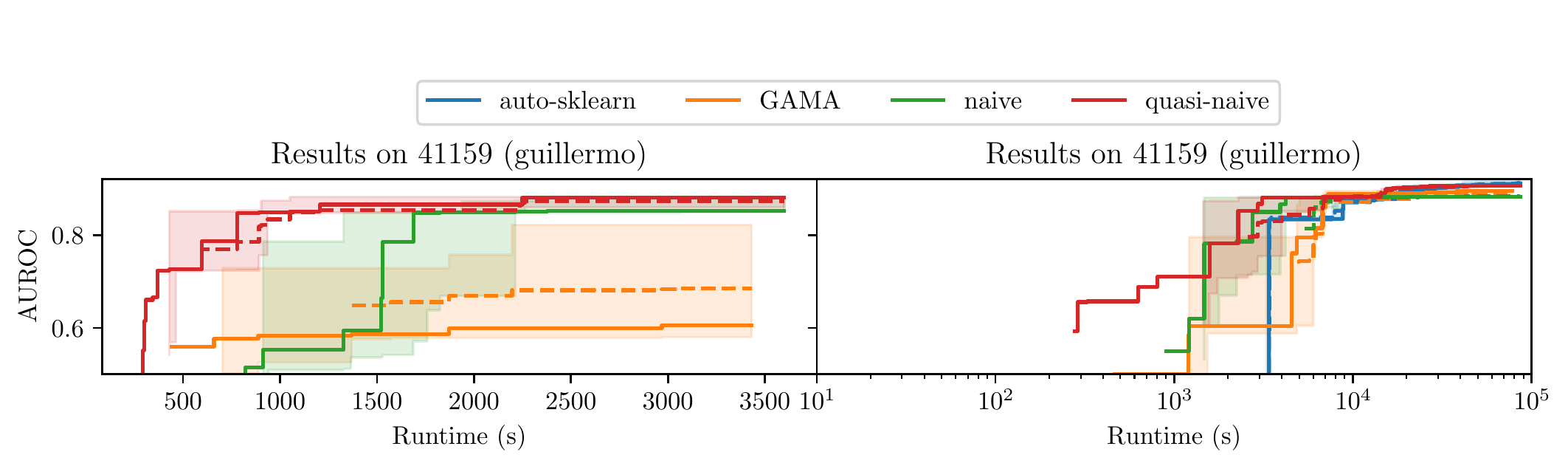}\\[-2em]
\includegraphics[width=\textwidth]{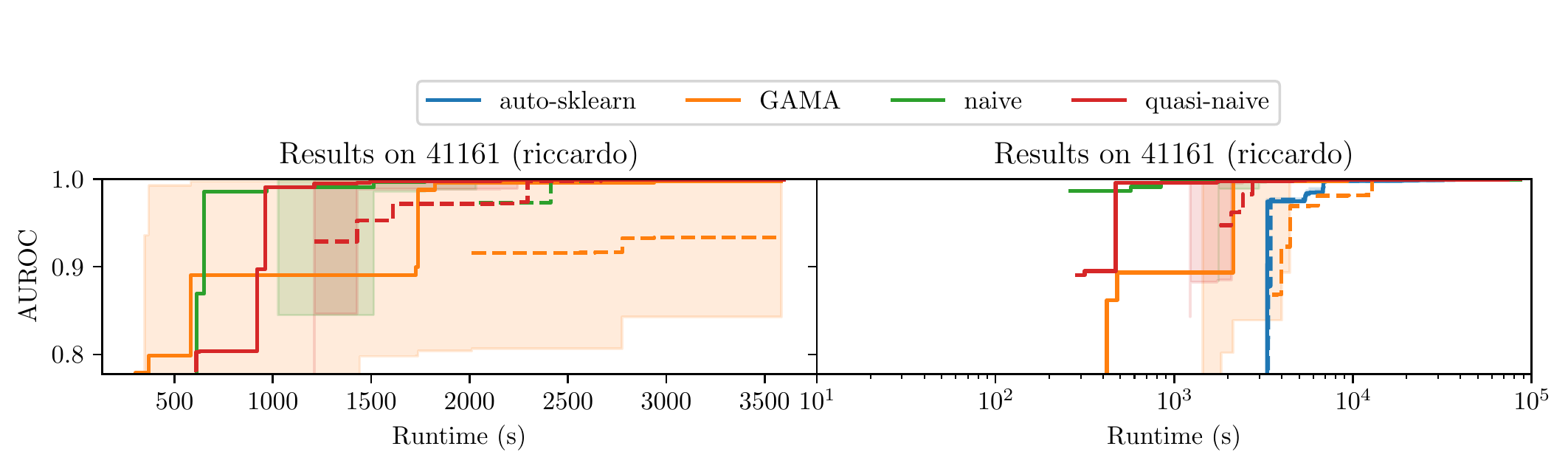}\\[-2em]
\includegraphics[width=\textwidth]{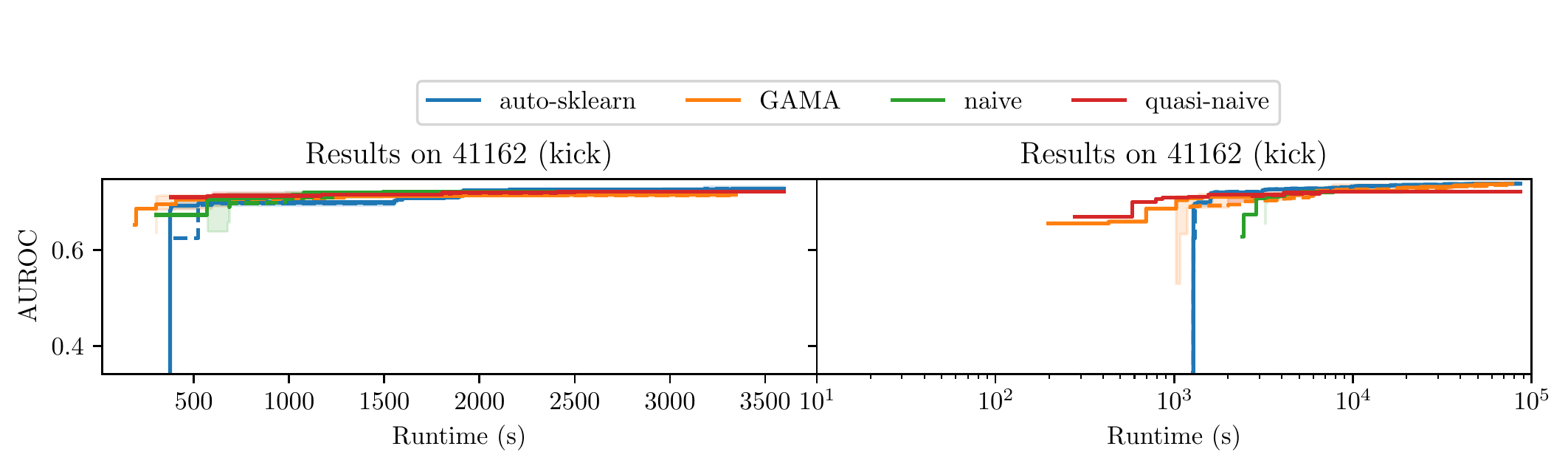}\\[-2em]
\includegraphics[width=\textwidth]{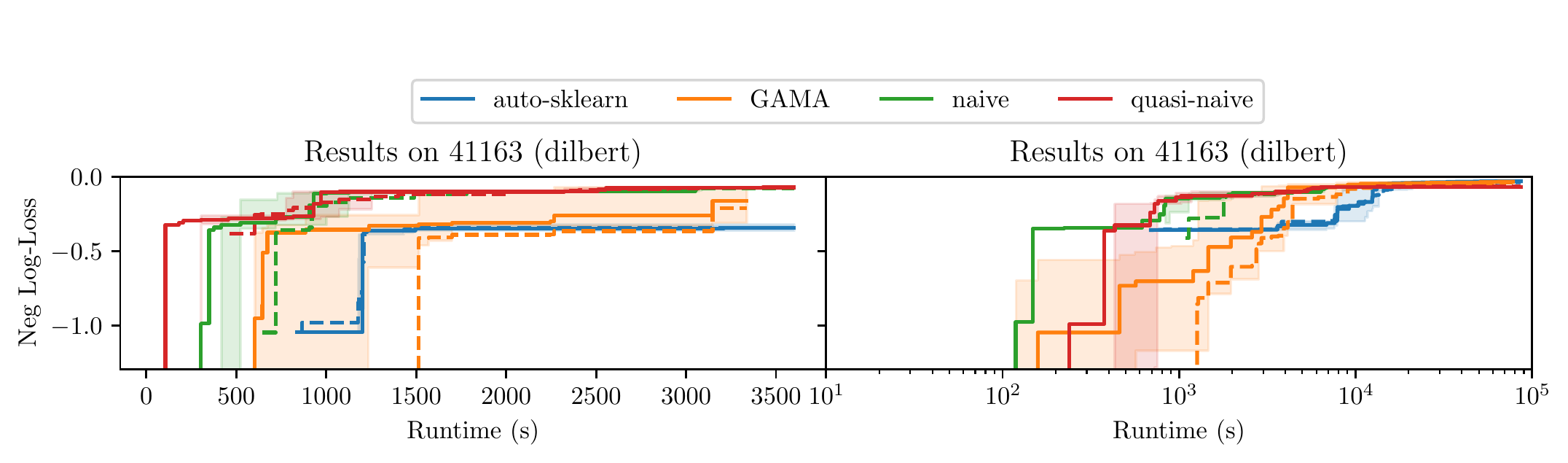}\\[-2em]
\includegraphics[width=\textwidth]{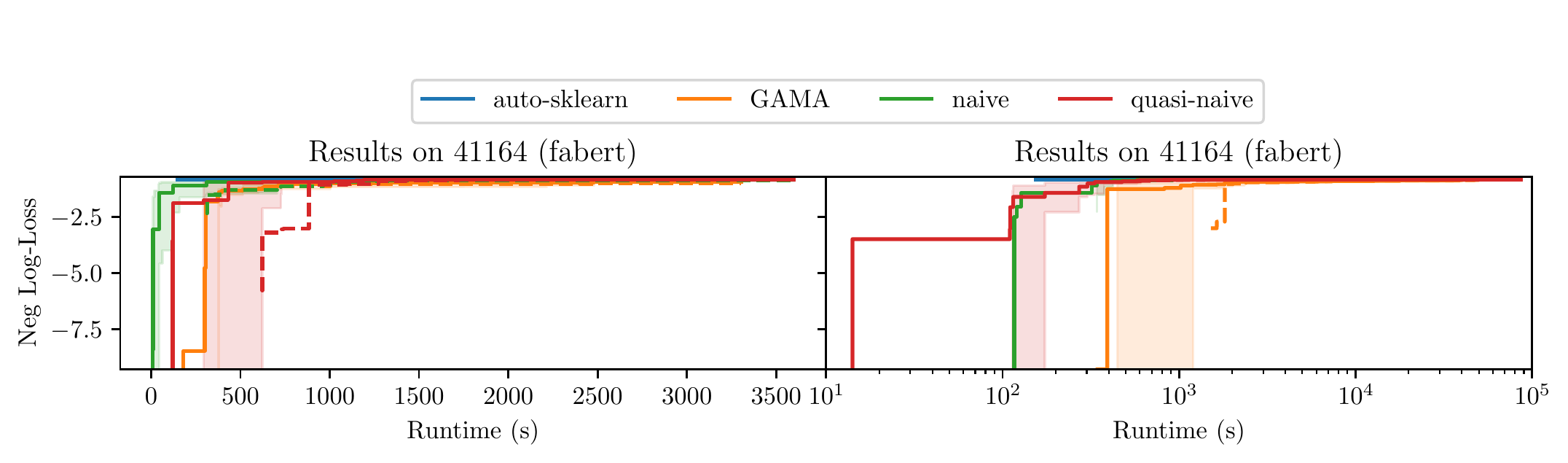}\\[-2em]
\includegraphics[width=\textwidth]{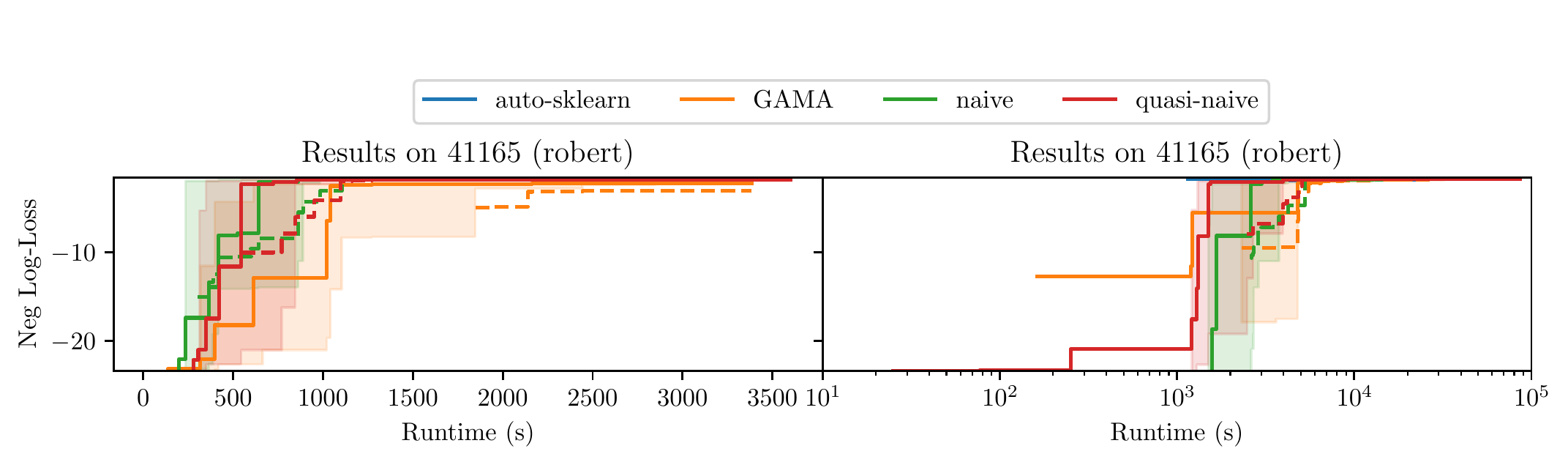}\\[-2em]
\includegraphics[width=\textwidth]{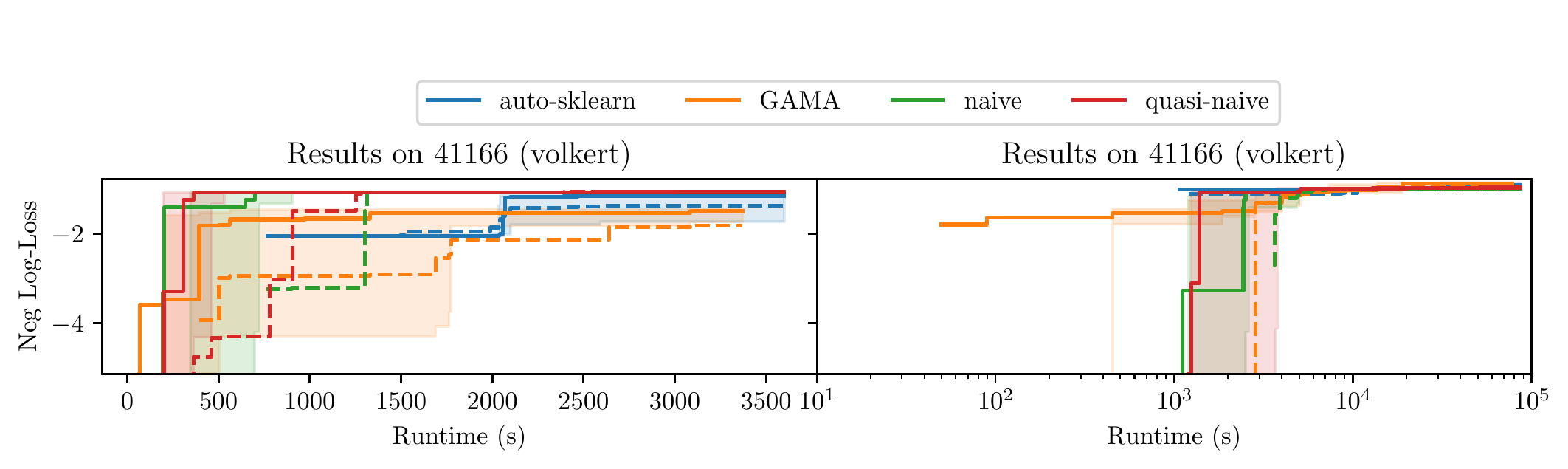}\\[-2em]
\includegraphics[width=\textwidth]{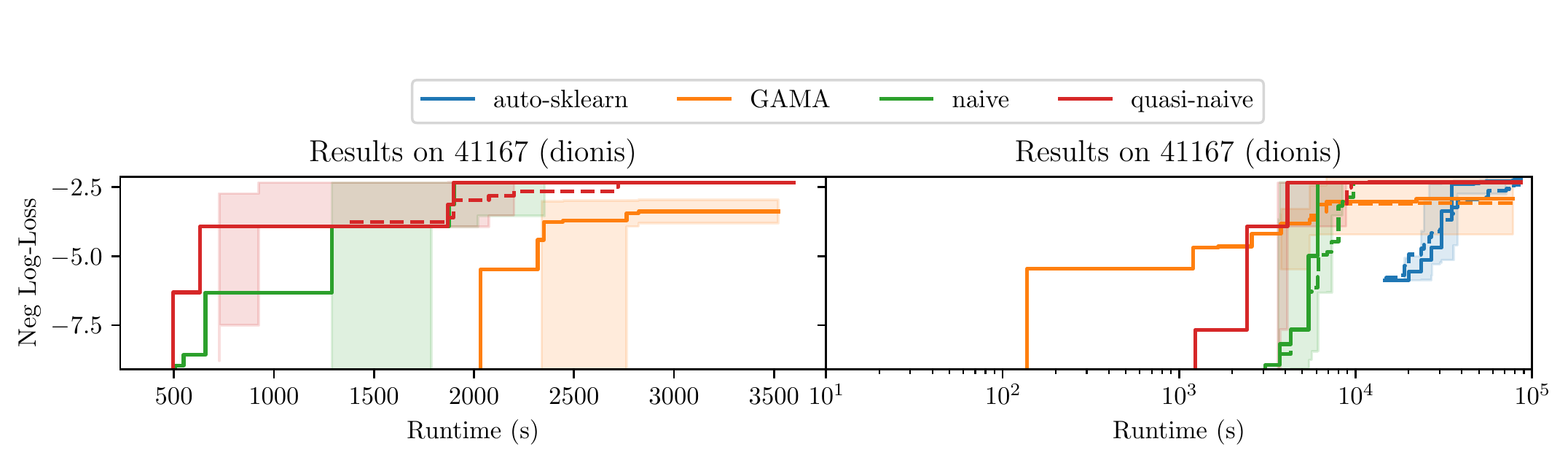}\\[-2em]
\includegraphics[width=\textwidth]{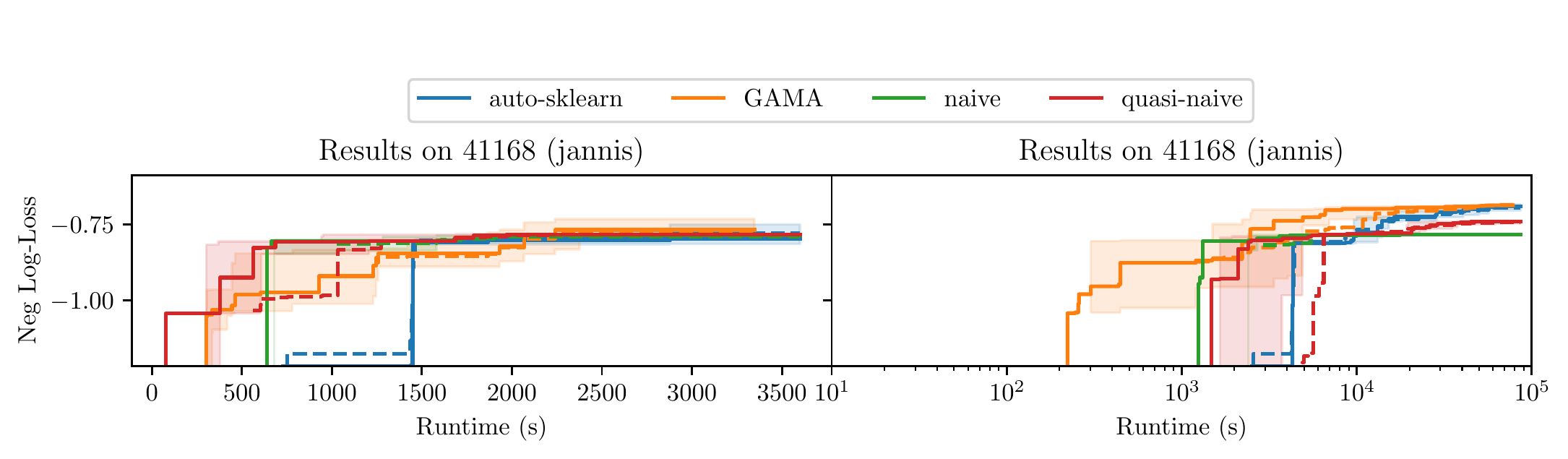}\\[-2em]
\includegraphics[width=\textwidth]{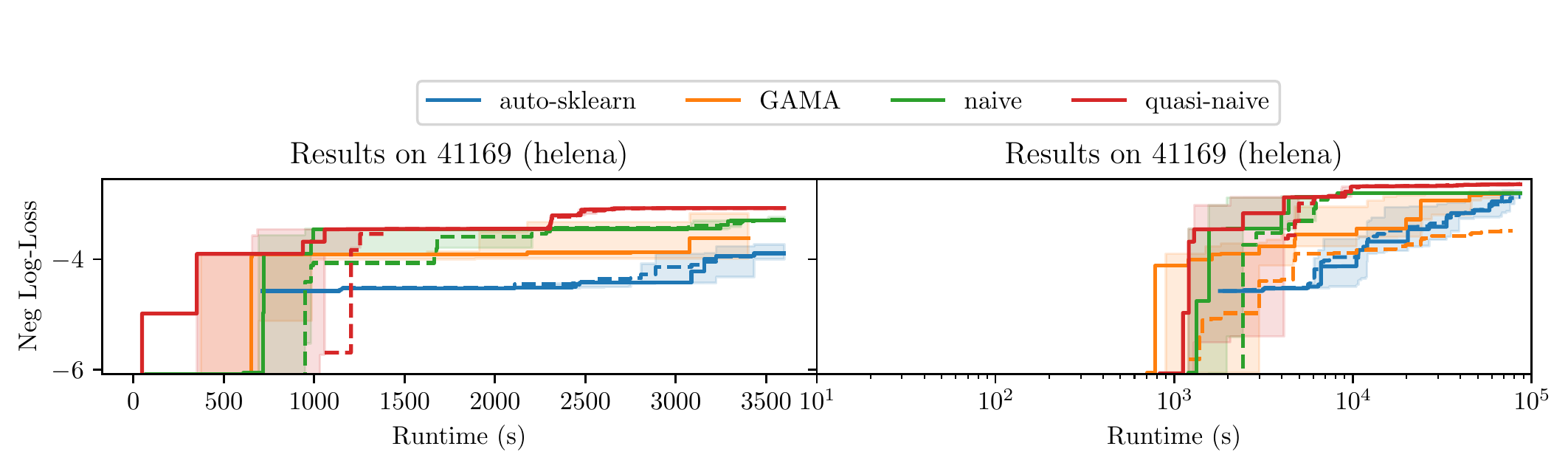}\\[-2em]
\includegraphics[width=\textwidth]{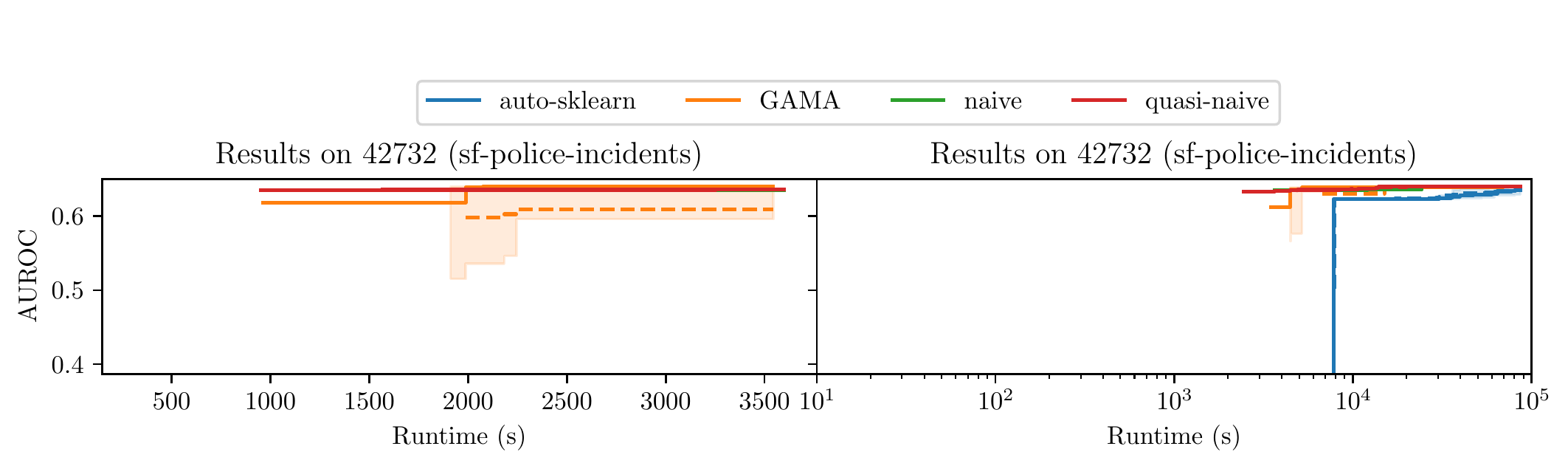}\\[-2em]
\includegraphics[width=\textwidth]{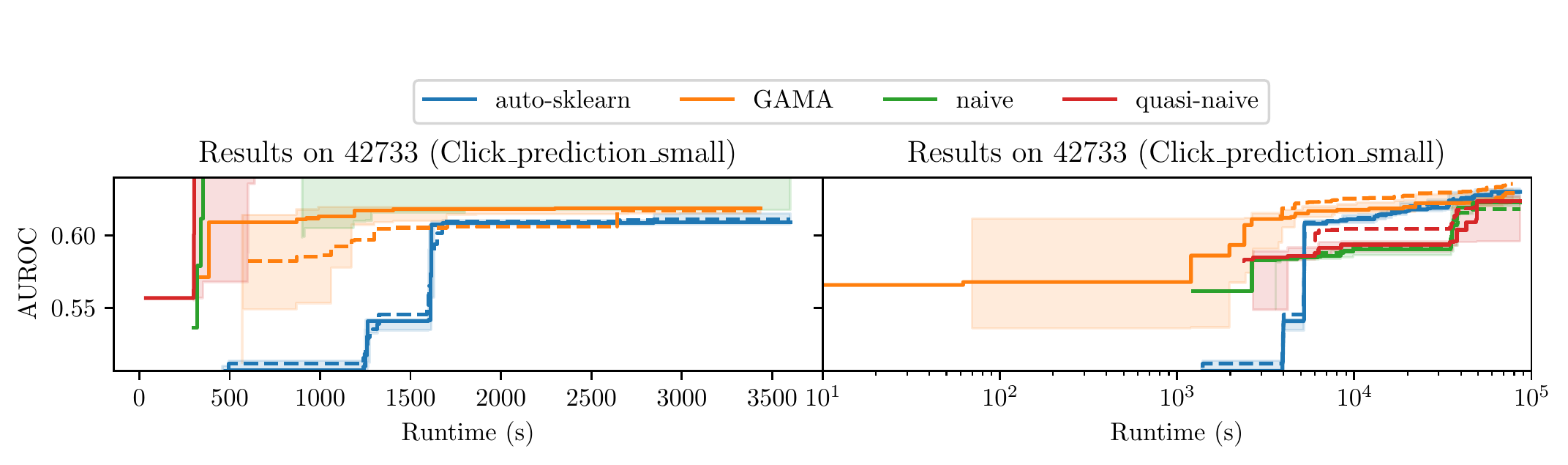}\\[-2em]
\includegraphics[width=\textwidth]{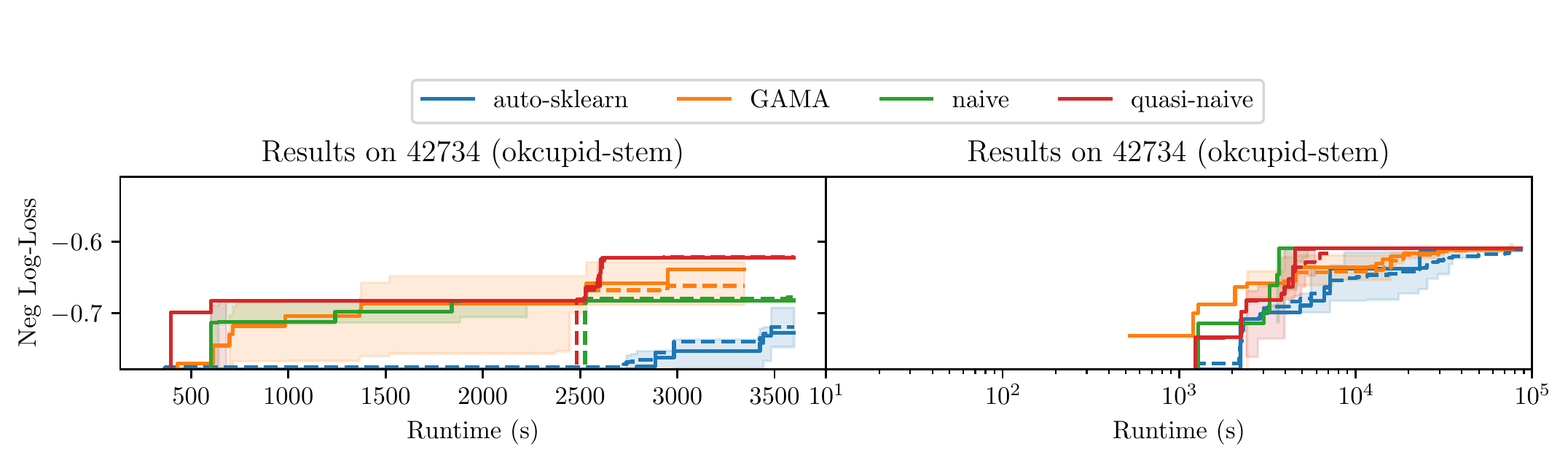}\\[-2em]

\end{document}